\pdfoutput=1
\documentclass[journal]{IEEEtran}
\usepackage{cite}
\usepackage{amsmath,amssymb,amsfonts}
\usepackage{algorithmic}
\usepackage{float}
\usepackage{tabularx}
\usepackage{graphicx} 
\graphicspath{{Figures/}}
\usepackage{subcaption} 
\usepackage{booktabs} 
\usepackage{xcolor}
\usepackage{hyperref} 
\usepackage{comment} 
\usepackage{textcomp}
\def\BibTeX{{\rm B\kern-.05em{\sc i\kern-.025em b}\kern-.08em
    T\kern-.1667em\lower.7ex\hbox{E}\kern-.125emX}}

\begin{document}
\title{Early versus Late RGB–Infrared Fusion for Counter-UAV Detection: A Comparative Study}
\author{\uppercase{Alberto De Zan}\textsuperscript{1}, \uppercase{Denis Tavaris}\textsuperscript{1},
\uppercase{Gian Luca Foresti}\textsuperscript{1}, and \uppercase{Ivan Scagnetto\textsuperscript{1}}\\[6pt]
{\small\textsuperscript{1}Department of Mathematics, Computer Science, and Physics, University of Udine, Via delle Scienze 206, 33100 Udine, Italy}\\[2pt]
{\small Corresponding author: Ivan Scagnetto (e-mail: ivan.scagnetto@uniud.it).}\\[10pt]

\fbox{\parbox{0.94\textwidth}{\footnotesize \textbf{Note:} This arXiv version differs from the article published in \emph{IEEE Access}, vol.~14, pp.~149702--149712, 2026, DOI: \href{https://doi.org/10.1109/ACCESS.2026.3735919}{10.1109/ACCESS.2026.3735919}. It includes revisions requested by the reviewers during the final review round that could not be incorporated into the published version. The main changes include additional references, a more detailed description of the paired dataset, an additional fusion model (IFCNN) in the comparison, an extended discussion of model performance and an explanation of the distance threshold used for late fusion.}}}

\markboth{Alberto De Zan, Denis Tavaris, Gian Luca Foresti, Ivan Scagnetto}{Early versus Late RGB–Infrared Fusion for Counter-UAV Detection: A Comparative Study}

\maketitle

\begin{abstract}
This paper presents a novel strategy for UAV detection through visible (RGB) and infrared (IR) fusion, in order to improve accuracy and precision in harsh environments and weather conditions. In this research, two opposite approaches have been studied and implemented: the first concerns image-level (early) fusion, which merges two different spectra into a single image to which one detector is applied. On the other hand, the second approach, called decision-level (late) fusion, focuses on blending together the output bounding boxes of two different detection pipelines, aiming to create a common result that can be applied to both spectra simultaneously. Both strategies have been extensively tested in a dedicated environment, which included alternating low and high brightness scenes, different backgrounds (sky and ground), distances (5, 25 and 60 m), altitudes and zoom levels (1X, 2X and 4X). These tests allowed for the assessment of actual detection capabilities and system reliability in real-world applications. Furthermore, the two fusion strategies were evaluated and compared over a set of aligned images in order to highlight the better approach of the two. The results indicate that early fusion achieves better raw performance, since it incorporates the thermal information from the beginning of the training process. However, late fusion presents an innovative and useful detection workflow that can be applied in cases of sensor or frame misalignment. Both positive and negative aspects have also been pointed out and discussed, in order to give a more comprehensive overview. Finally, the paper outlines implementation considerations and practical recommendations based on operational priorities. Overall, the findings contribute to the research in RGB–IR fusion modalities, with the goal of enhancing UAV detection capabilities.
\end{abstract}


\section{Introduction}
\label{sec:introduction}
\IEEEPARstart{U}{nmanned} Aerial Vehicles (UAVs) are becoming increasingly common in military and civilian applications. They can be used in surveillance, mapping and transporting materials. However, there are safety and security concerns which have to be taken into consideration, since modern UAVs might be able to operate in varied environments and evade traditional surveillance methods. In order to reduce these weaknesses, many advanced sensors are used in current scenarios. For example, visible light (RGB) cameras are widely adopted, as they offer excellent spatial resolutions and reduced costs \cite{zhao2022vision}. However, they have problems in low light, adverse weather conditions \cite{Samadzadegan2022DroneDetection} and crowded backgrounds. Concerning this particular aspect, the terrain texture may be difficult to separate from the drone's surface, since UAVs are usually made from composite and camouflaged materials. On the other hand, infrared (IR) cameras work effectively in low light situations and night scenes. Nevertheless, they have reduced thermal and spatial resolution and are expensive. Therefore, this research focuses on fused solutions for UAV detection, aiming to combine the strengths of both sensors \cite{skorka2022tradeoffs}.\\
There are some studies in Computer Vision which have explored the capabilities of sensor fusion strategies in object detection systems \cite{sun2024visible}. However, the Counter-UAV setting still remains an underexplored research area for fusion. Most of the existing approaches require significant adjustments and fixes to the sensors before fusion can be applied. In this paper, the following innovative fusion pipelines for YOLO-based Counter-UAV detection have been adopted and evaluated:
\begin{itemize}
	\item An image-level (early) fusion that composes synchronized and aligned RGB and IR inputs into a unified representation for a single detector.
	\item A decision-level (late) fusion that processes RGB and IR streams independently and merges their detections post-inference.
\end{itemize}

This research presents two logical strategies and the framework designed and applied for each fusion type.\\
Finally, in order to present a complete evaluation, extensive performance analyses have been conducted, including quantitative metrics (such as average detection precision, recall and F1-score) and qualitative visualizations through image samples.

\paragraph{Synopsis.} This work presents two RGB–IR fusion architectures for drone-to-drone detection using YOLO: an image-level (early) fusion combining synchronized inputs into a single stream, and a decision-level (late) fusion merging outputs from separate detectors. Evaluated across varied conditions, the study compares accuracy, efficiency, and robustness, offering practical guidance for Counter-UAV deployment.

\section{Related Work}
\label{sec:RelatedWork}

\subsection{Detection Techniques}
The detection of small unmanned aerial vehicles (UAVs) has attracted increasing attention due to their rapid proliferation and the associated security implications \cite{seidaliyeva2024advances}. Early vision-based UAV detection methods relied primarily on classical image processing and handcrafted feature extraction techniques. Background subtraction, motion-based segmentation and orientation detection were commonly employed in static-camera surveillance scenarios, leveraging temporal consistency to isolate moving aerial targets \cite{9743918}. Feature-based approaches utilizing Histogram of Oriented Gradients (HOG) \cite{8588047}, Haar-like features and Scale-Invariant Feature Transform (SIFT) \cite{liu2024boost} descriptors were also explored for detecting aerial objects against sky backgrounds. While computationally efficient, such methods often suffer from limited robustness under dynamic backgrounds, illumination changes and cluttered scenes.\\
The emergence of deep learning has significantly advanced object detection performance \cite{zamri2024deep}. Two-stage detectors, such as Faster R-CNN \cite{7485869}, achieve high detection accuracy through region proposal mechanisms followed by classification and regression refinement. In addition, they offer strong capabilities in feature extraction tasks. However, their computational complexity can limit real-time applicability in resource-constrained operational environments. In contrast, one-stage detectors, including Single Shot MultiBox Detector (SSD) \cite{10.1007/978-3-319-46448-0_2} and the You Only Look Once (YOLO) \cite{7780460} family, directly predict bounding boxes and class probabilities in a unified framework, enabling significantly higher inference speeds \cite{hidayatullah2025yolov8}.\\
Moreover, models based on transformers, such as Vision Transformers (ViT) \cite{9716741,rebbapragada2024c2fdrone}, DEtection TRansformer (DETR) \cite{10.1007/978-3-030-58452-8_13} and Spatio-Temporal Transformer \cite{sangam2023transvisdrone}, have increased overall robustness by capturing long-range dependencies within images. Furthermore, advanced multimodal deep learning models with hybrid CNN-transformer architectures enhance both feature extraction and decision making through data obtained from numerous sensors. Additionally, the integration of multi-sensor systems into embedded platforms has substantially increased their value in practical applications, including drone surveillance and search-and-rescue operations, demonstrating the effectiveness of these multi-sensor approaches.\\
Recent UAV vision research has also explored multimodal approaches beyond conventional object detection. Lu et al. \cite{lu2025mtcnet} proposed MTCNet, a multimodal network for time-sensitive target concealment in UAV aerial imagery, while Tang et al. \cite{tang2025r2ploc} introduced R2PLoc, a region-to-point framework for UAV visual geo-localization. These works further demonstrate the increasing adoption of multimodal and hierarchical visual representations in UAV-based perception systems. Classification-and-regression-based detectors, such as YOLO, have become particularly attractive for UAV detection due to their favorable trade-off between accuracy and latency and wide compatibility with many devices. Several studies have adapted YOLO architectures for small-object detection, aerial surveillance and drone monitoring tasks \cite{aldosari2024drone}. Modifications such as multi-scale feature aggregation and anchor box optimization have been proposed to improve sensitivity to small aerial targets. Nevertheless, most existing UAV detection approaches rely solely on visible-spectrum imagery, which inherently limits performance under low-light, night-time or adverse weather conditions \cite{bilal2024cost}. These limitations motivate the exploration of complementary sensing modalities to enhance robustness in real-world Counter-UAV applications, potentially involving other drones in the detection process \cite{ashraf2021dogfight}.

\subsection{Fusion Techniques}
Overall, RGB and infrared modalities exhibit complementary characteristics. Therefore, their integration represents a promising direction for enhancing UAV detection performance in operational Counter-UAV scenarios \cite{9916096,YANG2024104179}. Methods such as deep fusion networks and attention mechanisms have shown significant improvements in accuracy, particularly in scenarios such as low-visibility conditions and occlusions \cite{Wang_Chen_Choi_Kuo_2019}, but also in challenging cases such as the resemblance between drones and birds. Furthermore, several studies have shown interest in RGB and IR image fusion methods, with the goal of creating highly informative multi-modal images that integrate features from both spectra. These techniques employ two-scale image decomposition \cite{6423909}, contrast enhancement \cite{liu2022infrared} and swin transformer \cite{ma2022swinfusion}. However, they usually need aligned image pairs for achieving fusion, which are not always available due to the sensors' physical limitations. For this reason, some fusion methods have been developed to also integrate image registration and alignment \cite{tang2022superfusion,ozer2024visirnet}. However, recent works have explored approaches for fusing spatially misaligned yet temporally synchronized images \cite{hao2025tmrgbtd2d}. Furthermore, multi-modal object detection has already been explored by Yue Cao et al. \cite{cao2023multimodal} and Fan Xu et al. \cite{xu2024antidrone}. More recently, Zhang et al. proposed CCSFuse \cite{zhang2026ccsfuse}, a collaborative compensation and selective fusion framework specifically designed for UAV-based RGB-IR object detection, further demonstrating the growing interest in feature-level fusion for multimodal aerial detection. Some studies also involve Bayesian fusion for enhancing perception \cite{chen2022eccv}. Some studies also involve Bayesian fusion for enhancing perception \cite{chen2022eccv}. Modern approaches also integrate additional data sources such as LiDAR and radar, in order to enhance depth perception and robustness, although these methods are mainly oriented toward pedestrian detection \cite{guan2018fusion}. In the UAV domain, RGB-T fusion detectors such as SF-YOLO \cite{zhang2023sfyolo} have also been proposed for drone scenes. Nevertheless, despite these recent advances, publicly available UAV detection methods operating on multimodal fused data remain comparatively limited, particularly when considering practical constraints such as sensor misalignment and real-time deployment.\\
After this brief introduction, in the following subsections we delve deeper into the three most important fusion methods families: early-fusion, mid-fusion and late-fusion.

\subsubsection{Early-Fusion (Pixel-Level)}
It combines the information of the two modalities at the pixel level. It is the most intuitive fusion approach, where IR and RGB images are concatenated to generate a 4-channel image, which is then fed into a regular object detection architecture. Several image-level fusion networks have been proposed in the literature, ranging from autoencoder-based methods such as DenseFuse \cite{li2019densefuse} to illumination-aware architectures such as PIAFusion \cite{tang2022piafusion}, IAIFNet \cite{yang2024iaifnet} and FCDFusion \cite{li2024fcdfusion}. These techniques rely on dedicated, relatively complex fusion architectures to better merge the images in the two spectra. However, as Wagner et al. show \cite{wagner2016multispectral}, pixel-level fusion was outperformed by deeper fusion variants in their pedestrian detection experiments, even though there are ways to improve it, such as the ones presented by Tianyi Zhao et al. \cite{zhao2024ieee}.
Since aligned RGB-Thermal datasets are not available, the method developed by Lizhi Yang et al. \cite{yang2022drone} has also been evaluated, which uses a CycleGAN to create aligned synthetic RGB/IR data \cite{zhu2017unpaired,isola2017image}. In fact, it is important that the image pairs are perfectly aligned before performing early-fusion. In a more simplified way, this architecture can be described as a neural network composed of two different branches, dedicated respectively to RGB and IR image processing. In addition, this network presents a decision layer, which chooses how much importance to give to which of the two sub-networks. \\
In order to merge RGB and IR images, it is possible to use Python Imaging Library (PIL or Pillow) or dstack. The approach that these two techniques follow is based on the raw channel information overlapping. In fact, they take into consideration the bit depth of RGB images, which refers to the number of bits used to represent the color of a single pixel. It is usually 24 bits (8 bits for each color channel) plus other 8 bits for transparency (alpha channel). However, the transparency channel is not always present and it depends on the extension in which the image file is saved. In fact, if more than 3 channels are needed, the \verb|.png| format needs to be considered, since \verb|.jpg| supports a maximum of 3 channels. For IR images, only their thermal information (8 bits) should be considered, which will be added to the final fused image. Therefore, the resulting merged image will be of 32 bits.

\subsubsection{Late-Fusion (Decision-Level)}
This fusion type merges detection results from separate models, improving robustness by leveraging complementary sensor characteristics. Furthermore, the aligned inputs (RGB and the respective IR counterpart) of the two modalities are fed separately into two unimodal object detection models to generate bounding boxes and the predicted boxes are fused using statistical methods, such as weighted averaging of the bounding box coordinates, which is based on the confidence scores from each modality. This method gives more importance to the bounding boxes with higher confidence scores, and this is a suitable option for the purpose of UAV detection. \\
This method works with both aligned and misaligned images. For this reason, it could be considered more resilient to sensor misalignment with respect to early fusion. However, when the same object appears with different orientations or in different positions across the two images, the resulting bounding boxes may be less precise in terms of localization. In practice, the bounding box tends to expand, and especially in the case of unaligned images, the object is still correctly identified while the box remains largely consistent across both views. As a result, the final bounding box may show slight variations in size and position.\\
In addition to the adopted strategy, there exist also ensemble methods such as probabilistic ensembling \cite{chen2022eccv}, in which a bounding box is considered valid if it is detected by both modalities.

\subsubsection{Mid-Fusion (Feature-Level)}
This strategy, which falls between early and late fusion techniques \cite{gadzicki2020early}, might be the best and most balanced option to choose when trying to develop a computer vision system that exploits sensor fusion. Essentially, this fusion integrates feature maps extracted separately from RGB and IR modalities before applying detection networks. More specifically, features are captured from each of the two spectra using specific convolutional pipelines in the first stages. Subsequently, feature maps are merged using methods like concatenation or sum, which prepare information to be used for the final layers of detection. Taking into account these aspects, the mid-fusion idea has the ability to improve the robustness of computer vision and object detection systems in harsh environments with low light \cite{liu2023self,tian2023rgb}.\\
This type of fusion has also many advantages compared with the other two. First of all, images do not need to be merged into a single one as early-fusion does, but all their channels are kept in the original form, preserving modality-specific information. Nor are the final detection results combined using statistical methods. Therefore, this strategy appears to be more flexible with respect to the input data type and the final detection goals that we may have. In fact, it has already been studied and implemented in some feature alignment methods for RGB and IR images \cite{wang2019rgb}. Recent UAV-specific work further supports this direction. Zhang et al. \cite{zhang2026ccsfuse} proposed CCSFuse, which performs collaborative compensation and selective fusion for RGB-IR object detection, illustrating how modality-specific features can be selectively combined to improve multimodal UAV perception. However, despite these positive aspects, only the other two strategies have been implemented, as mid-fusion implies creating a custom detection architecture, markedly different from the state-of-the-art YOLO one used for the tests. In addition, early-fusion and late-fusion allowed us to obtain useful metrics for comparing the strategies in several scenarios, which were not possible with mid-fusion.

\section{Logical Proposed Architectures}
\label{Sec:Dataset}
The architecture proposed is based on the YOLO framework, since it represents the state-of-the-art in the real-time object detection field. Moreover, two separate fusion pipelines have been developed for enhancing detection performance and accuracy. Finally, a multimodal dataset \cite{11177201} of real-world data on still and moving drones was built, for testing the validity of the following fusion methods. This approach has been preferred to the synthetic one because real acquired data better reflect operational conditions and lead to superior performance. In addition, the actual behaviour of a drone, that is, its ability to rotate, shift and change pose over the three axes (roll, pitch and yaw), has also been considered while collecting data. The paired dataset comprises 400 aligned images (300 for training and 100 for validation, following a 75/25 split) for each fusion strategy evaluated, starting from a base of aligned RGB-IR pairs.

\subsection{Early Fusion}
\begin{figure}[!ht]
	\centering
	\begin{minipage}{0.15\textwidth}
		\centering
		\includegraphics[width=\linewidth]{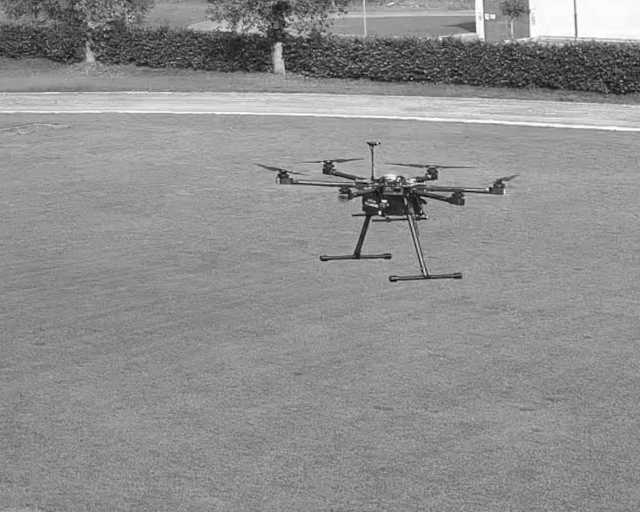}
		\label{fig:red}
	\end{minipage}
	\hfill
	\begin{minipage}{0.15\textwidth}
		\centering
		\includegraphics[width=\linewidth]{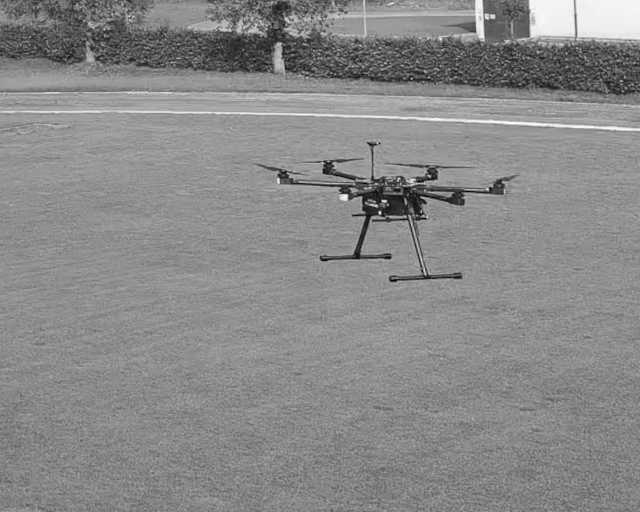}
		\label{fig:green}
	\end{minipage}
    \hfill
	\begin{minipage}{0.15\textwidth}
		\centering
		\includegraphics[width=\linewidth]{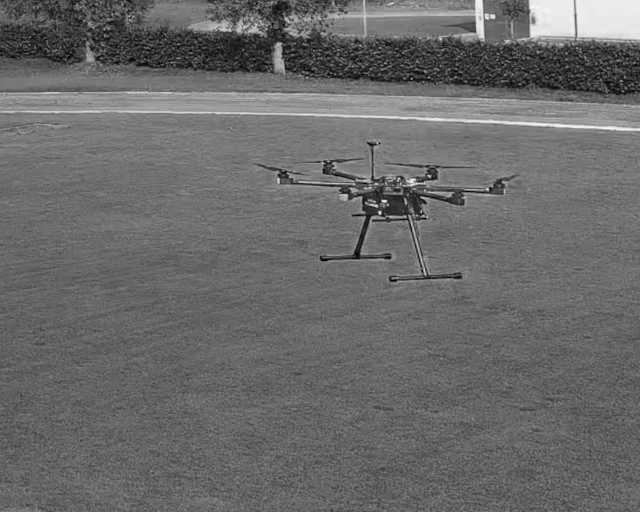}
		\label{fig:blue}
	\end{minipage}
    \caption{From left to right: red, green and blue channel's intensity maps.}
    \label{fig:intensity_maps}
\end{figure}

\begin{figure}[!ht]
	\centering
	\includegraphics[width=\linewidth]{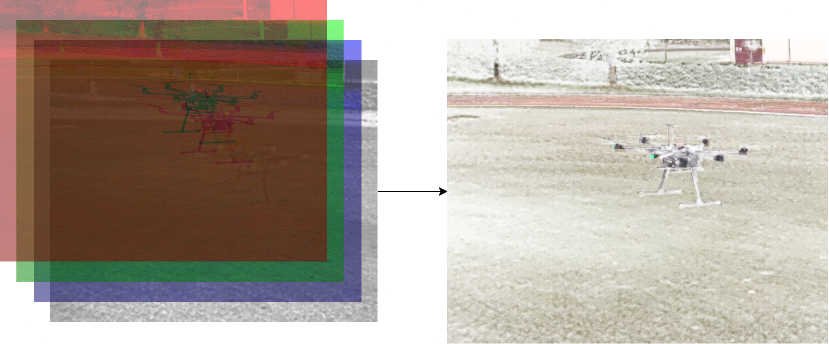}
	\caption{Graphical representation of how layers are overlapped by Pillow library in order to get the resulting image on the right.}
	\label{fig:4_layers}
\end{figure}

\begin{figure}[!ht]
	\centering
	\begin{minipage}[b]{0.2\textwidth}
		\centering
		\includegraphics[width=\linewidth]{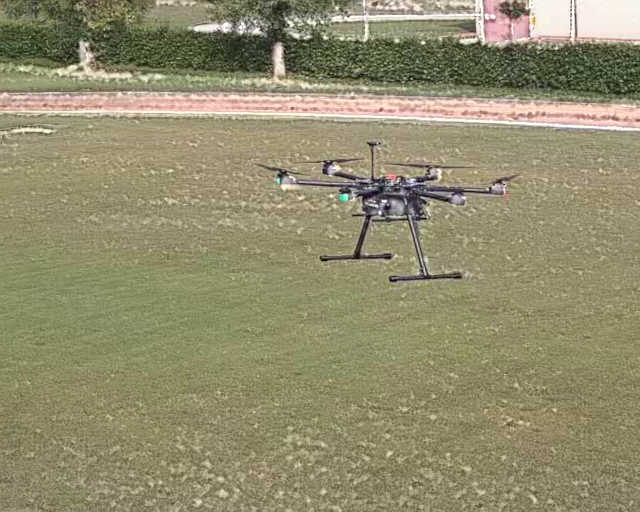}
		\caption{PIAFusion}
		\label{fig:pia}
	\end{minipage}
	\hfill
	\begin{minipage}[b]{0.2\textwidth}
		\centering
		\includegraphics[width=\linewidth]{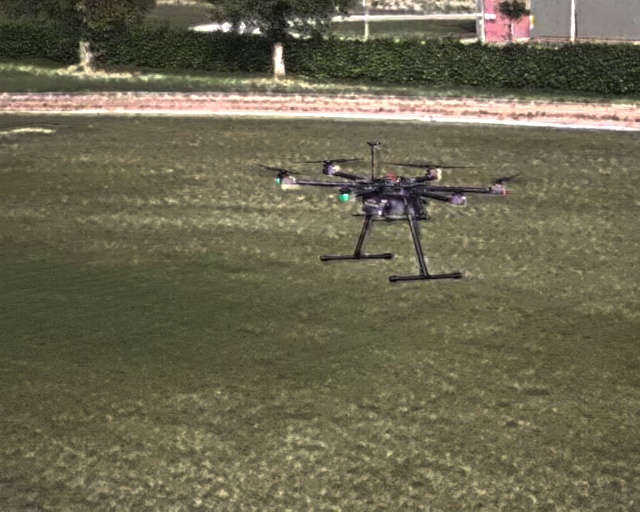}
		\caption{FCDFusion}
		\label{fig:fcd}
	\end{minipage}
    \hfill
    \begin{minipage}[b]{0.2\textwidth}
		\centering
		\includegraphics[width=\linewidth]{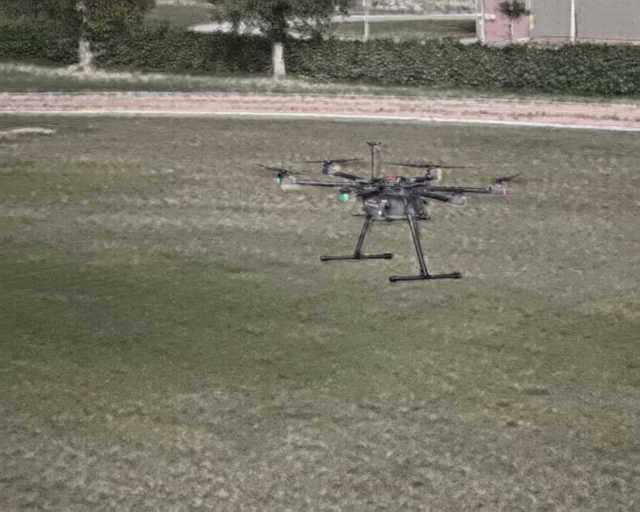}
		\caption{DenseFuse}
		\label{fig:densefuse}
	\end{minipage}
    \hfill
    \begin{minipage}[b]{0.2\textwidth}
		\centering
		\includegraphics[width=\linewidth]{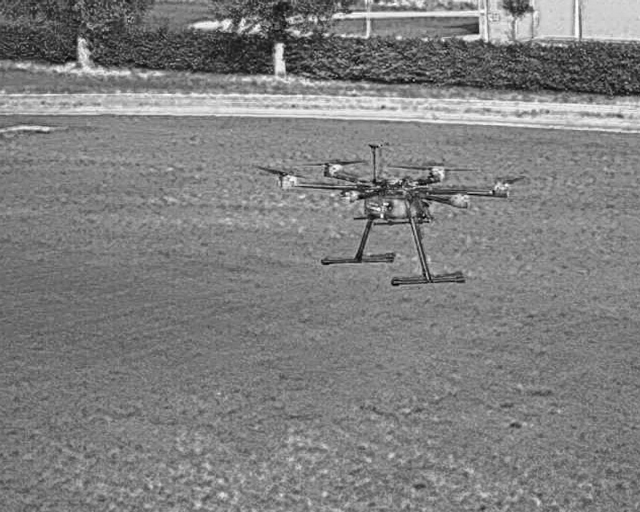}
		\caption{IFCNN}
		\label{fig:ifcnn}
	\end{minipage}
    \caption{Visible comparison between state-of-the-art fusion strategies, exploiting IAN framework and convolutional neural networks for multimodal image generation.}
\end{figure}

\begin{figure}[!ht]
	\centering
	\includegraphics[width=0.8\linewidth]{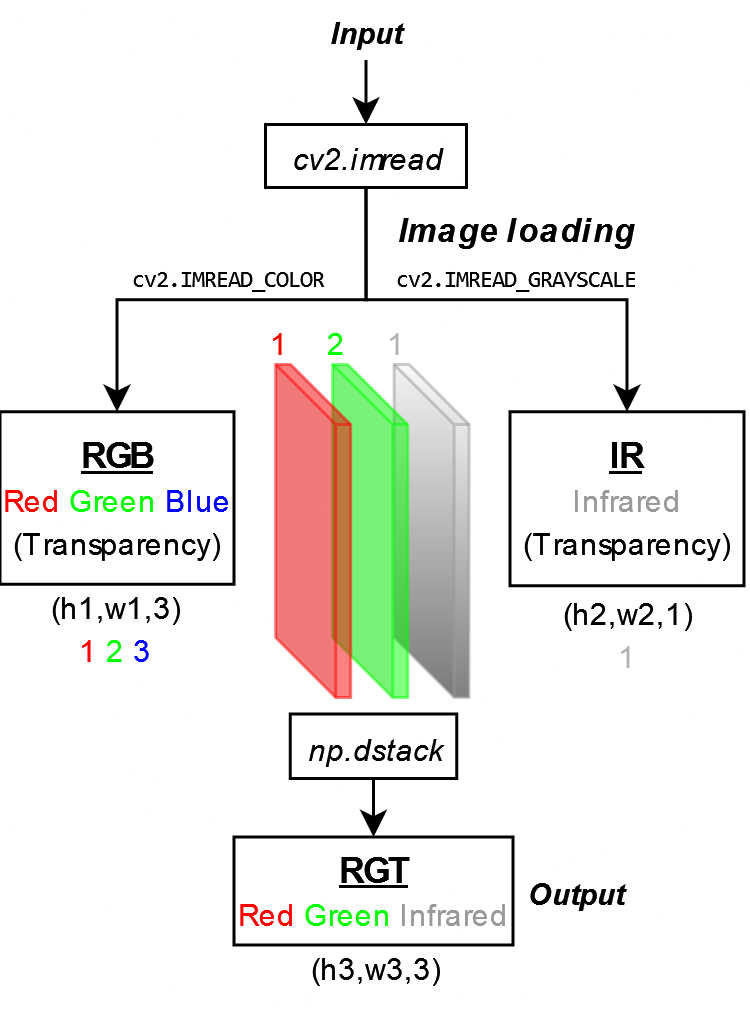}
	\caption{Early Fusion framework.}
	\label{fig:efusion}
\end{figure}

\begin{figure}[!ht]
	\centering
	\includegraphics[width=\linewidth]{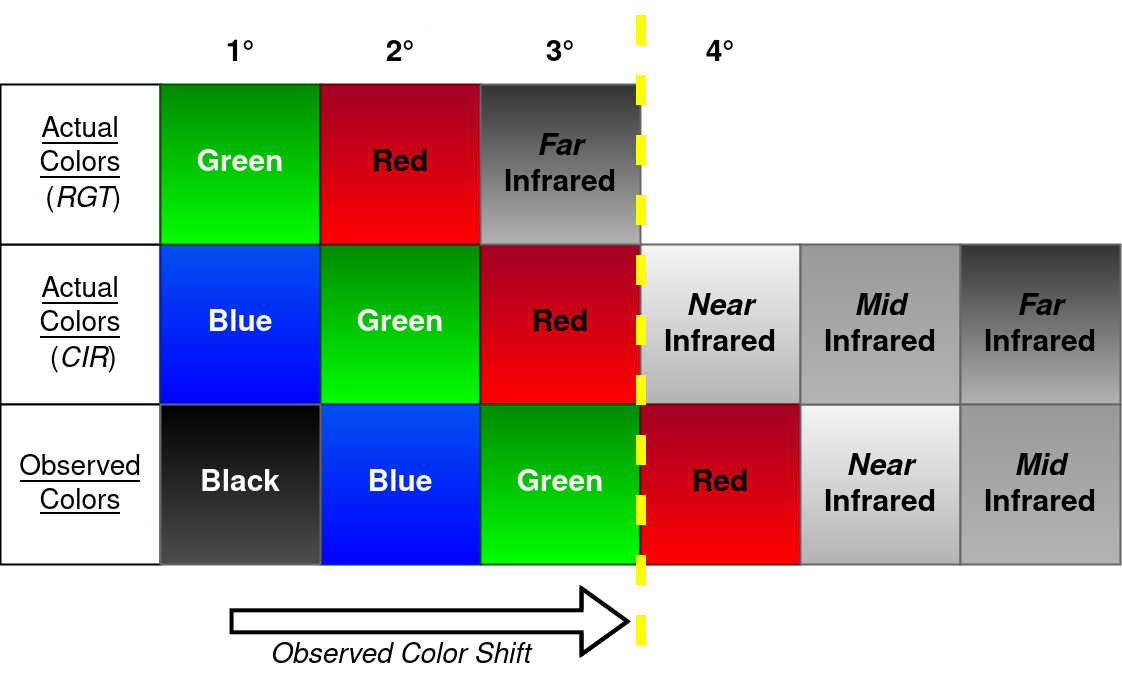}
	\caption{Channel shifting process. As it can be seen, there is a clear difference between CIR and RGT representations. Color-infrared imagery uses a portion of the electromagnetic spectrum known as near infrared, which ranges from $0.70 \mu m$ to $1.0 \mu m$. RGT uses instead far infrared. In both cases, blue wavelengths are shifted out of the visible portion of the spectrum. However, only 3 channels are considered part of the RGT representation, without considering the non-visible part of the spectrum.}
	\label{fig:shifting}
\end{figure}

\begin{figure}[!ht]
\centering

\begin{subfigure}{0.12\textwidth}
    \includegraphics[width=\linewidth]{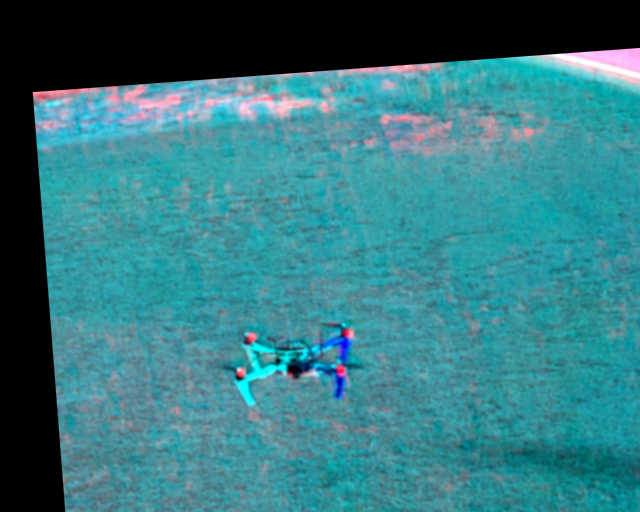}
\end{subfigure}%
\begin{subfigure}{0.12\textwidth}
    \includegraphics[width=\linewidth]{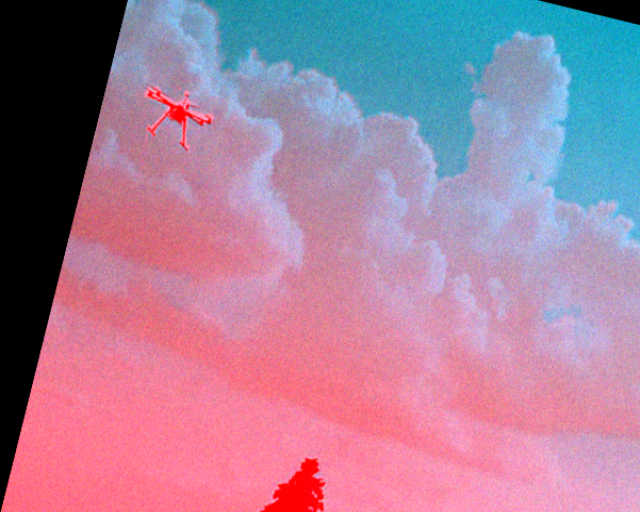}
\end{subfigure}%
\begin{subfigure}{0.12\textwidth}
    \includegraphics[width=\linewidth]{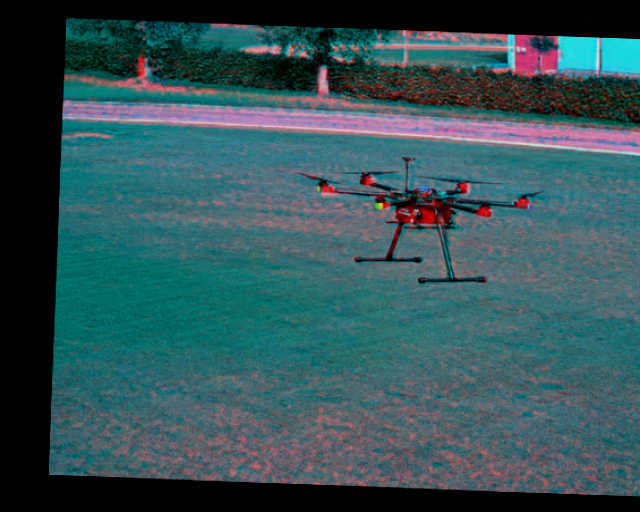}
\end{subfigure}%
\begin{subfigure}{0.12\textwidth}
    \includegraphics[width=\linewidth]{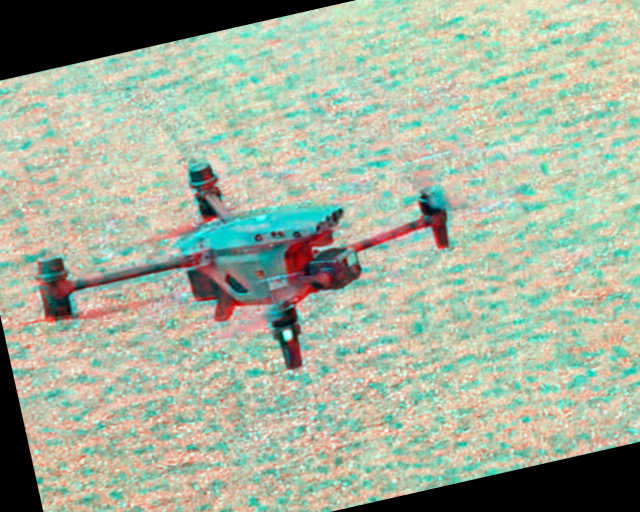}
\end{subfigure}%

\begin{subfigure}{0.12\textwidth}
    \includegraphics[width=\linewidth]{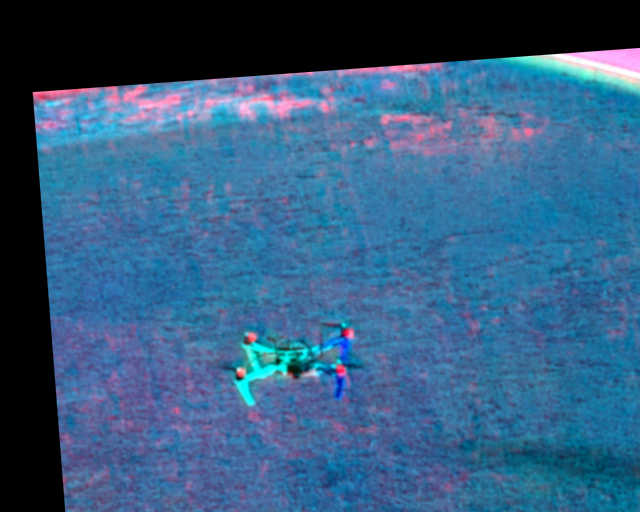}
\end{subfigure}%
\begin{subfigure}{0.12\textwidth}
    \includegraphics[width=\linewidth]{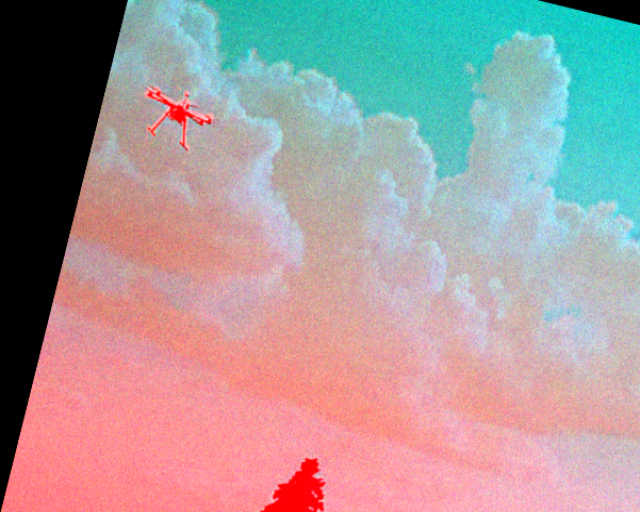}
\end{subfigure}%
\begin{subfigure}{0.12\textwidth}
    \includegraphics[width=\linewidth]{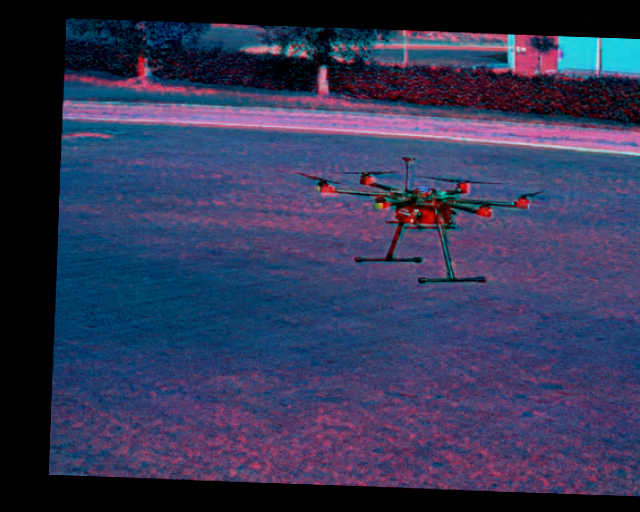}
\end{subfigure}%
\begin{subfigure}{0.12\textwidth}
    \includegraphics[width=\linewidth]{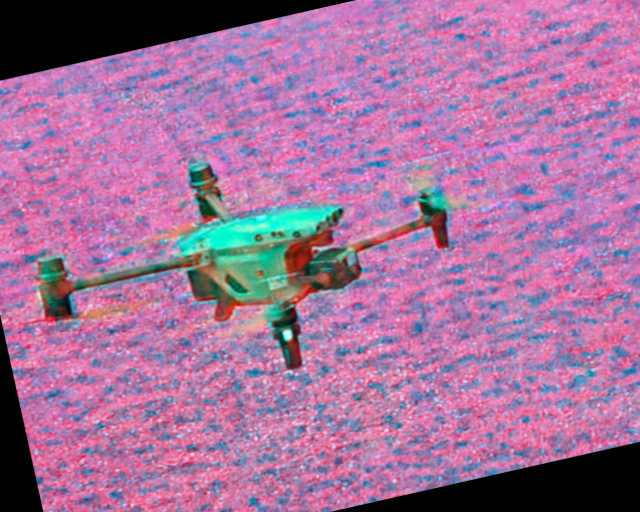}
\end{subfigure}%

\begin{subfigure}{0.12\textwidth}
    \includegraphics[width=\linewidth]{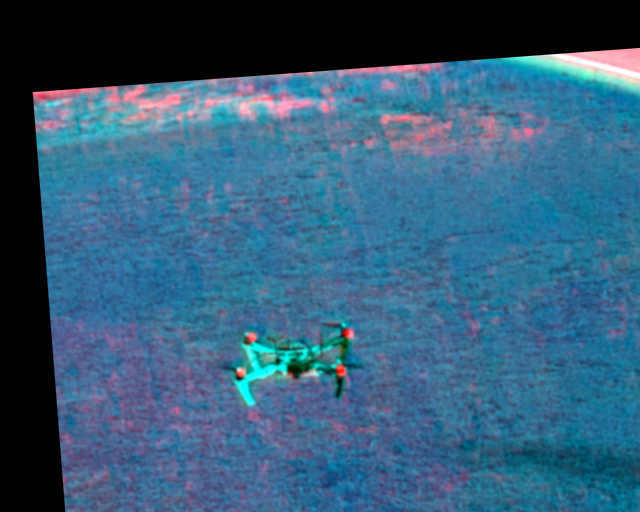}
\end{subfigure}%
\begin{subfigure}{0.12\textwidth}
    \includegraphics[width=\linewidth]{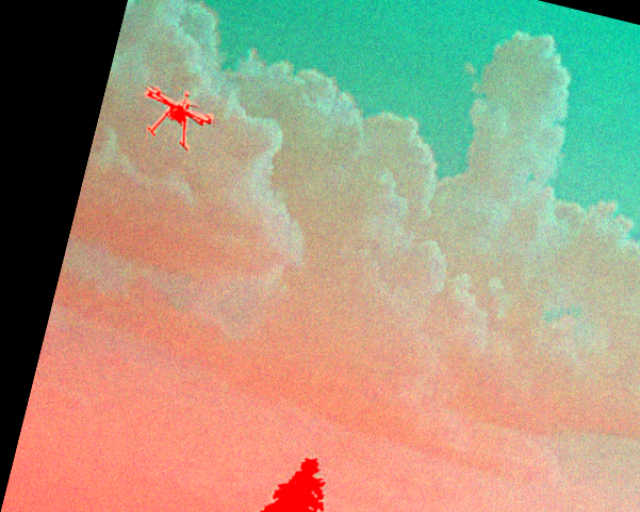}
\end{subfigure}%
\begin{subfigure}{0.12\textwidth}
    \includegraphics[width=\linewidth]{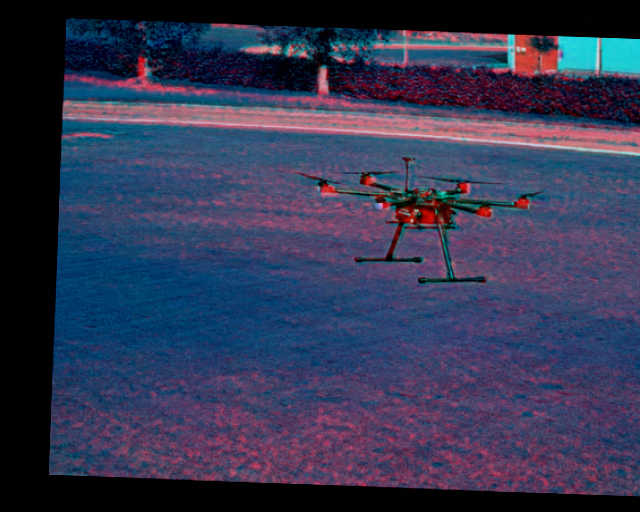}
\end{subfigure}%
\begin{subfigure}{0.12\textwidth}
    \includegraphics[width=\linewidth]{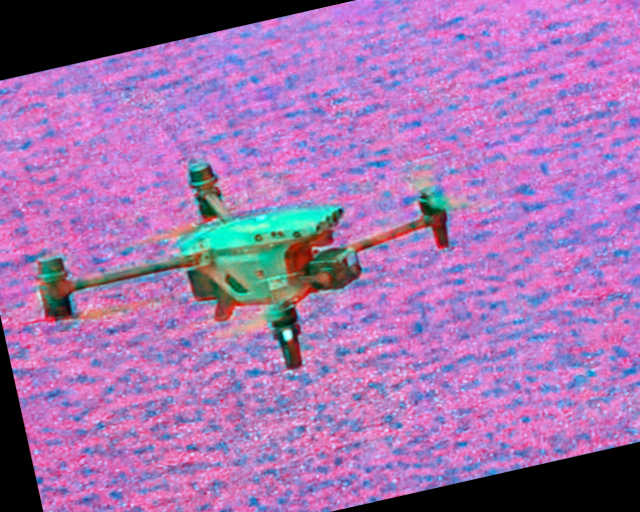}
\end{subfigure}

\caption{From the top: RGT, RBT and GBT fusion strategies. In all the cases only 3 of the 4 channels are considered for creating the fusions. For RGT the blue channel has been omitted in favour of the thermal one. A similar type of image is usually called CIR or false-color image \cite{usgs_cir}. The other types have only been considered for a technical comparison.}
\label{fig:3_layers}
\end{figure}

The early fusion approach combines RGB and IR images at the pixel level to create a single multimodal input. This approach requires frames to be aligned. This process was performed by applying transformation steps to each image, such as interpolation, perspective distortion and rotation. After this, an overlap of the channels was performed, while removing the transparency level (also called "A" in RGBA image format), which was present in \verb|.png| images. However, since the focus of this research strictly concerns the analysis of fusion strategies, the reader is referred to the paper on autonomous UAV detection architecture development \cite{architecture}, in which all alignment steps are described.\\
Once spatial alignment between RGB and infrared frames was achieved, the fusion was performed through overlap, by stacking the image channels into a single array of 4 spectra. The RGB image was split into three channels: Red, Green and Blue. Each of them is represented by a two-dimensional intensity map, where every pixel contains intensity values for that particular channel. A visual interpretation is shown in Figure \ref{fig:intensity_maps}.
On the other hand, the IR image was converted into a grayscale single-channel image. These channels were then combined into a 4-channel image (RGBT), where the fourth channel corresponded to the thermal information. The operation result is shown in Figure \ref{fig:4_layers}.\\
For implementation simplicity, the Python Imaging Library (PIL) was employed, as it allows independent access to individual image channels. Initial experiments involved the creation of four-channel RGBT images obtained by stacking the red, green, blue, and thermal components into a single tensor.
In addition to direct channel stacking, pretrained fusion models such as PIAFusion in Figure \ref{fig:pia}, FCDFusion in Figure \ref{fig:fcd} and DenseFuse in Figure \ref{fig:densefuse} were evaluated to assess alternative image-level fusion strategies. These approaches generated fused outputs without requiring manual channel manipulation.
However, Ultralytics introduced support for multi-channel training only recently \cite{yolov8_multispectral_2025}, subsequent to the experiments reported here. Hence, although OpenCV2 library, commonly used within YOLO pipelines, allows image loading with an arbitrary number of channels using the \verb|cv2.IMREAD_UNCHANGED| flag, in previous versions of Ultralytics, in order to use 4-channel inputs, extensive modifications to the YOLO architecture would be required, including adaptation of the model's input layer and associated preprocessing routines.
In order to preserve architectural consistency and ensure fair comparison with baseline experiments, modifications to the core framework and alteration to the network structure were avoided. \\
Instead, the characteristics of RGB channels were taken into consideration. RGB usually exhibits a high degree of spectral correlation, since most objects in the environment reflect light similarly across the visible spectrum. Both limitations and contributions of each channel have been summarized in Table \ref{tab:rgb_channels}.
\begin{table}[!ht]
\caption{Summary of the characteristics of the RGB channels \cite{5350902}.}
\label{tab:rgb_channels}
\centering
\footnotesize
\renewcommand{\arraystretch}{1.2} 
\begin{tabularx}{\columnwidth}{>{\bfseries}l X X}
\toprule
Channel & Main contribution & Limitations / Remarks \\
\midrule
Green &
Contributes the most to the perception of luminance and fine details. &
Provides high structural information. \\
Red &
Crucial for discriminating between minerals and soils. &
Complementary for material recognition. \\
Blue &
Captures wavelengths in the 450--495\,nm range, useful for identifying water bodies and atmospheric effects. &
Lower contrast, limited thermal and low-light contributions, sensitive to Rayleigh scattering, noisier in long-distance imagery. \\
\bottomrule
\end{tabularx}
\renewcommand{\arraystretch}{1}
\end{table}
Due to these highlights reported, a modified 3-channel tensor was created by discarding the blue channel, which turned out to be redundant in several outdoor UAV scenes where red and green spectral content are often predominant. The blue channel was then replaced with the thermal component, mitigating some of the atmospheric noise. This resulted in an RGT image (Red, Green, Thermal), maintaining compatibility with YOLOv8 architecture while integrating complementary spectral information. Consequently, a 3-channel fusion representation was adopted. Therefore, the same channel-stacking principles used in the 4-channel implementation were applied to construct the 3-channel fused images. The methodological workflow is summarized in the dedicated flowchart in Figure \ref{fig:efusion}.
The process shown in Figure \ref{fig:shifting} can be summarized as a shift on the left of all the image channels, in order to include the thermal one inside the final image.
Therefore, the resulting fused image consisted of red, green and thermal channels, forming an RGT representation. Some examples are shown in Figure \ref{fig:3_layers}. The distinction between a CIR image \cite{mngeo_cir_imagery} and the proposed RGT one relies on the infrared type chosen. In fact, the camera sensor used for the tests operates in far-infrared (FIR) range, whereas CIR relies on near-infrared (NIR) imagery. \\
This design enabled YOLOv8s model training without architectural modification and allowed operation with the traditional YOLO detection pipeline while reducing potentially redundant spectral information. \\
Formally, each spectral channel is defined as a function over pixel coordinates $(i, j)$:
\begin{align}
    R(i,j) &:\ \text{Red intensity at pixel } (i,j) \\
    G(i,j) &:\ \text{Green intensity at pixel } (i,j) \\
    T(i,j) &:\ \text{Thermal intensity at pixel } (i,j)
\end{align}
The fused image tensor is obtained by stacking these channels:
\begin{equation}
	I_{RGT}(i, j) =
	\begin{bmatrix}
		R(i, j) \\
		G(i, j) \\
		T(i, j)
	\end{bmatrix}
\end{equation}
Each pixel location $(i, j) \in \mathbb{Z}^2$ is therefore mapped to a three-dimensional spectral vector in $\mathbb{R}^3$:
\begin{equation}
	I_{RGT} : \mathbb{Z}^2 \rightarrow \mathbb{R}^3,
	\quad (i, j) \mapsto (R(i, j), G(i, j), T(i, j)).
\end{equation}
The complete fused image can thus be expressed as a tensor:
\begin{equation}
	I_{RGT} \in \mathbb{R}^{H \times W \times 3},
\end{equation}
where $H$ and $W$ denote image height and width, respectively.\\
The same pipeline was also adopted for the other two channel combinations, respectively RBT and GBT, each obtained by omitting one of the three RGB channels. This enabled an ablation study on channel selection, providing a formal justification for the choice of RGT in the proposed testing scenario.

After generating and augmenting a substantial number of fused images, the YOLOv8s detector was trained using the standard (visible and infrared) and the fused datasets. All of them contain a total of 300 images for the training part and 100 for the validation one, corresponding to a split of 75\% and 25\% respectively. The results obtained after training and validation processes were compared against models trained exclusively on RGB and IR data.\\
These findings can be interpreted in light of the Data Fusion Principle\footnote{The integration of multiple information sources should yield performance superior to that of each individual modality considered separately.}. Accordingly, a fusion-based system is expected to outperform its unimodal counterparts if complementary information is effectively exploited.\\
The comparative performance metrics obtained on unimodal and multimodal datasets are summarized in Table \ref{tab:metrics}.
\begin{table}[!ht]
    \scriptsize
	\centering
	\caption{Performance comparison of YOLOv8s model (11.2M parameters, 28.6B FLOPs) over standard spectra (RGB, IR), literature fusions (FCD, PIA, Dense, IFCNN) and proposed early fusion approaches (RGT, RBT, GBT). The model has been trained for 300 epochs with \texttt{imgsz=640} over the same dataset, only changing the color channels and the fusion strategies for the images inside it.}
	\begin{tabular}{lccccc}
		\hline
		Dataset & Precision & Recall & mAP@50 & mAP@50-95 & F1-Score \\
		\hline
		\textbf{RGT} & \textbf{95.9\%} & \textbf{94.6\%} & \textbf{97.8\%} & \textbf{56.2\%} & \textbf{95.2\%}\\
        \textbf{RBT} & \textbf{96.9\%} & \textbf{93.8\%} & \textbf{96.9\%} & \textbf{55.6\%} & \textbf{95.3\%} \\
        \textbf{GBT} & \textbf{93.2\%} & \textbf{85\%} & \textbf{92.4\%} & \textbf{53.3\%} & \textbf{88.9\%} \\
        FCDFusion\cite{li2024fcdfusion} & 98.9\% & 91.8\% & 96.8\% & 56.9\% & 95.2\% \\
        PIAFusion\cite{tang2022piafusion} & 96.2\% & 76.4\% & 84.6\% & 42.2\% & 85.2\% \\
        DenseFuse\cite{li2019densefuse} & 95.5\% & 84.7\% & 91.5\% & 50.3\% & 89.8\% \\
        IFCNN\cite{zhang2020IFCNN} & 97.8\% & 90.7\% & 93.8\% & 50.4\% & 94.8\% \\
        IR & 89\% & 80.8\% & 89.8\% & 51.5\% & 84.7\% \\
        RGB & 97.9\% & 91.7\% & 97\% & 48.8\% & 94.7\%\\
		\hline
	\end{tabular}
	\label{tab:metrics}
\end{table}
While the standard RGB configuration achieved higher precision, the overall detection performance did not uniformly surpass the fused models across all evaluation metrics. In fact, the superiority of the three proposed fusion strategies first emerges on mAP@50-95. The highest one is reached by RGT with 56.2\%. A similar improvement can be observed on mAP@50, where the proposed fused approach achieves a result of 97.8\%, compared to 89.8\% of Infrared. These are indexes of good model performance, since mAP evaluates it over all confidence thresholds, rather than at a single operating point.\\
Another important point regards F1-Score, which penalizes models with unbalanced precision and recall values. \\
All these results highlight the improvements of RGT fusion in comparison to the other strategies reported. In fact, the metrics suggest that the proposed multimodal approach has gained the capability to enrich the scope and improve the stability of the model in object detection tasks. In addition, the comparison between fusion strategies enriches the study, highlighting the importance of a dedicated fusion approach with respect to the use-case scenario. \\
Another interesting point regards the results obtained from literature fusion strategies. All of them cannot be considered for real-time purposes, since they rely on dedicated fusion networks (illumination-aware or autoencoder-based) whose computational cost is unsuitable for real-time processing. Thus, even though the FCD approach gives promising results, it is not usable in real-time scenarios.\\
According to the results obtained, the RGT fusion strategy was also better than the other two presented (RBT and GBT). For this reason, it was selected for tests in real-world scenarios. Nevertheless, it should be pointed out that this is a preliminary evaluation step for the topic. It could surely be enriched and extended with a larger fused dataset and other object detection models, in order to carry on the research in the field.\\
A further consideration can be made by comparing the proposed RGT strategy with FCDFusion, which turned to be the two best models. Although FCDFusion outperforms RGT in precision and mAP@50-95, RGT achieves higher recall and mAP@50, while both methods obtain the same F1-Score of 95.2\%, indicating comparable overall detection performance. Therefore, the superiority of RGT should not be interpreted exclusively in terms of individual detection metrics, but rather considering the relationship between performance and computational complexity. In fact, FCDFusion requires a dedicated fusion network to generate the fused representation, introducing an additional deep learning model into the processing pipeline. This results in additional computational requirements and increases the overall complexity of the system. Conversely, RGT performs the fusion directly at the frame level, without requiring a dedicated fusion network during inference. The multimodal information can therefore be combined in real time and directly provided to the object detector, which represents a significant advantage for real-world applications.

\subsection{Late Fusion}
In the late fusion approach aligned RGB and IR images are processed separately by two object detection models. The resulting bounding boxes are then fused using a weighted average method. For each pair of boxes, the center coordinates are averaged based on their confidence scores, while the maximum width and height are kept to avoid box shrinking. A distance threshold $\delta$ is used to decide whether two boxes should be fused. In particular, if the centers of two boxes are within this range, they are fused, otherwise they are left as they are. More in detail, the Euclidean distance between the center points of two bounding boxes in the two spectra is computed. If this distance falls in the threshold set, weighted average is performed. Otherwise, the original box created by the model is preserved. This approach has been designed to avoid discarding detections that may only exist in one modality and to work in case of misaligned frames. In fact, if the centers of the two bounding boxes are near, the system fuses them. Alternatively, it preserves what each model has decided during detection. \\
In the decision-level fusion strategy, two spatially aligned images, one RGB and its corresponding infrared (IR) counterpart, are processed independently by their respective detection models. Each detector produces a set of bounding boxes with associated confidence scores. Fusion is subsequently performed at the bounding box level using a weighted averaging method.
\begin{figure}[!ht]
	\centering
	\includegraphics[width=\linewidth]{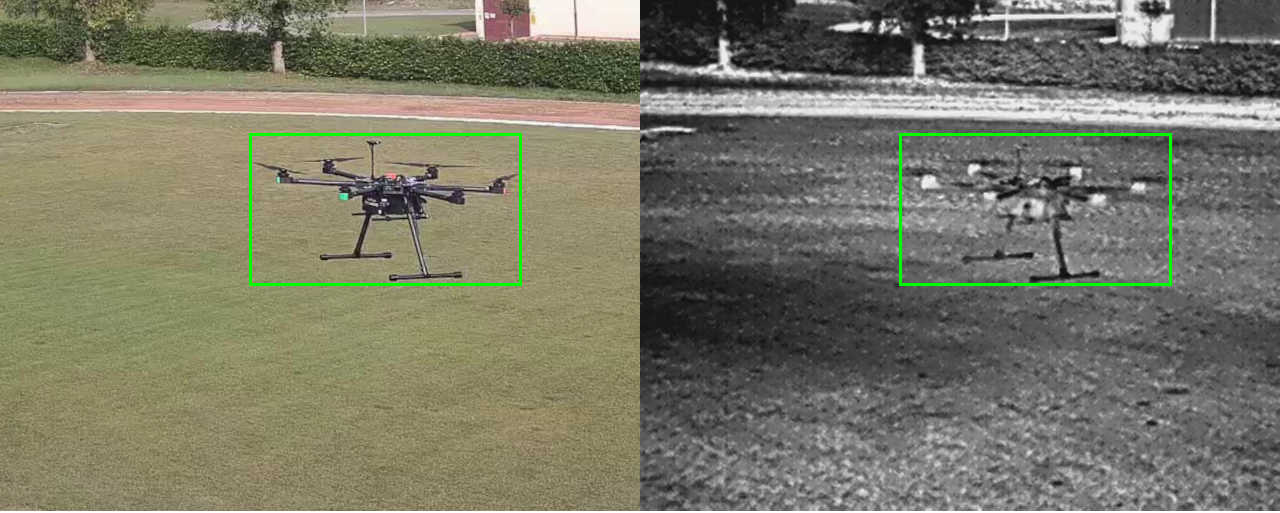}
	\caption{The bounding box is unique for both images. As it was seen in the experiments, it could work even with misaligned frames.}
	\label{fig:late_fusion}
\end{figure}
The weighted average approach computes the fused bounding box coordinates based on the confidence scores of the individual detections. This method can be applied to both perfectly aligned and moderately misaligned image pairs. However, if the same object appears with slight positional or orientational discrepancies across the two modalities, the resulting fused bounding box may exhibit reduced localization precision. In such cases, the fused bounding box may become enlarged, encompassing the object regions identified in both modalities within a single unified box.\\
The fusion procedure begins with the computation of the center coordinates of the new bounding box using confidence-weighted averaging. The width and height of the fused bounding box are then defined as the maximum of the corresponding dimensions from the RGB and IR detections. This design choice prevents the fused bounding box from shrinking excessively and ensures that the spatial extent of the detected object is fully preserved.\\
A dedicated fusion function was implemented to manage all detection scenarios. Since decision-level fusion requires detections from both modalities, situations may arise in which a UAV is detected by only one of the two models. In such cases, the bounding boxes generated by the detecting model are retained without modification.
\begin{figure}[!ht]
	\centering
	\includegraphics[width=\linewidth]{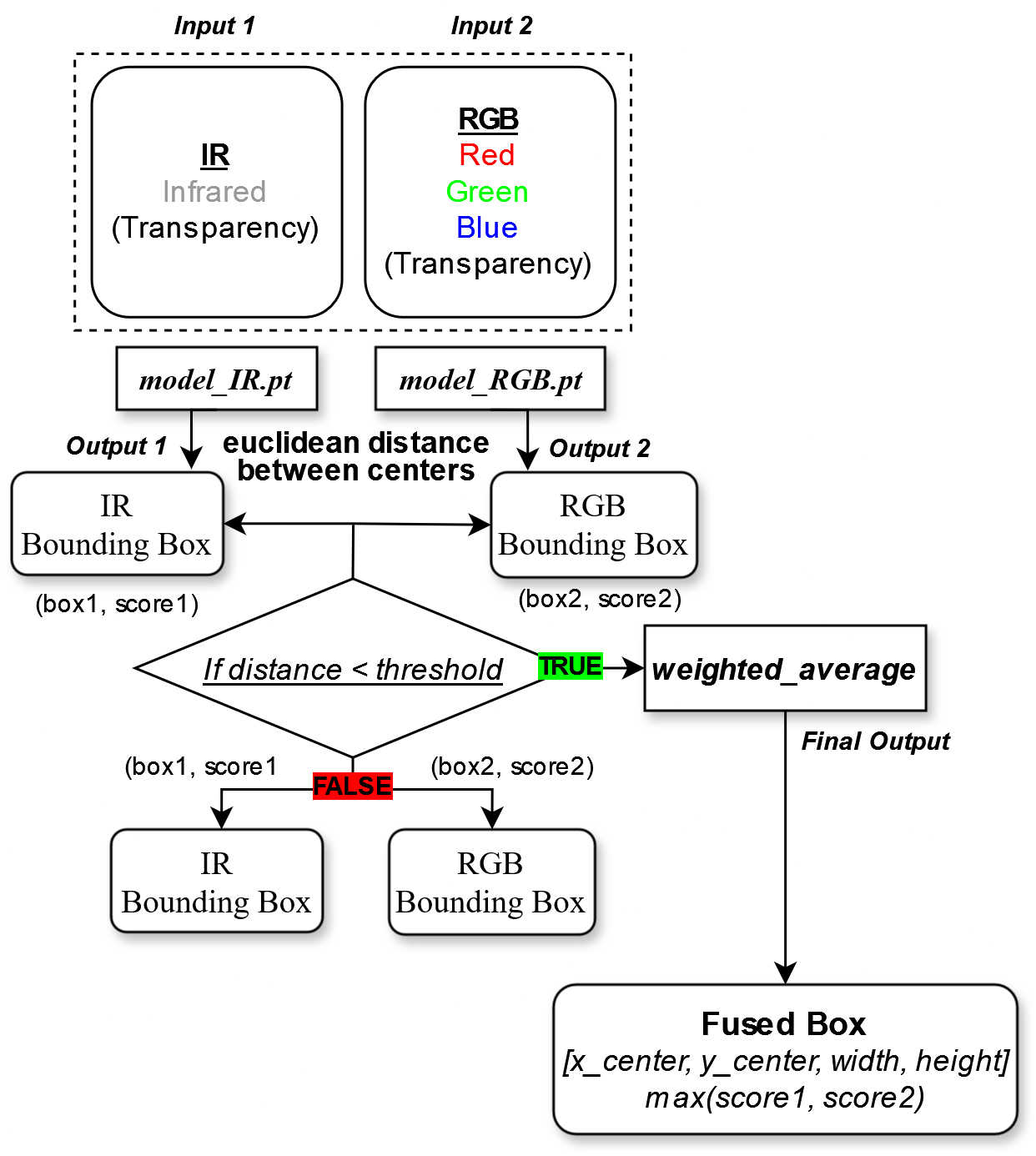}
	\caption{Late Fusion logic.}
	\label{fig:lfusion}
\end{figure}
In order to regulate the correspondence between detections from the two modalities, a fixed distance threshold ($\delta=50$ pixels) was introduced. This threshold is defined as a geometric correspondence tolerance for the spatial displacement that may occur between RGB and IR detections. Since the two optical axes are physically separated and therefore non-coaxial, the same UAV may appear at different image coordinates in the two modalities due to parallax and residual registration errors. The threshold consequently represents the maximum positional discrepancy considered compatible with the hypothesis that two detections correspond to the same physical target. A fixed pixel-domain threshold was adopted because the proposed fusion operates directly on detector outputs at a fixed input resolution of $640\times640$ pixels. The value of $\delta$ is therefore treated as a predefined correspondence tolerance rather than as an optimized detection hyperparameter, and is kept constant throughout the evaluation. In particular, the use of a distance tolerance allows the proposed method to tolerate spatial discrepancies between the two modalities while preventing detections with larger spatial separation from being merged.\\
For each pair of RGB and IR bounding boxes, the Euclidean distance between their center coordinates has also been computed. If this distance is below the threshold, weighted averaging is applied. Otherwise, the bounding boxes are preserved independently. This strategy enables the method to operate effectively under moderate sensor misalignment. When detections are spatially consistent across modalities, fusion is performed; when they differ significantly, each model's decision is maintained.\\
Formally, consider two bounding boxes
\[
B_1 = (x_1, y_1, w_1, h_1), \quad B_2 = (x_2, y_2, w_2, h_2),
\]
with associated confidence scores $s_1$ and $s_2$. The center coordinates of the fused bounding box are computed as confidence-weighted averages:
\begin{align}
	x_c = \frac{s_1 x_1 + s_2 x_2}{s_1 + s_2}\\
	y_c = \frac{s_1 y_1 + s_2 y_2}{s_1 + s_2}
\end{align}
The width $w$ and height $h$ of the fused bounding box are defined as the maximum between the values of the two original boxes.
The resulting fused bounding box is therefore represented as:
\begin{equation}\label{eq:fbb}
	B_f = (x_c, y_c, w, h)
\end{equation}
To determine whether fusion should be applied, the Euclidean distance between the centers $C_1 = (x_1, y_1)$ and $C_2 = (x_2, y_2)$ is computed as:
\begin{equation}
	D(C_1, C_2) = \sqrt{(x_1 - x_2)^2 + (y_1 - y_2)^2}
\end{equation}
Fusion is performed if:
\begin{equation}
	D(C_1, C_2) < \delta
\end{equation}
where $\delta$ is the empirical distance threshold. The final confidence score of the fused bounding box is the maximum of the two score values.
If the distance condition is not satisfied, the original bounding boxes are preserved. This formulation ensures robustness to moderate spatial discrepancies between modalities while maintaining reliable detection outputs in the presence of partial or asymmetric detections.

\section{Experimental Tests}\label{Sec:Experiments}
After computing the performance metrics of each model, the results gained from tests in real-world scenarios are analyzed and compared, both visually and empirically. The analysis will cover all the phases with both strategies: Early and Late Fusion.
\subsection{Acquisition Scenarios}
After recording all data in the two modalities, some results are presented as frames of the collected videos, discussing both benefits and limitations of the two systems in all the conditions tested.\\
The metric considered in external tests is the confidence score calculated by the model during detection. This value indicates the probability of an unidentified flying object belonging to the UAV class and it has allowed us to perform an early qualitative review of the system in different environmental conditions. In the final part of the section, some quantitative results are also presented in a dedicated table, in which other metrics are considered and analyzed, in order to give a review of the two frameworks as complete as possible.
\begin{figure}[!ht]
    \centering
    \includegraphics[width=\linewidth]{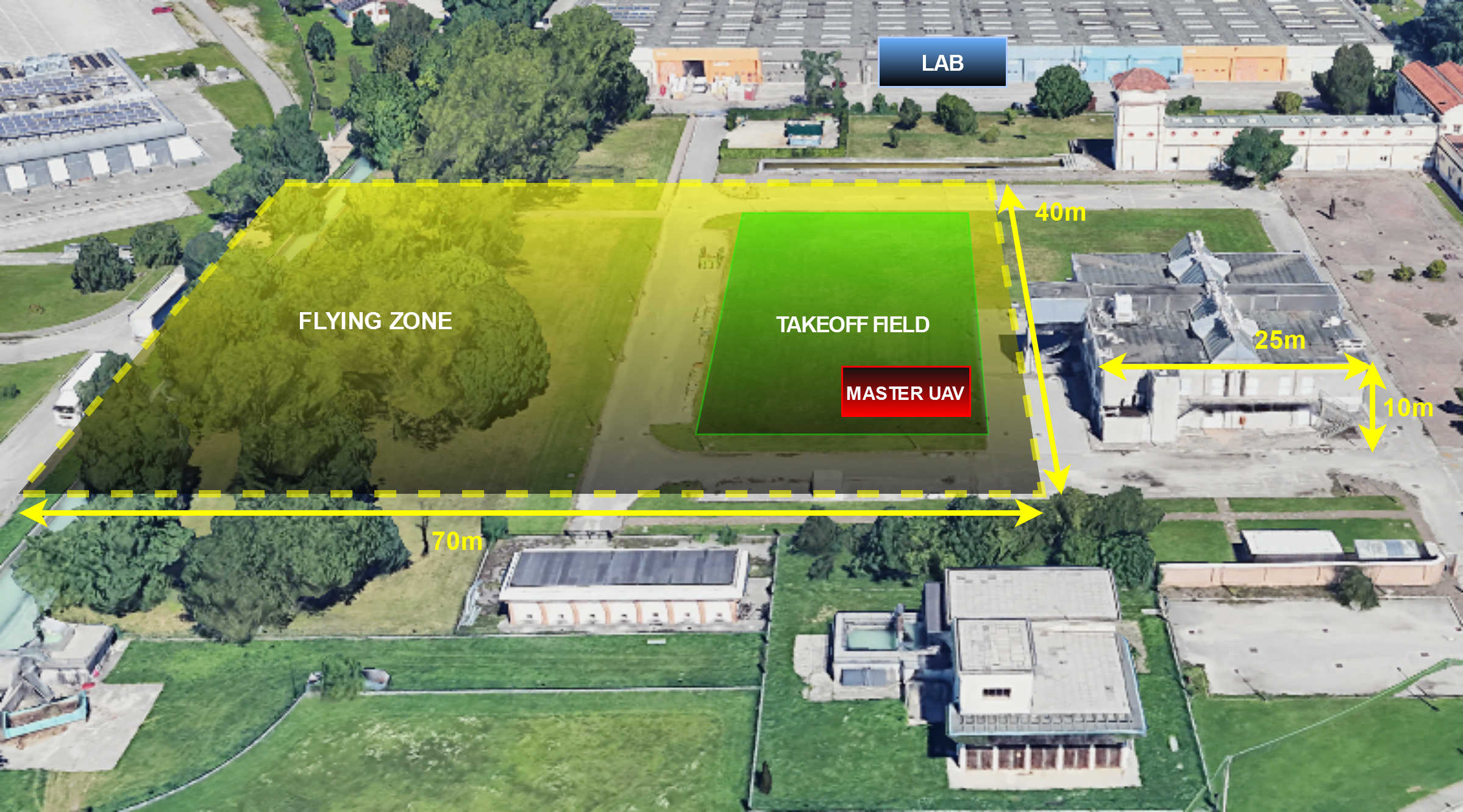}
    \caption{The outdoor testing field $\copyright$ Google Earth Pro. The takeoff field is highlighted in green. Inside it, in the red area, the master drone used for detection takes off. In the rest of the yellow scenario, a target drone flies at different altitudes, in order to test the capabilities of the detection models.}
    \label{fig:scenario}
\end{figure}
\begin{figure*}[!ht]
	\centering
	\begin{subfigure}{0.11\textwidth}
		\centering
		\includegraphics[width=\linewidth]{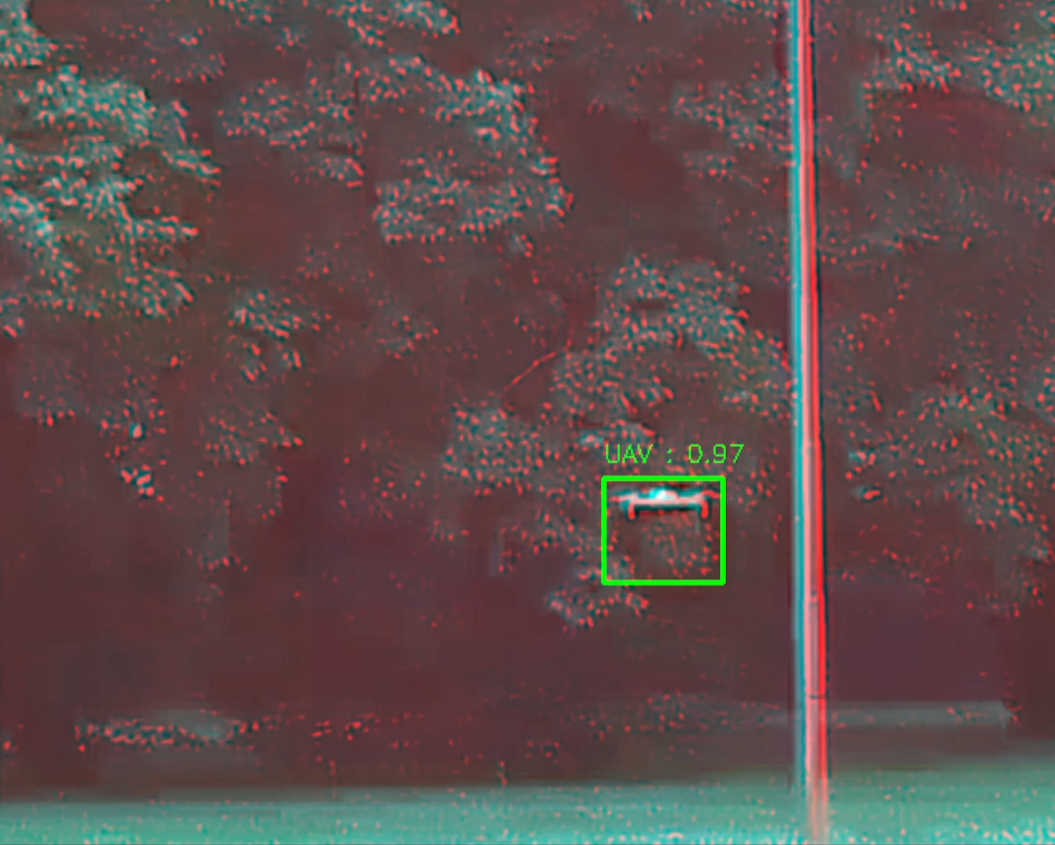}
	\end{subfigure}\hfill
	\begin{subfigure}{0.11\textwidth}
		\centering
		\includegraphics[width=\linewidth]{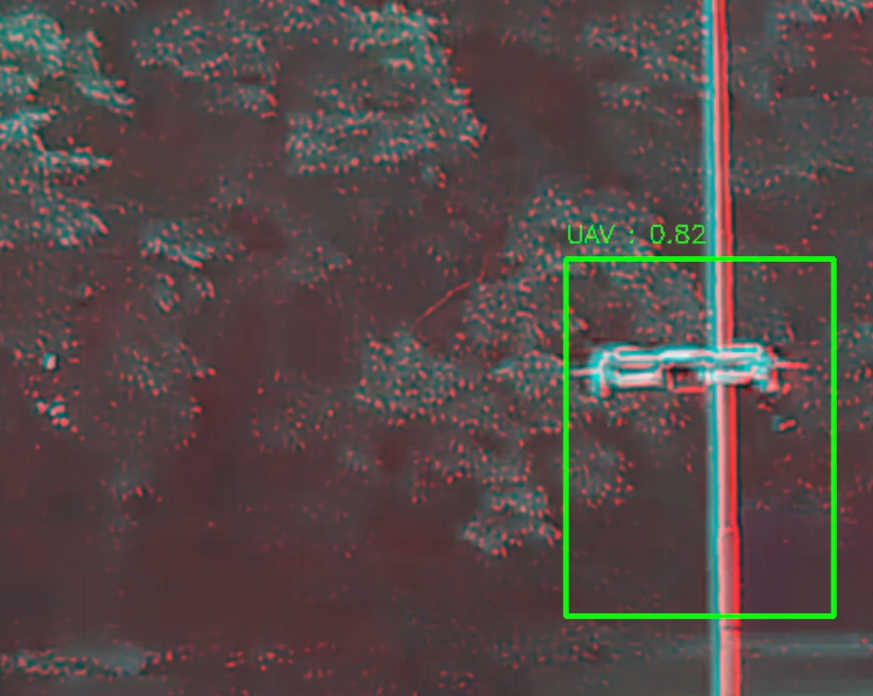}
	\end{subfigure}\hfill
	\begin{subfigure}{0.11\textwidth}
		\centering
		\includegraphics[width=\linewidth]{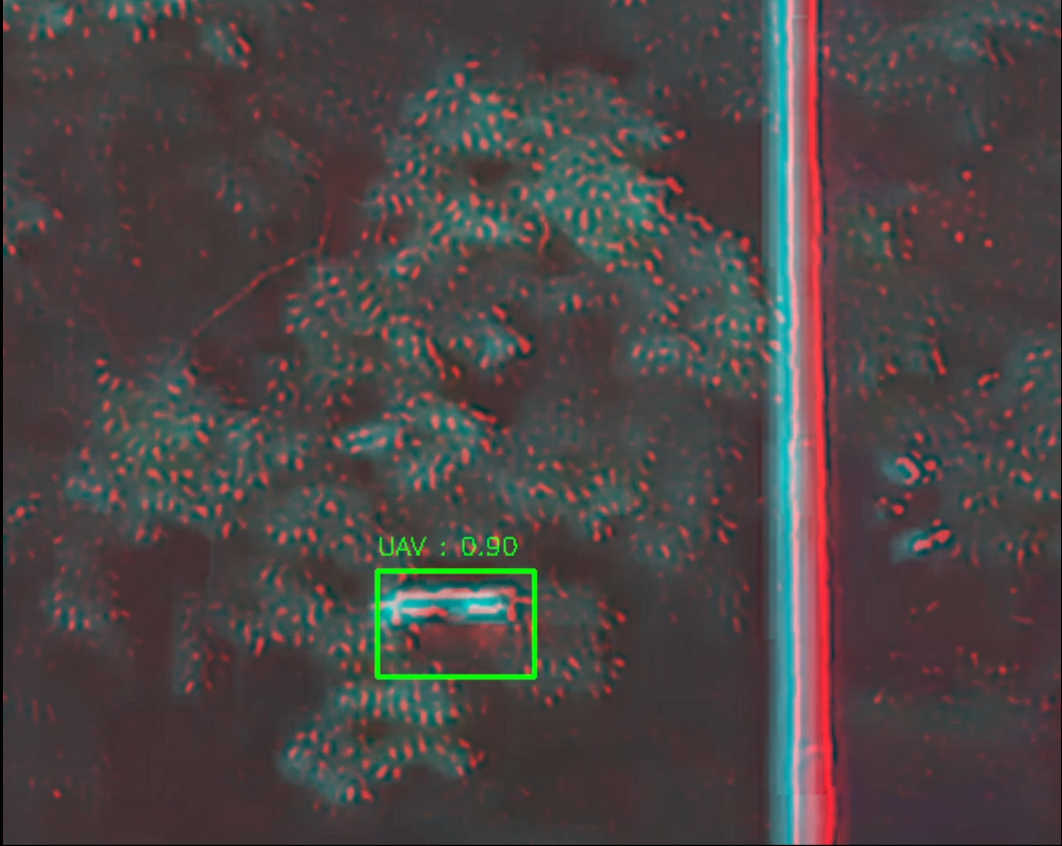}
	\end{subfigure}\hfill
	\begin{subfigure}{0.11\textwidth}
		\centering
		\includegraphics[width=\linewidth]{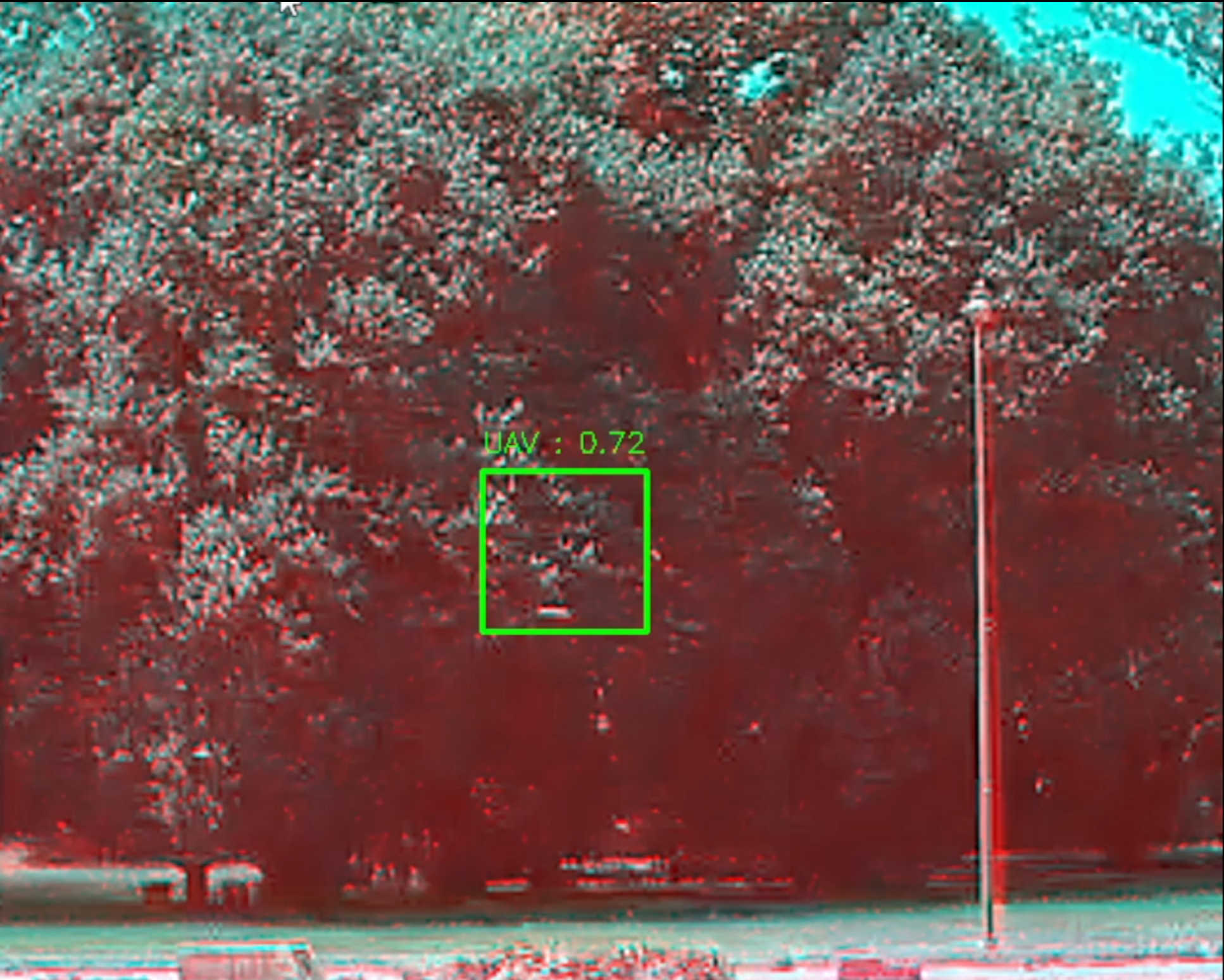}
	\end{subfigure}\hfill
	\begin{subfigure}{0.11\textwidth}
		\centering
		\includegraphics[width=\linewidth]{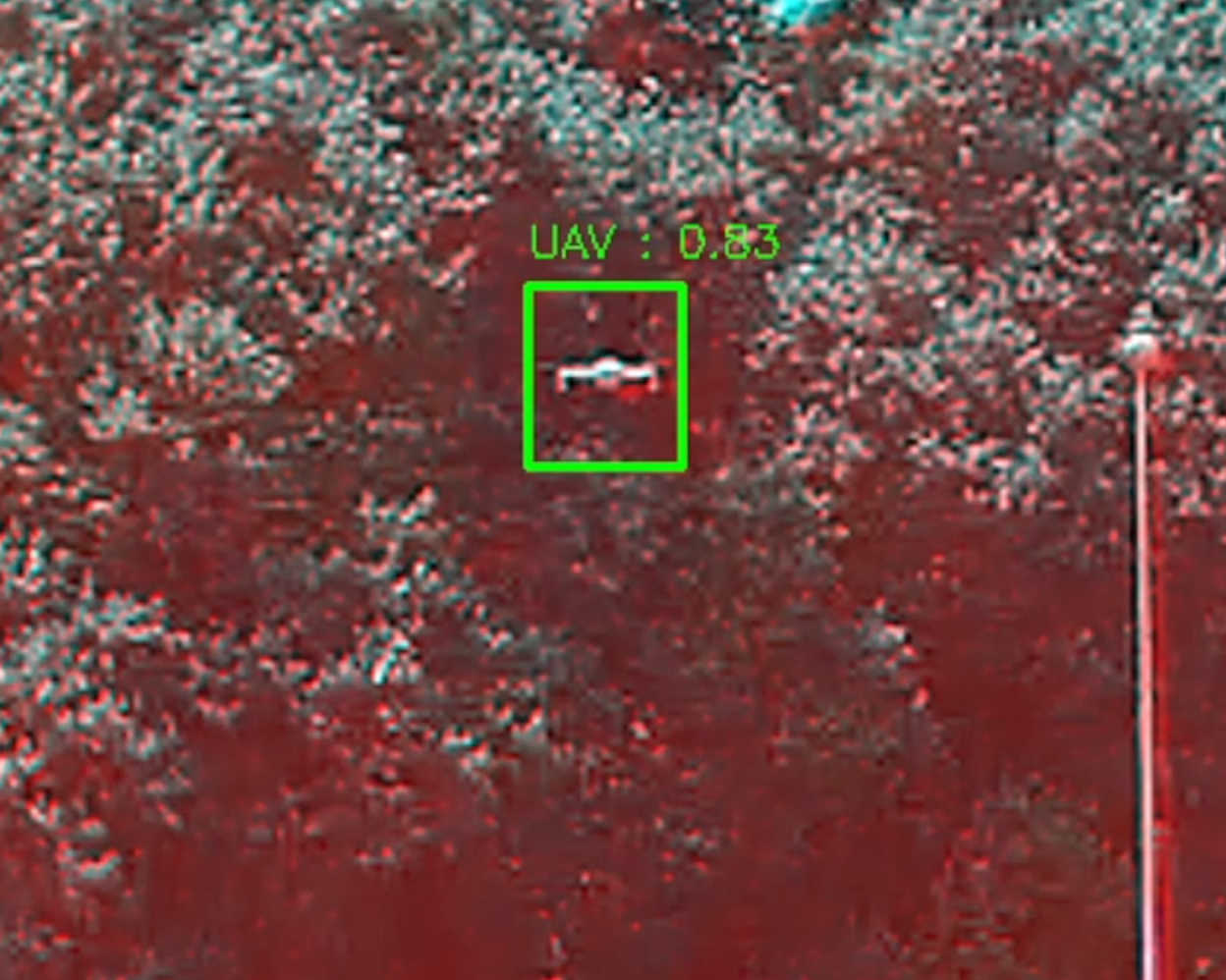}
	\end{subfigure}\hfill
	\begin{subfigure}{0.11\textwidth}
		\centering
		\includegraphics[width=\linewidth]{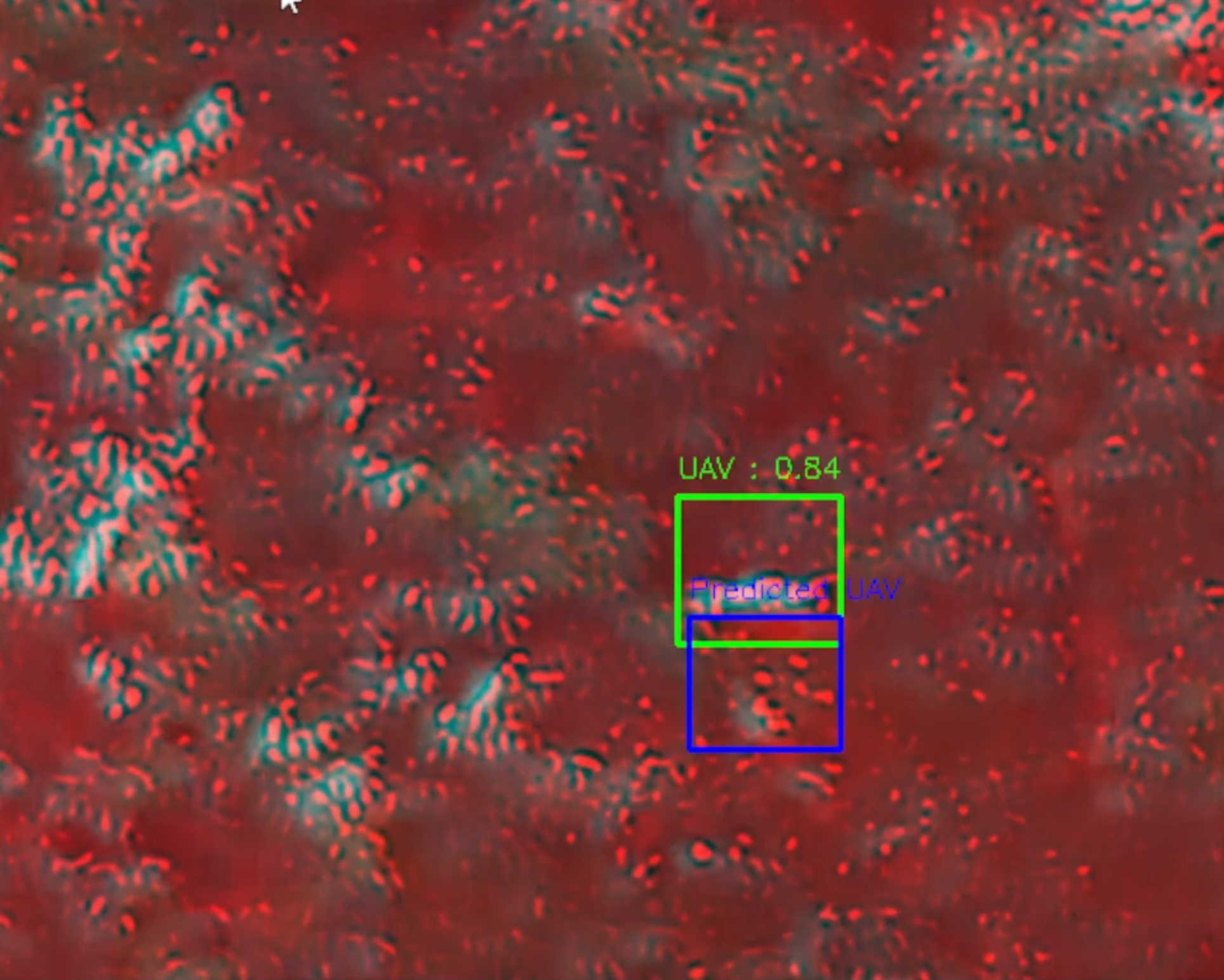}
	\end{subfigure}\hfill
	\begin{subfigure}{0.11\textwidth}
		\centering
		\includegraphics[width=\linewidth]{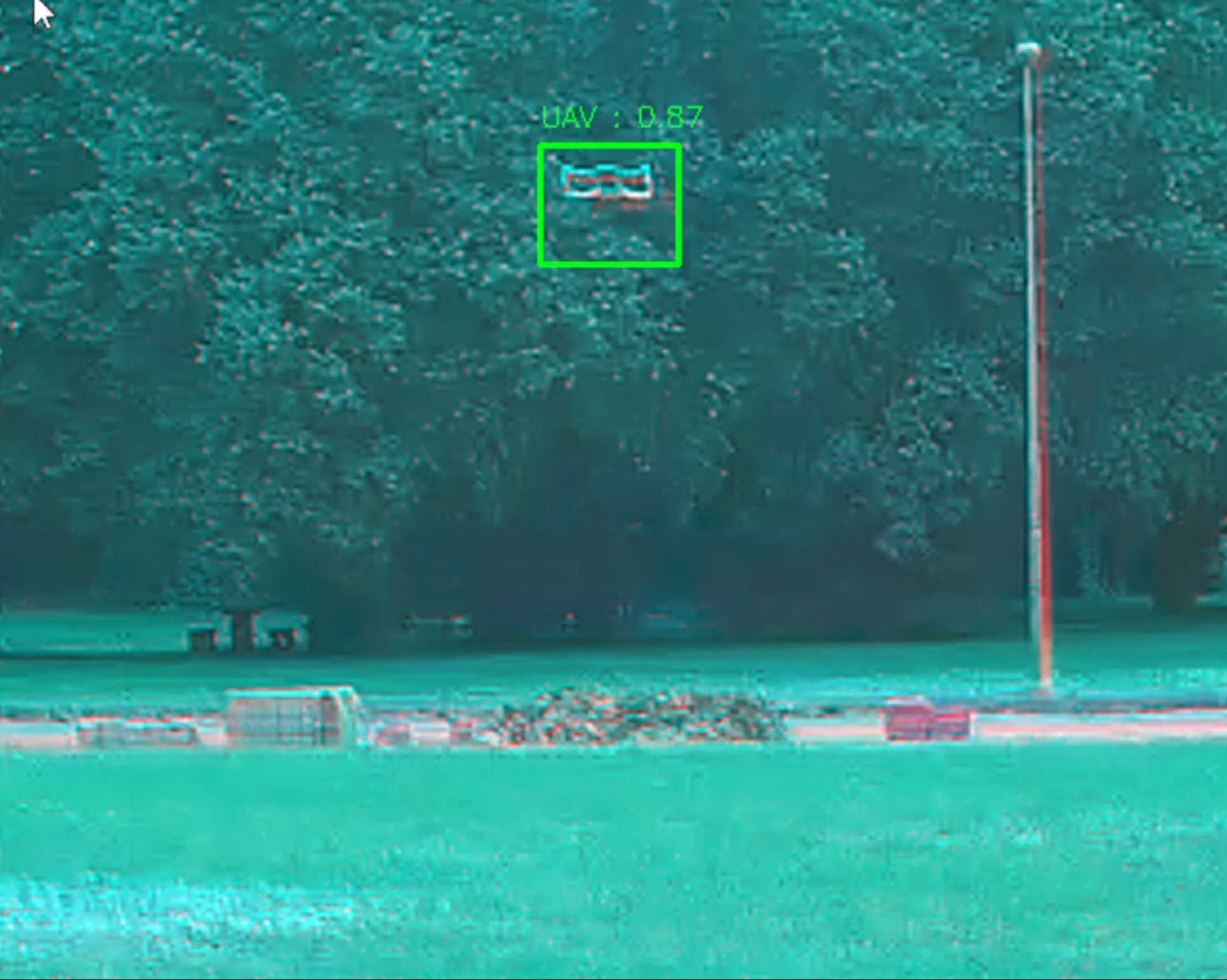}
	\end{subfigure}\hfill
	\begin{subfigure}{0.11\textwidth}
		\centering
		\includegraphics[width=\linewidth]{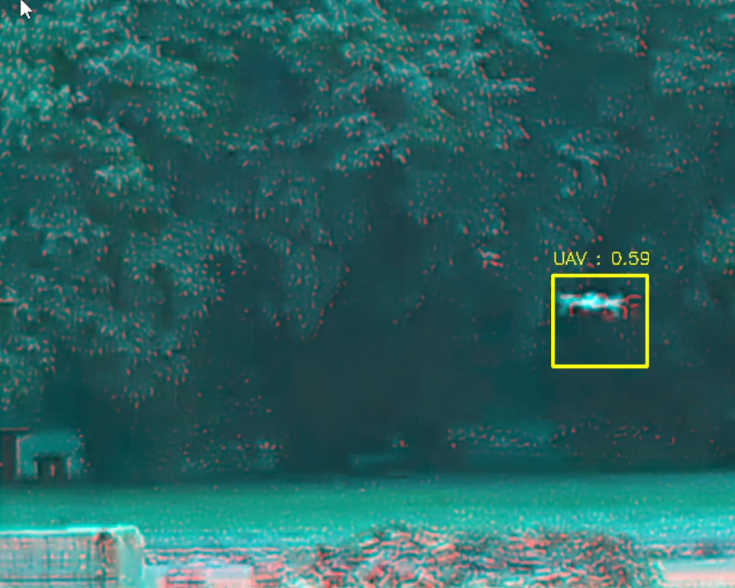}
	\end{subfigure}\hfill
	\begin{subfigure}{0.11\textwidth}
		\centering
		\includegraphics[width=\linewidth]{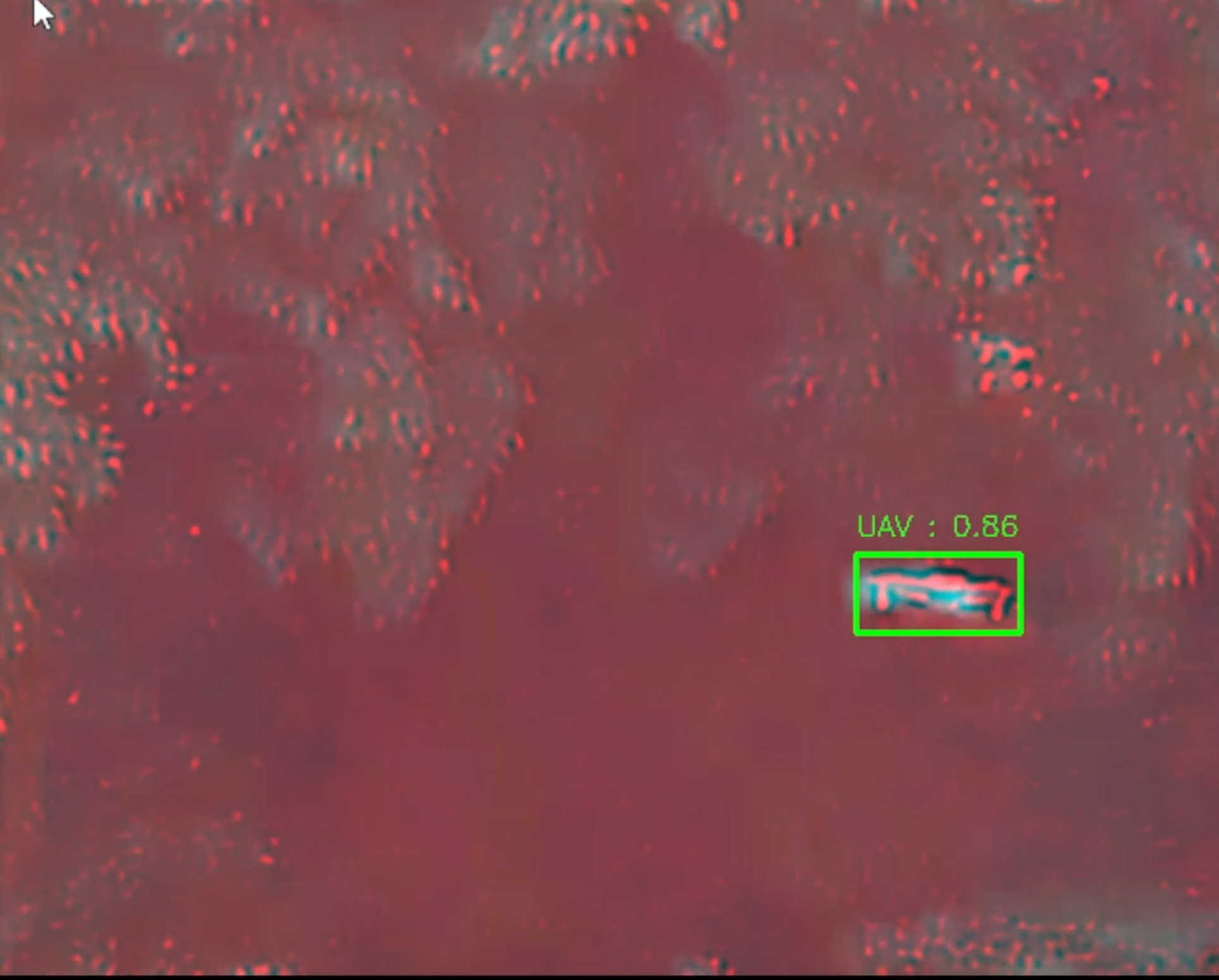}
	\end{subfigure}
	
	\begin{subfigure}{0.11\textwidth}
		\centering
		\includegraphics[width=\linewidth]{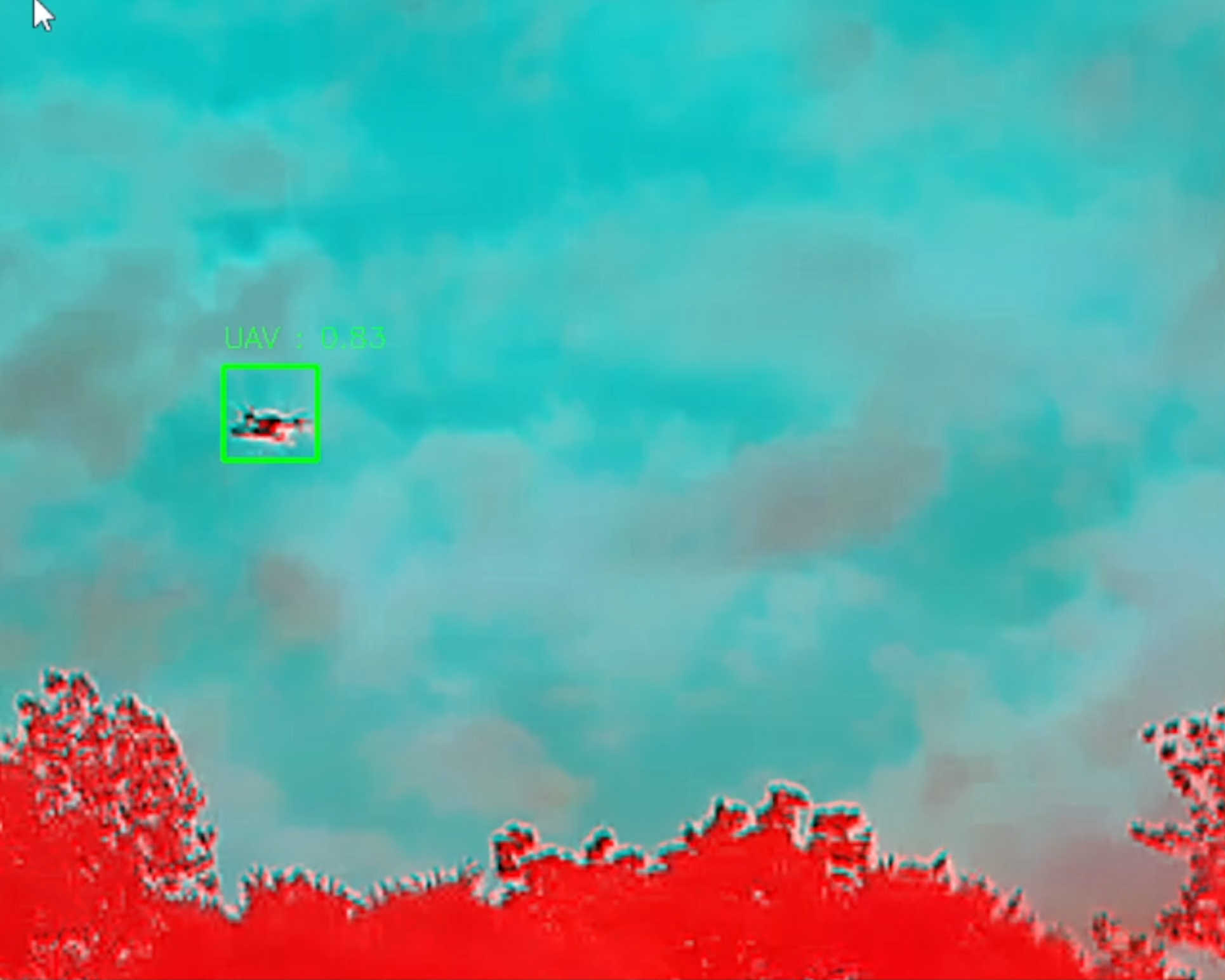}
	\end{subfigure}\hfill
	\begin{subfigure}{0.11\textwidth}
		\centering
		\includegraphics[width=\linewidth]{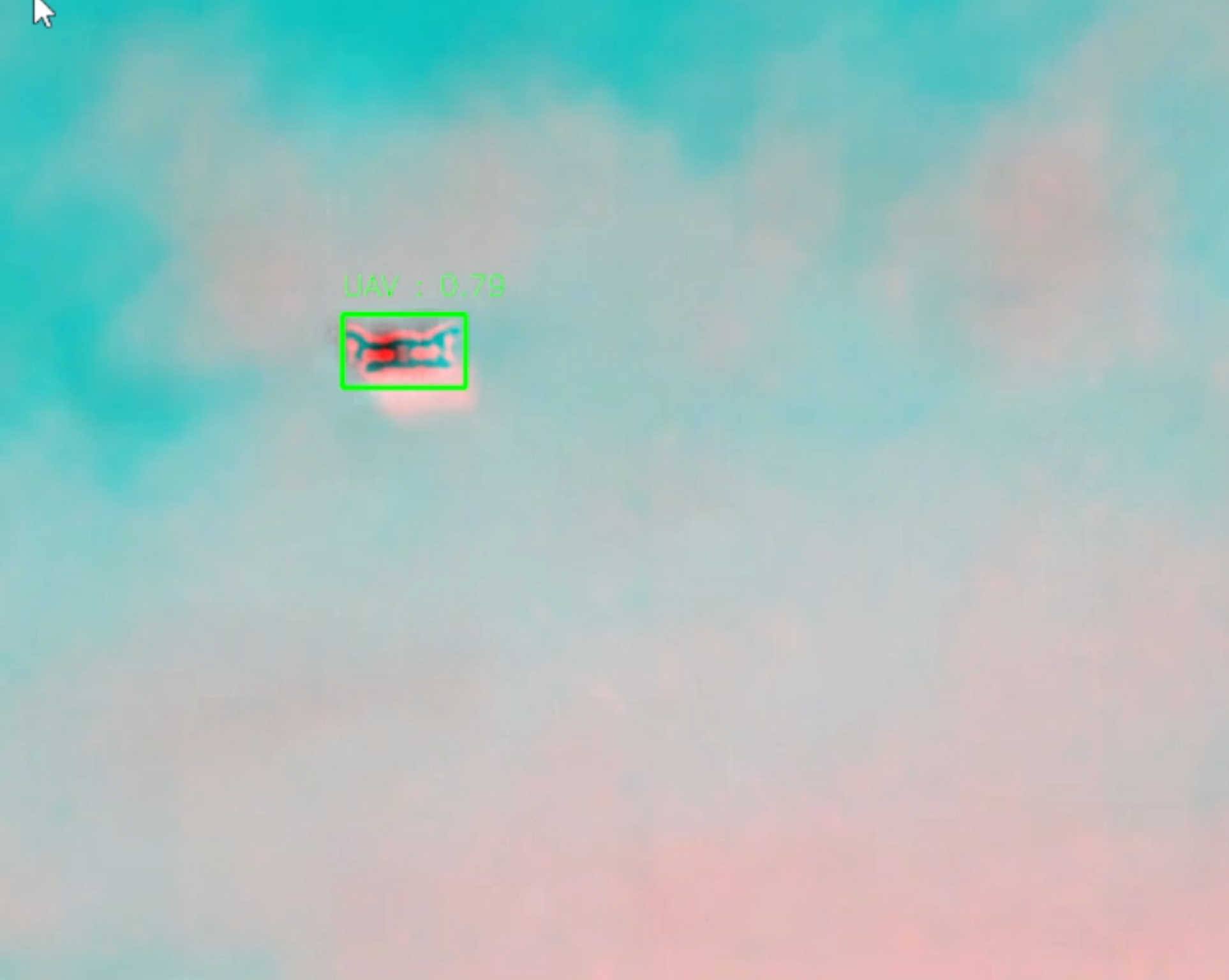}
	\end{subfigure}\hfill
	\begin{subfigure}{0.11\textwidth}
		\centering
		\includegraphics[width=\linewidth]{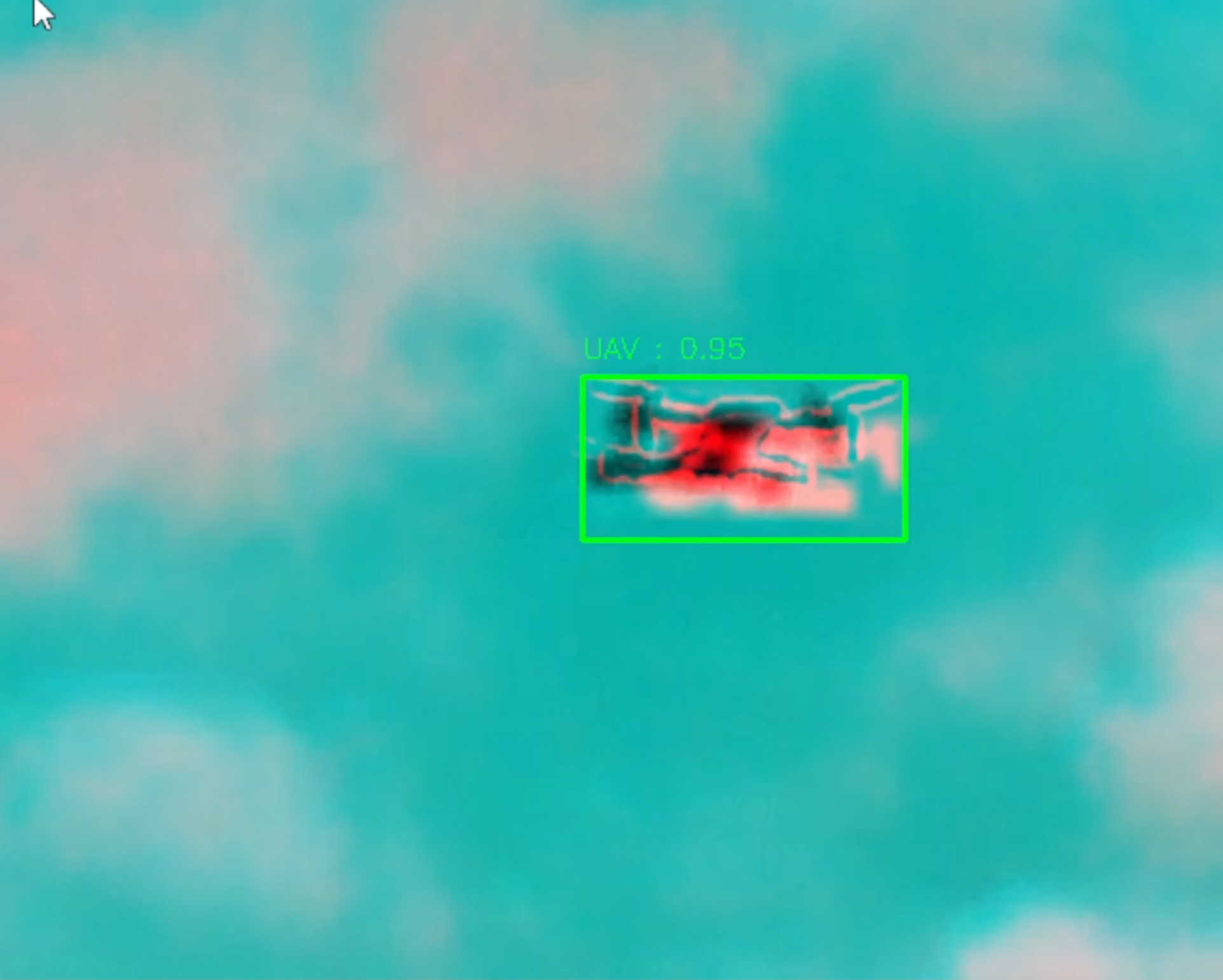}
	\end{subfigure}\hfill
	\begin{subfigure}{0.11\textwidth}
		\centering
		\includegraphics[width=\linewidth]{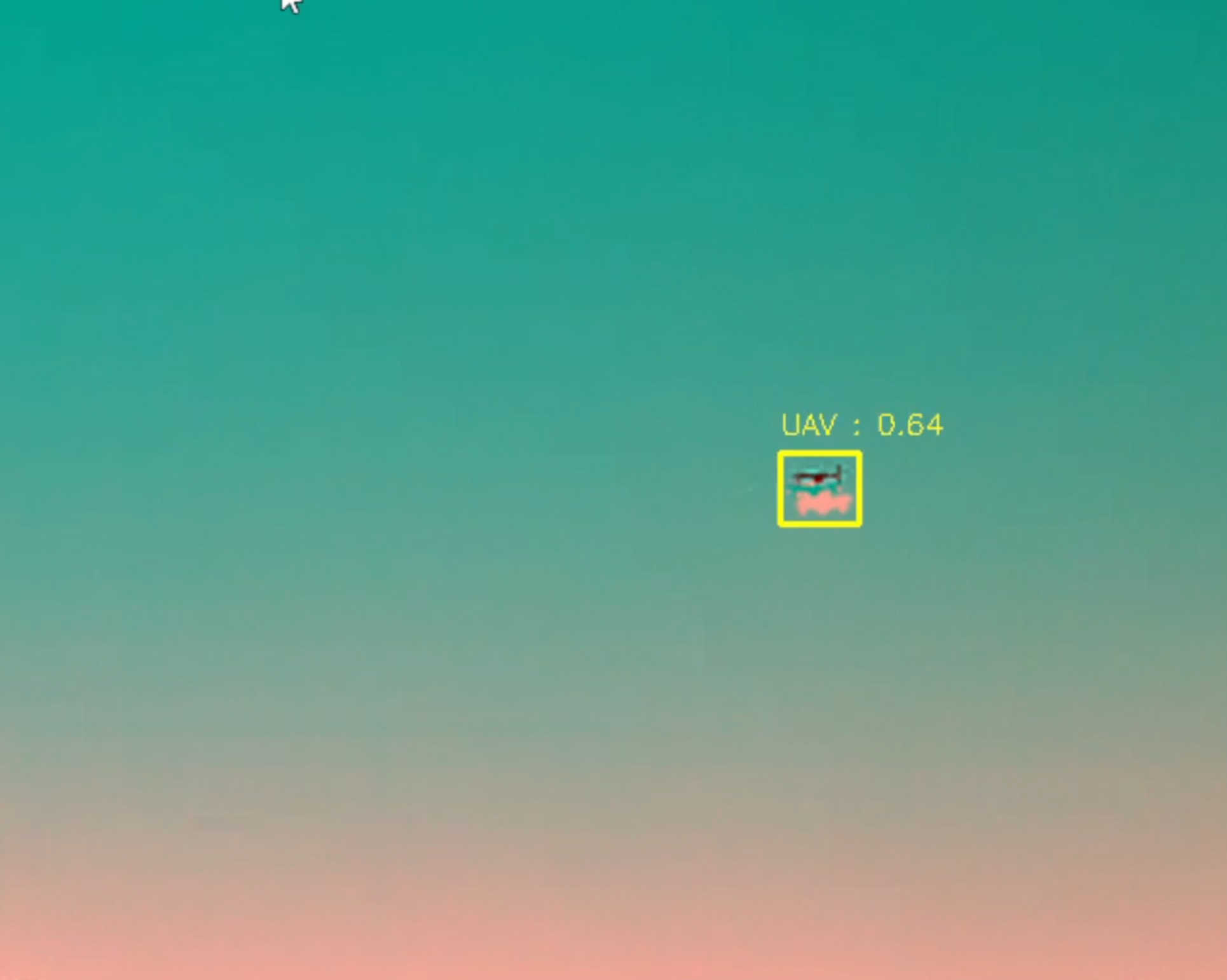}
	\end{subfigure}\hfill
	\begin{subfigure}{0.11\textwidth}
		\centering
		\includegraphics[width=\linewidth]{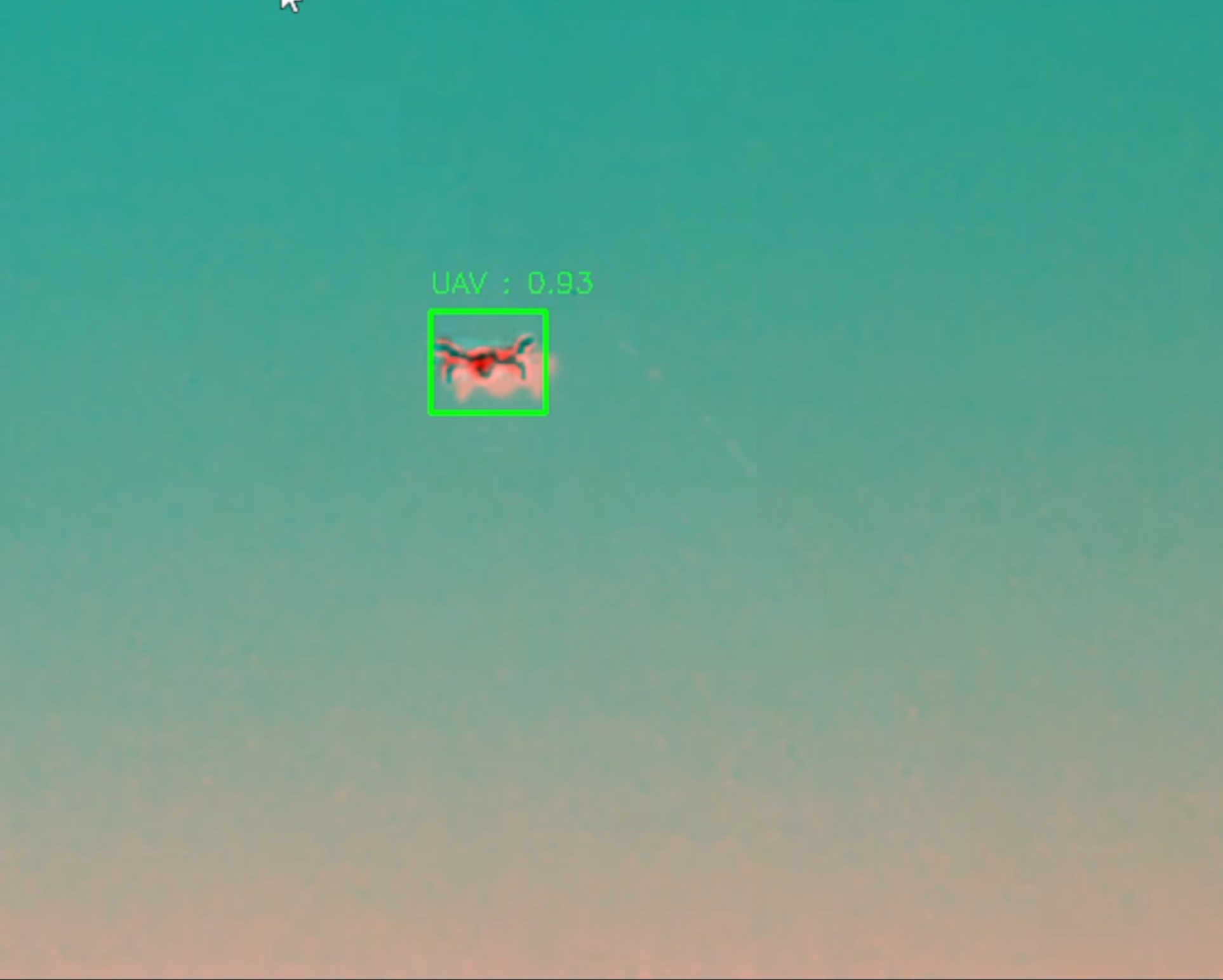}
	\end{subfigure}\hfill
	\begin{subfigure}{0.11\textwidth}
		\centering
		\includegraphics[width=\linewidth]{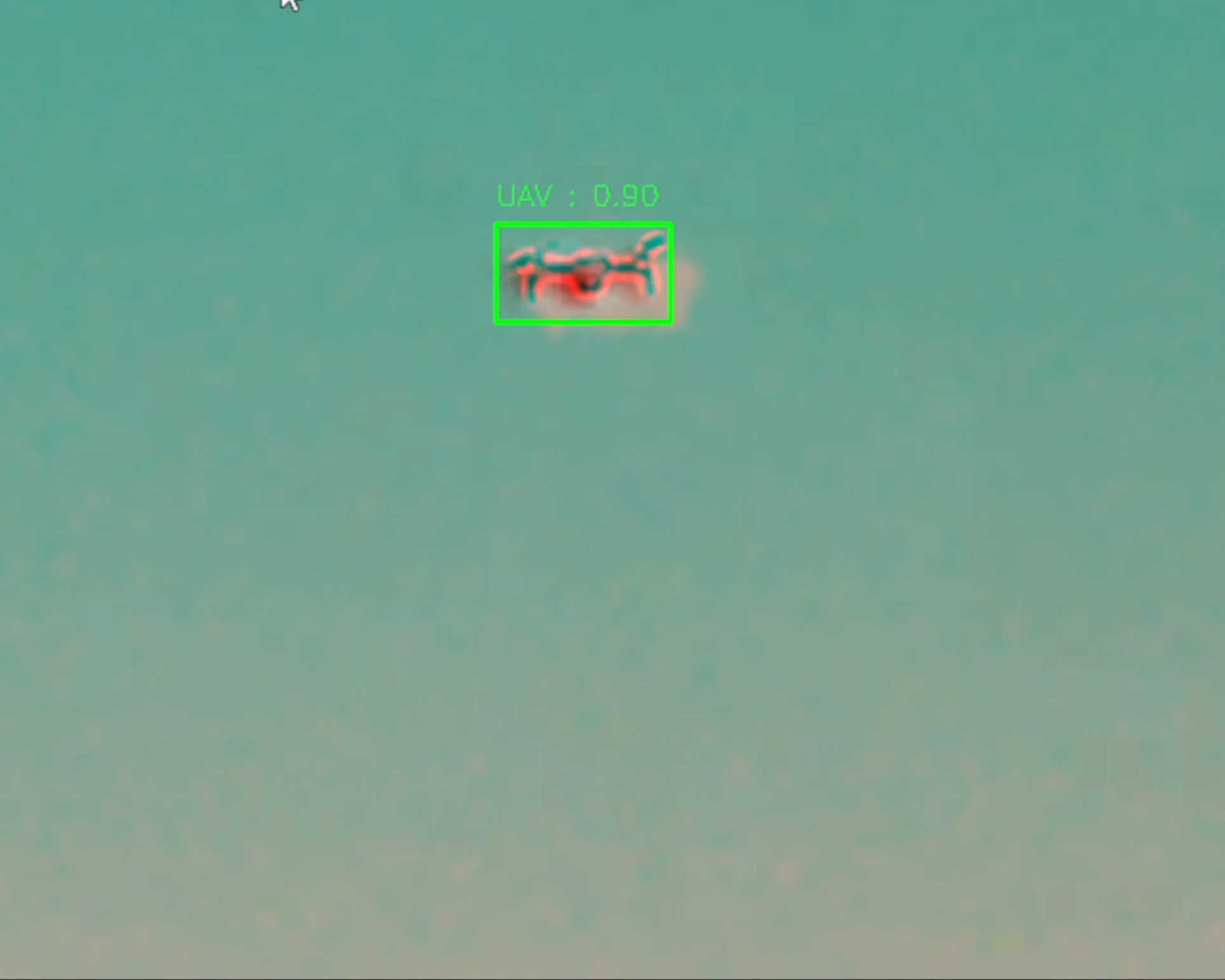}
	\end{subfigure}\hfill
	\begin{subfigure}{0.11\textwidth}
		\centering
		\includegraphics[width=\linewidth]{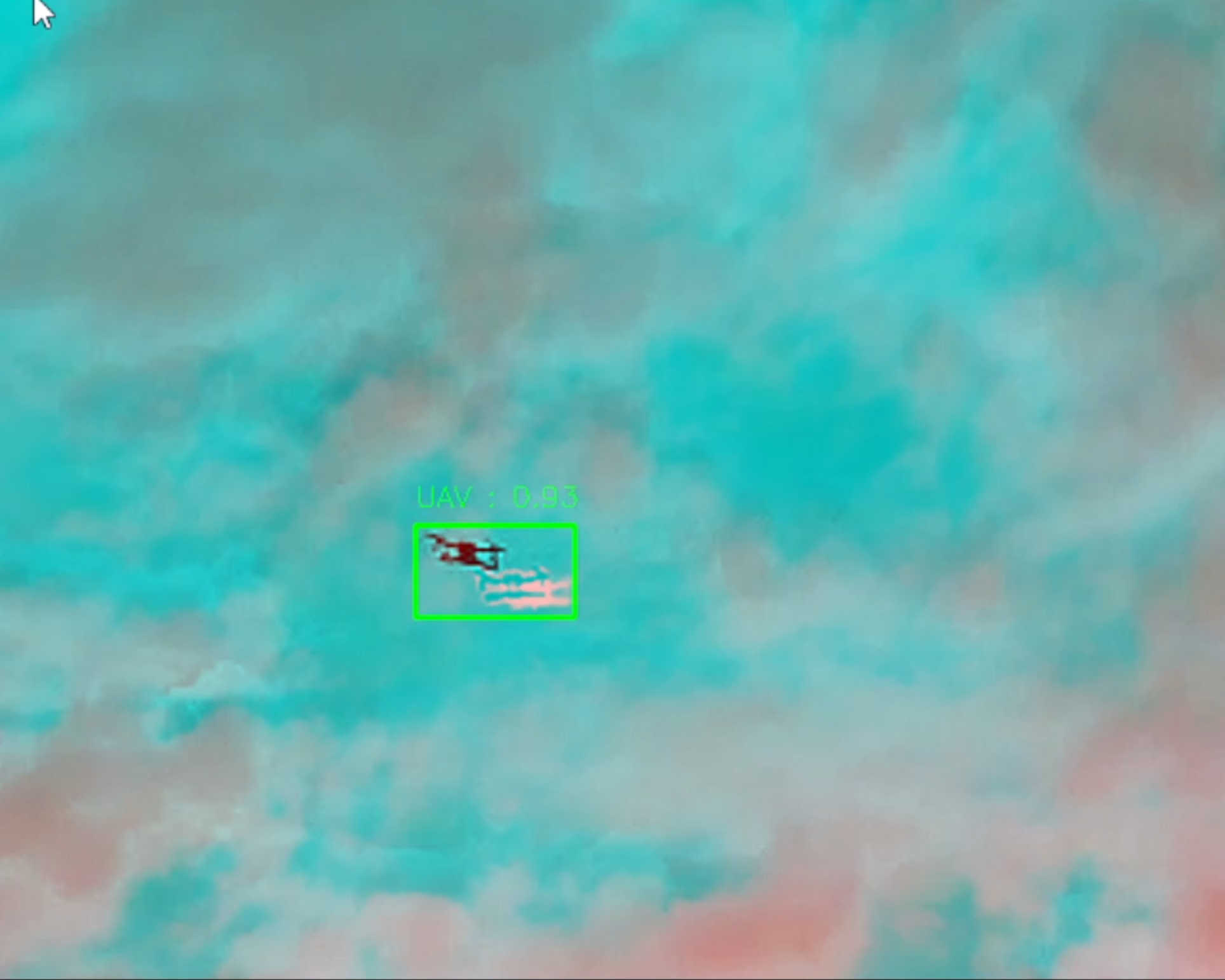}
	\end{subfigure}\hfill
	\begin{subfigure}{0.11\textwidth}
		\centering
		\includegraphics[width=\linewidth]{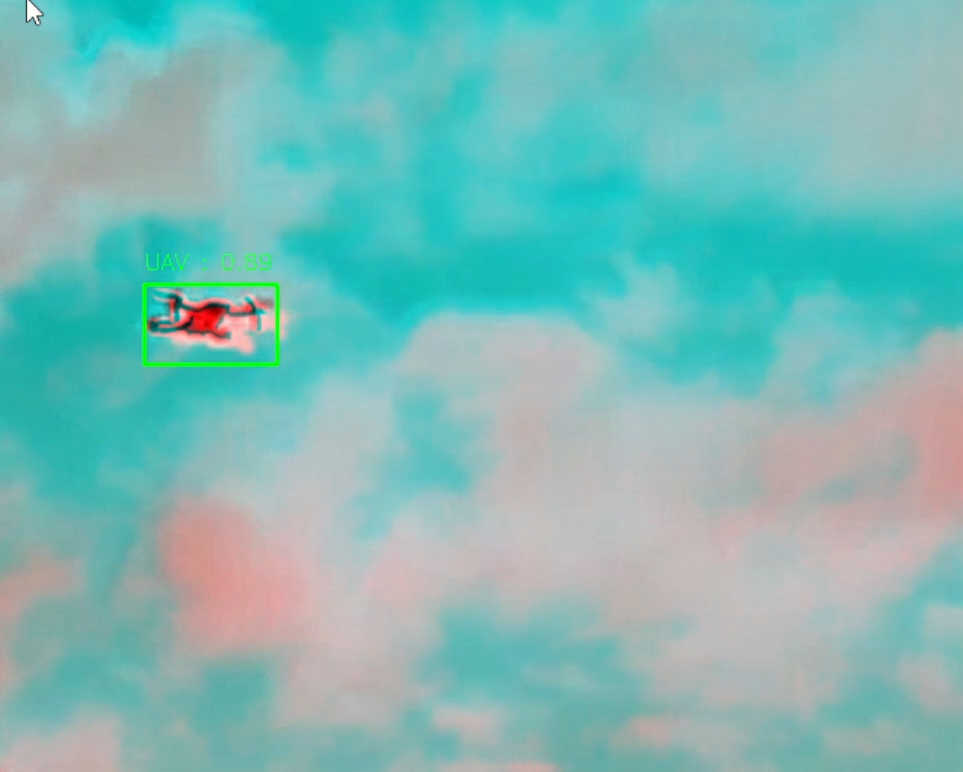}
	\end{subfigure}\hfill
	\begin{subfigure}{0.11\textwidth}
		\centering
		\includegraphics[width=\linewidth]{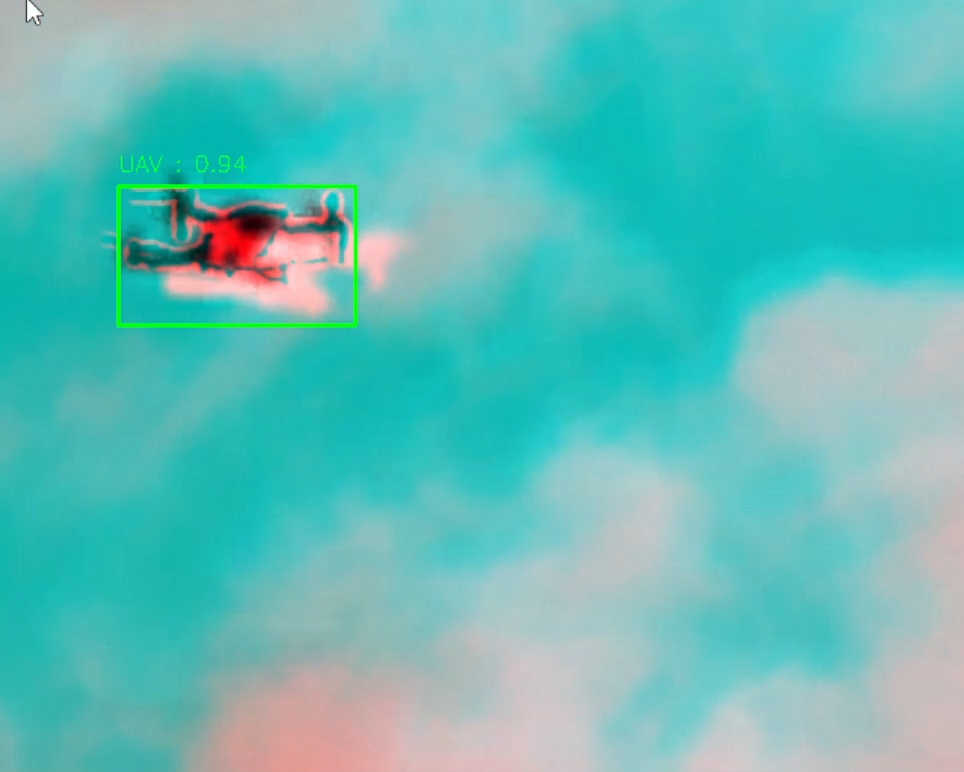}
	\end{subfigure}
	
	\begin{subfigure}{0.11\textwidth}
		\centering
		\includegraphics[width=\linewidth]{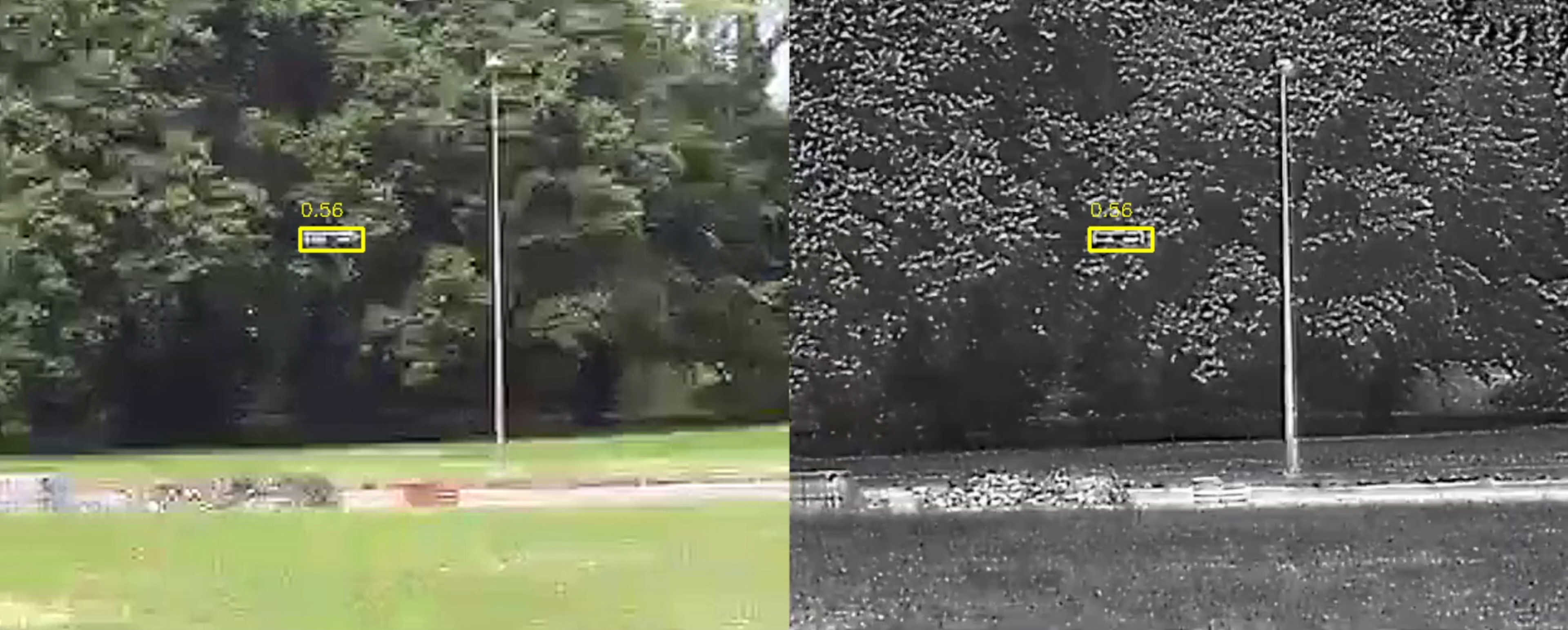}
	\end{subfigure}\hfill
	\begin{subfigure}{0.11\textwidth}
		\centering
		\includegraphics[width=\linewidth]{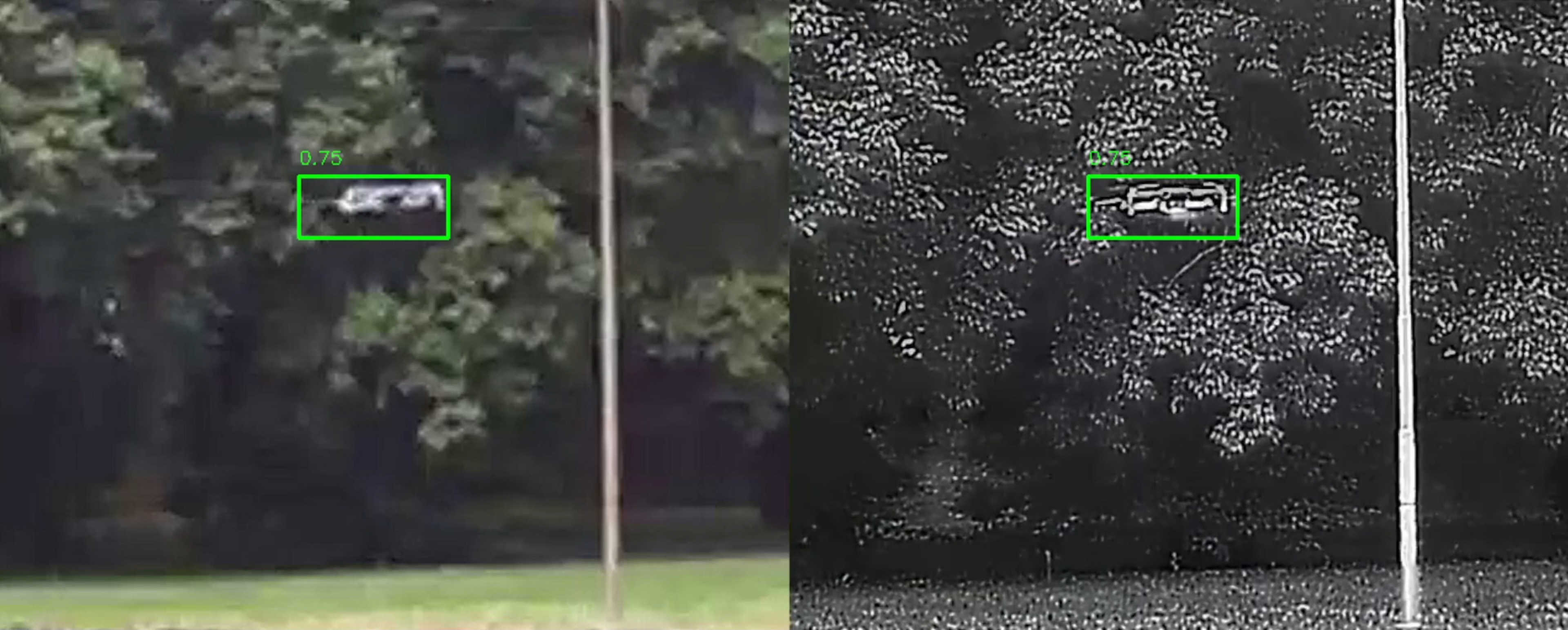}
	\end{subfigure}\hfill
	\begin{subfigure}{0.11\textwidth}
		\centering
		\includegraphics[width=\linewidth]{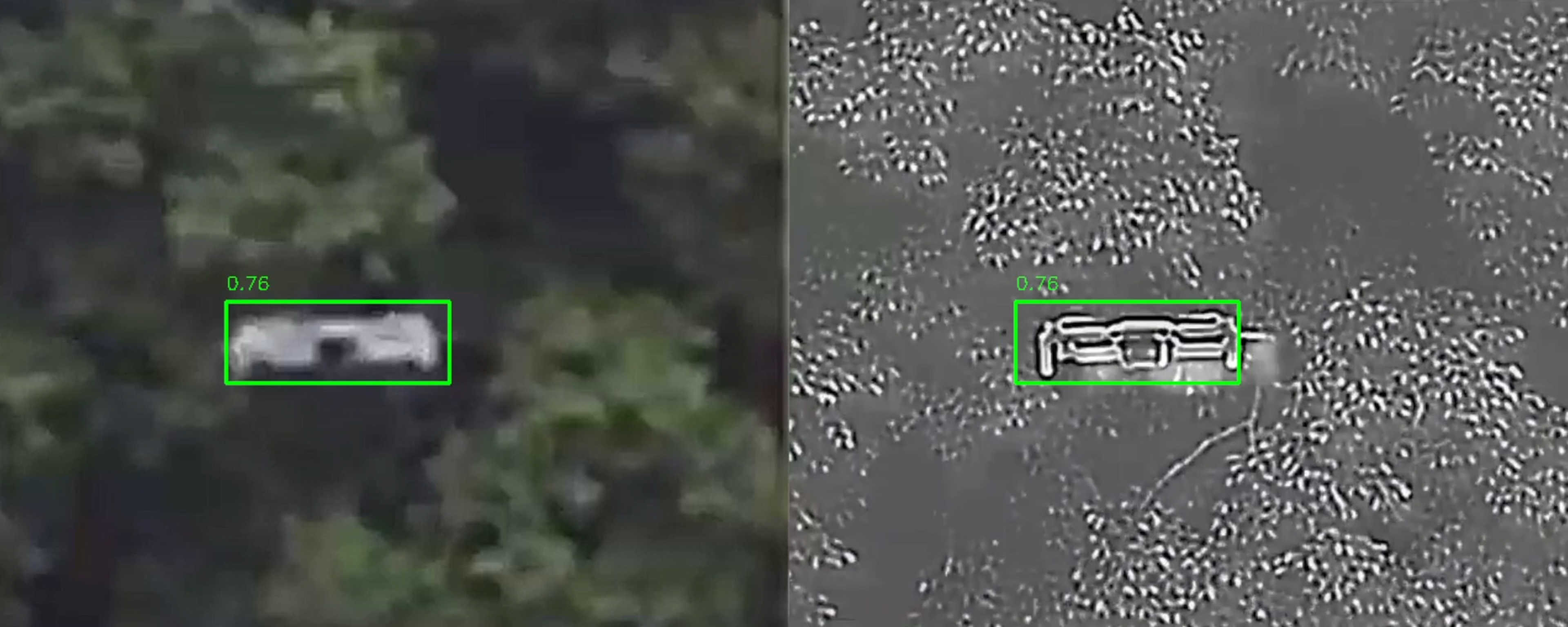}
	\end{subfigure}\hfill
	\begin{subfigure}{0.11\textwidth}
		\centering
		\includegraphics[width=\linewidth]{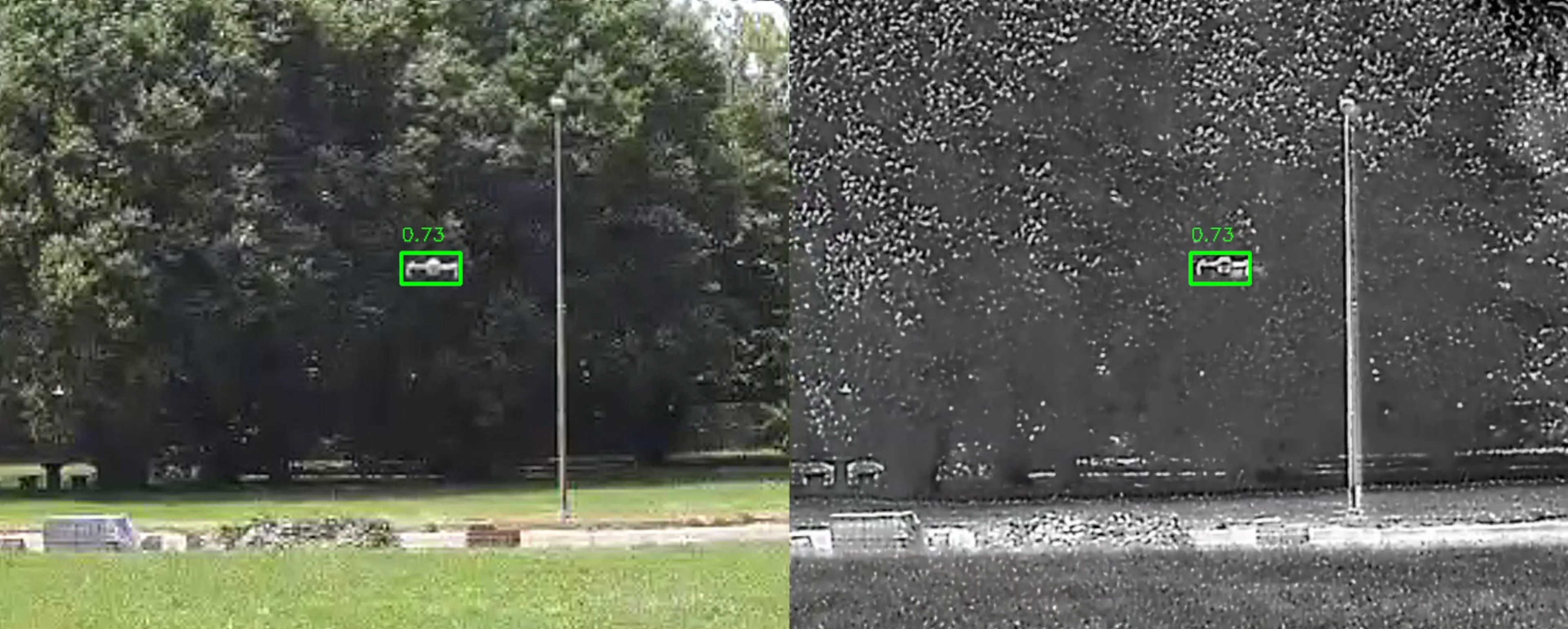}
	\end{subfigure}\hfill
	\begin{subfigure}{0.11\textwidth}
		\centering
		\includegraphics[width=\linewidth]{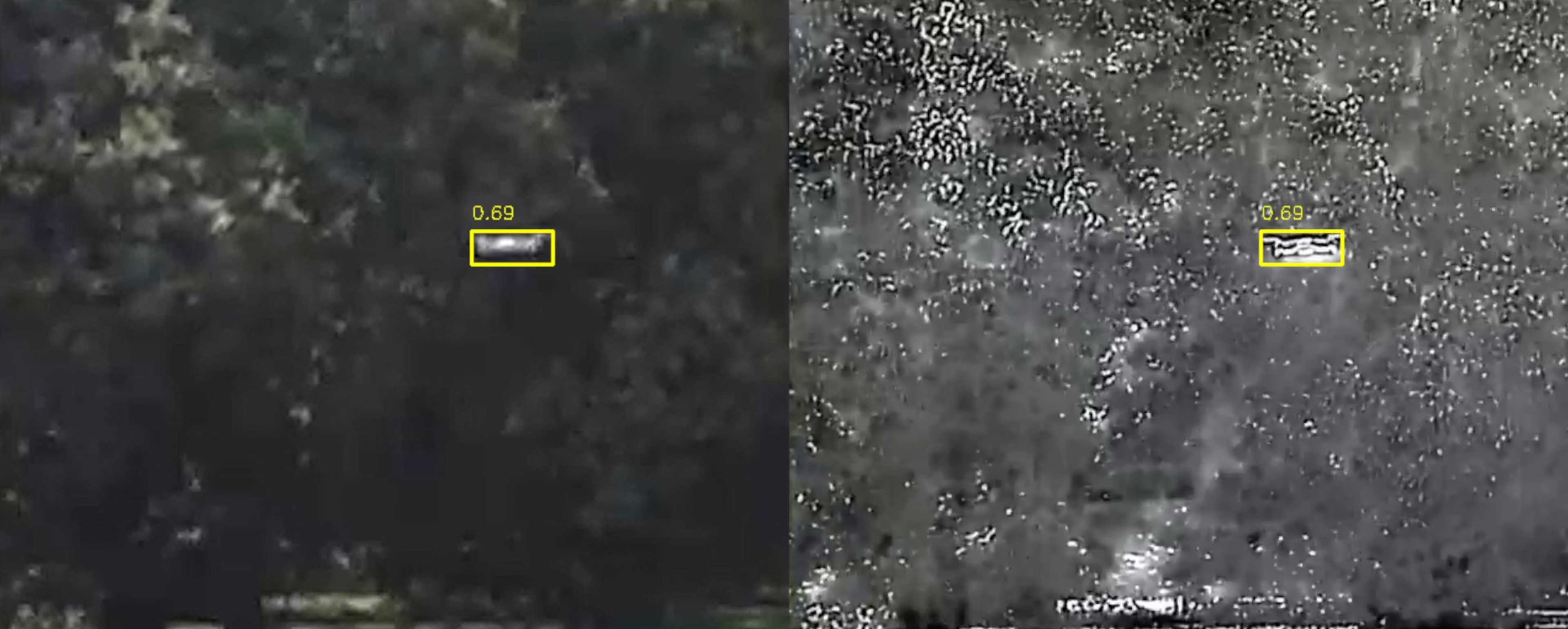}
	\end{subfigure}\hfill
	\begin{subfigure}{0.11\textwidth}
		\centering
		\includegraphics[width=\linewidth]{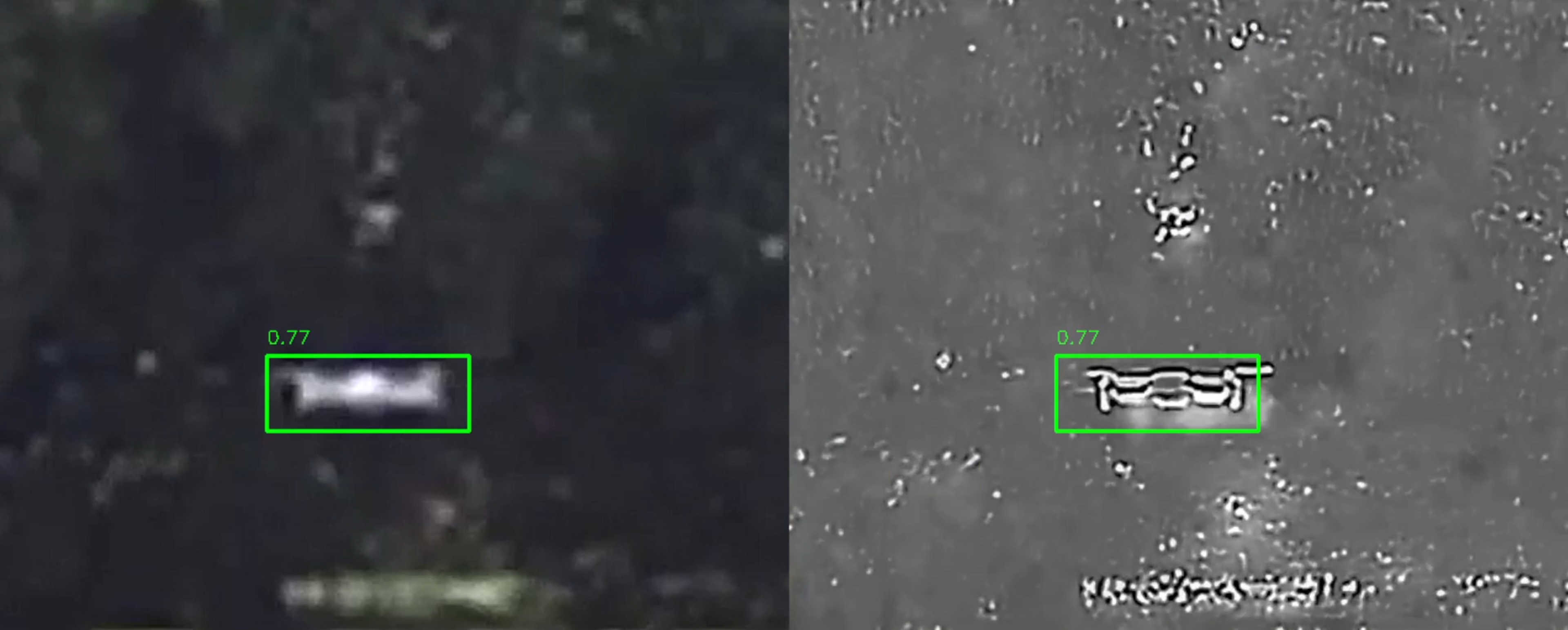}
	\end{subfigure}\hfill
	\begin{subfigure}{0.11\textwidth}
		\centering
		\includegraphics[width=\linewidth]{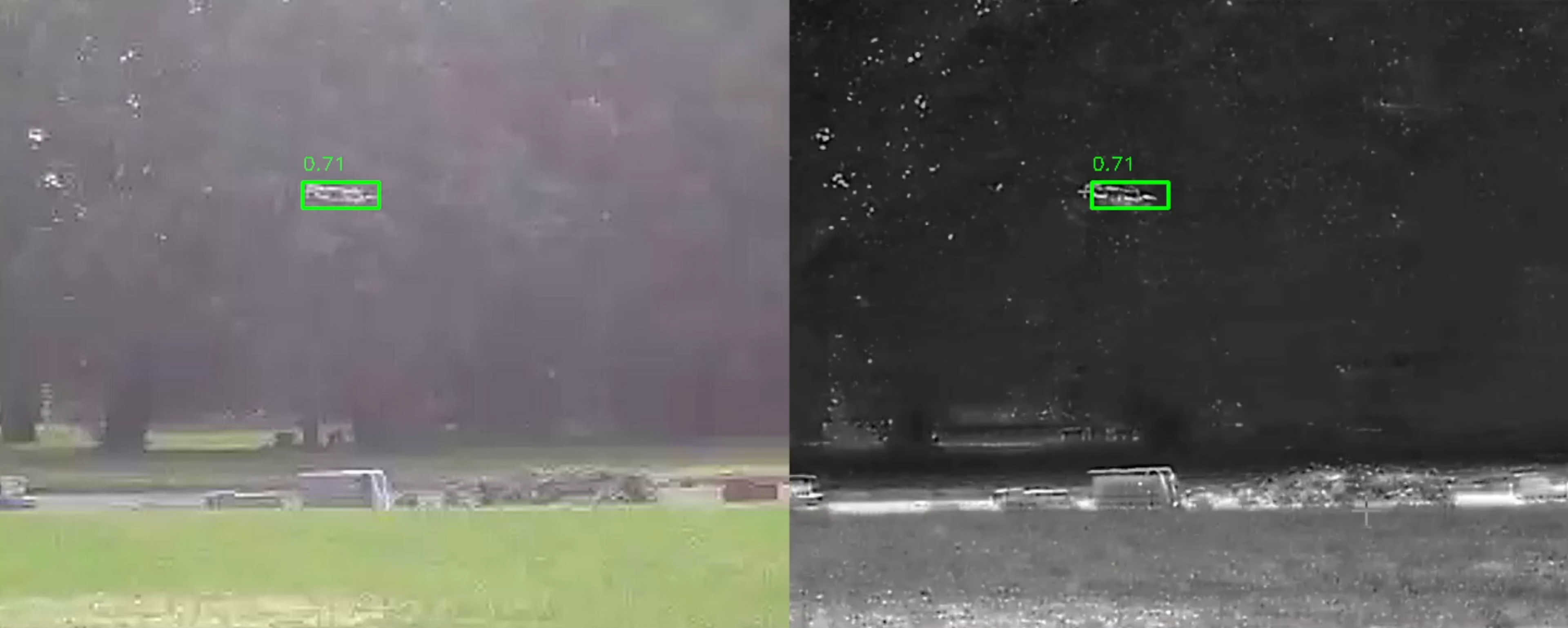}
	\end{subfigure}\hfill
	\begin{subfigure}{0.11\textwidth}
		\centering
		\includegraphics[width=\linewidth]{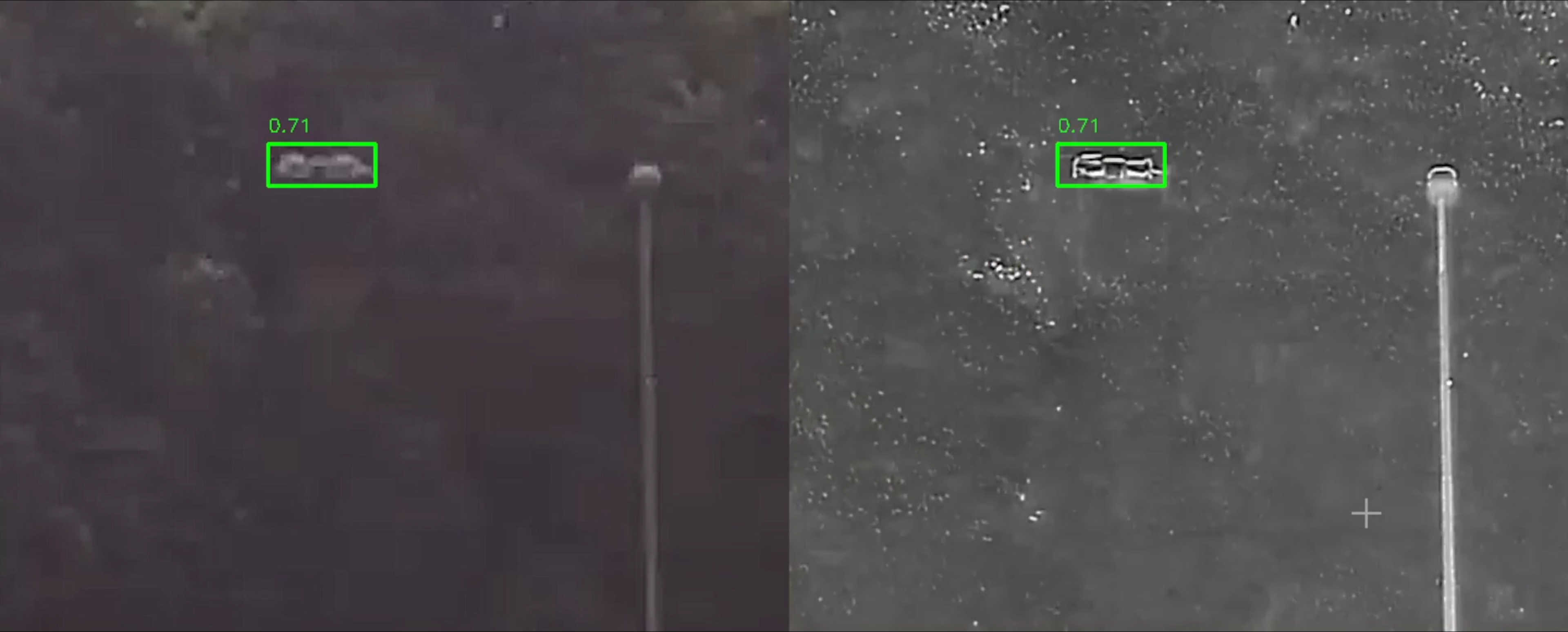}
	\end{subfigure}\hfill
	\begin{subfigure}{0.11\textwidth}
		\centering
		\includegraphics[width=\linewidth]{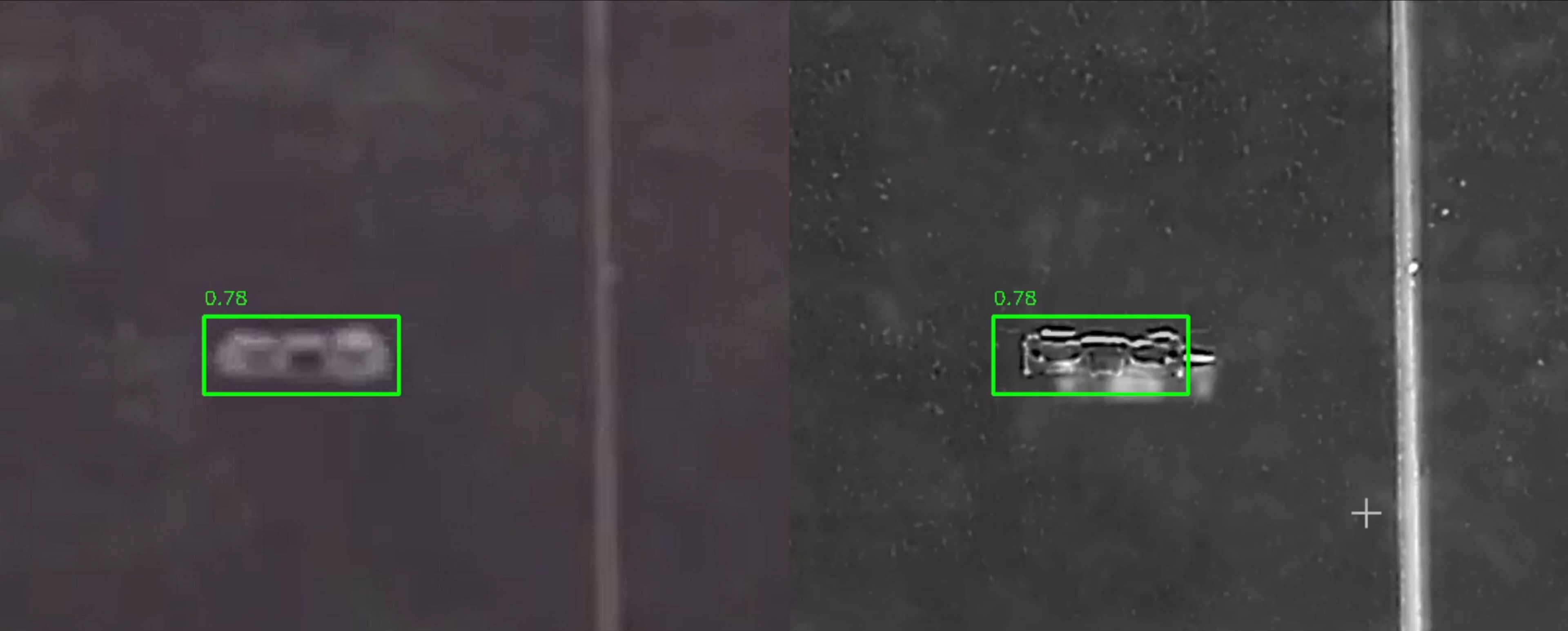}
	\end{subfigure}
	
	\begin{subfigure}{0.11\textwidth}
		\centering
		\includegraphics[width=\linewidth]{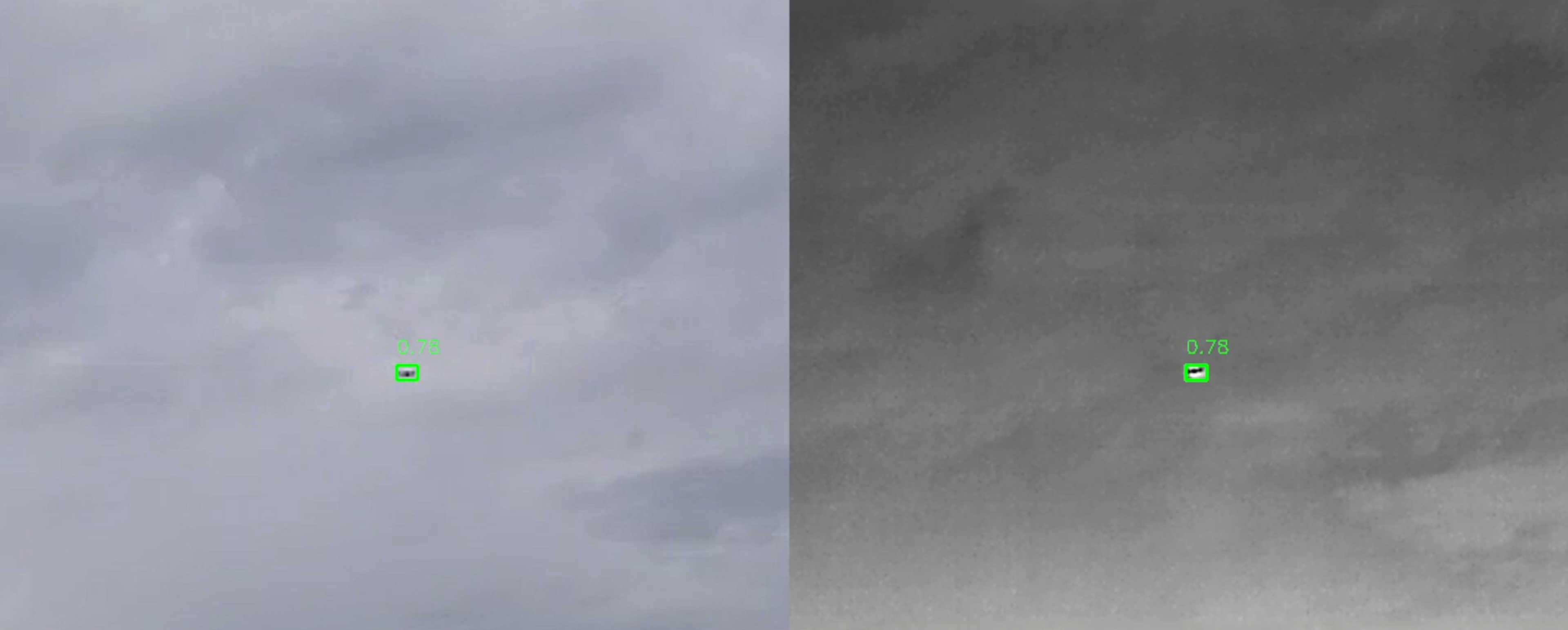}
	\end{subfigure}\hfill
	\begin{subfigure}{0.11\textwidth}
		\centering
		\includegraphics[width=\linewidth]{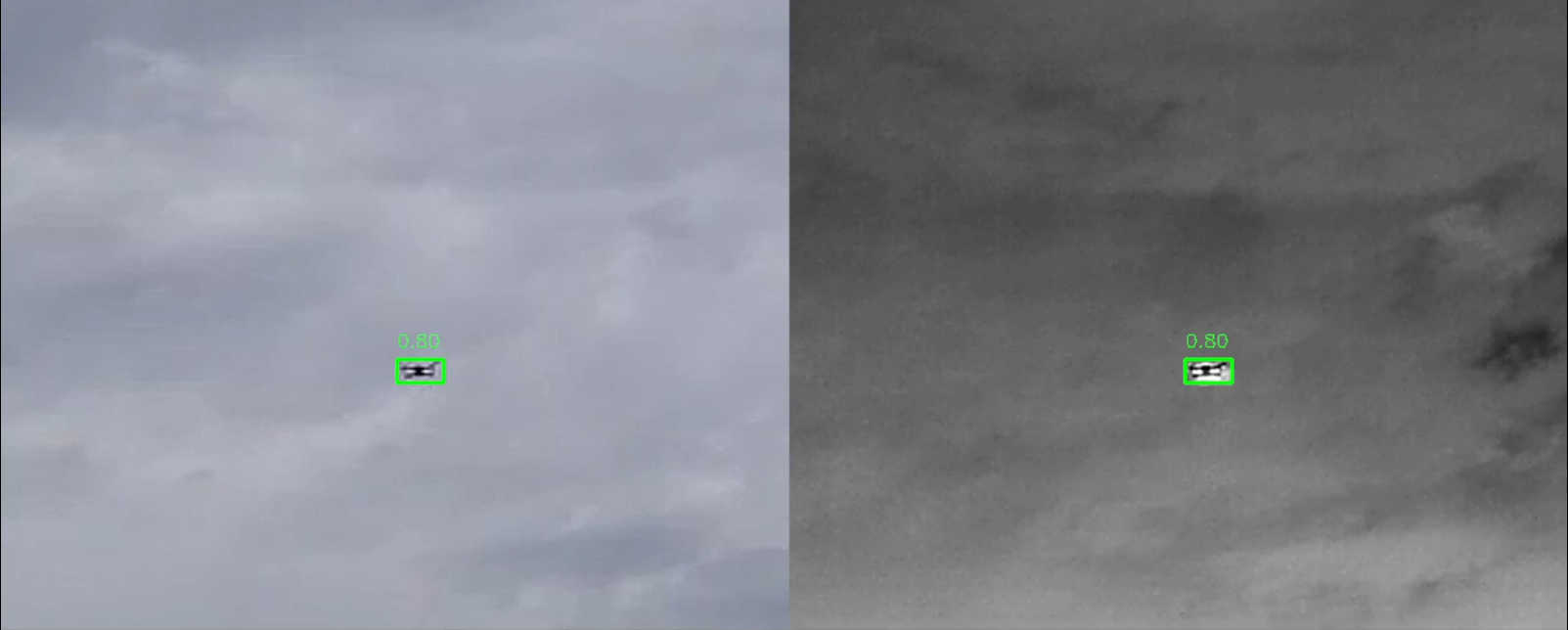}
	\end{subfigure}\hfill
	\begin{subfigure}{0.11\textwidth}
		\centering
		\includegraphics[width=\linewidth]{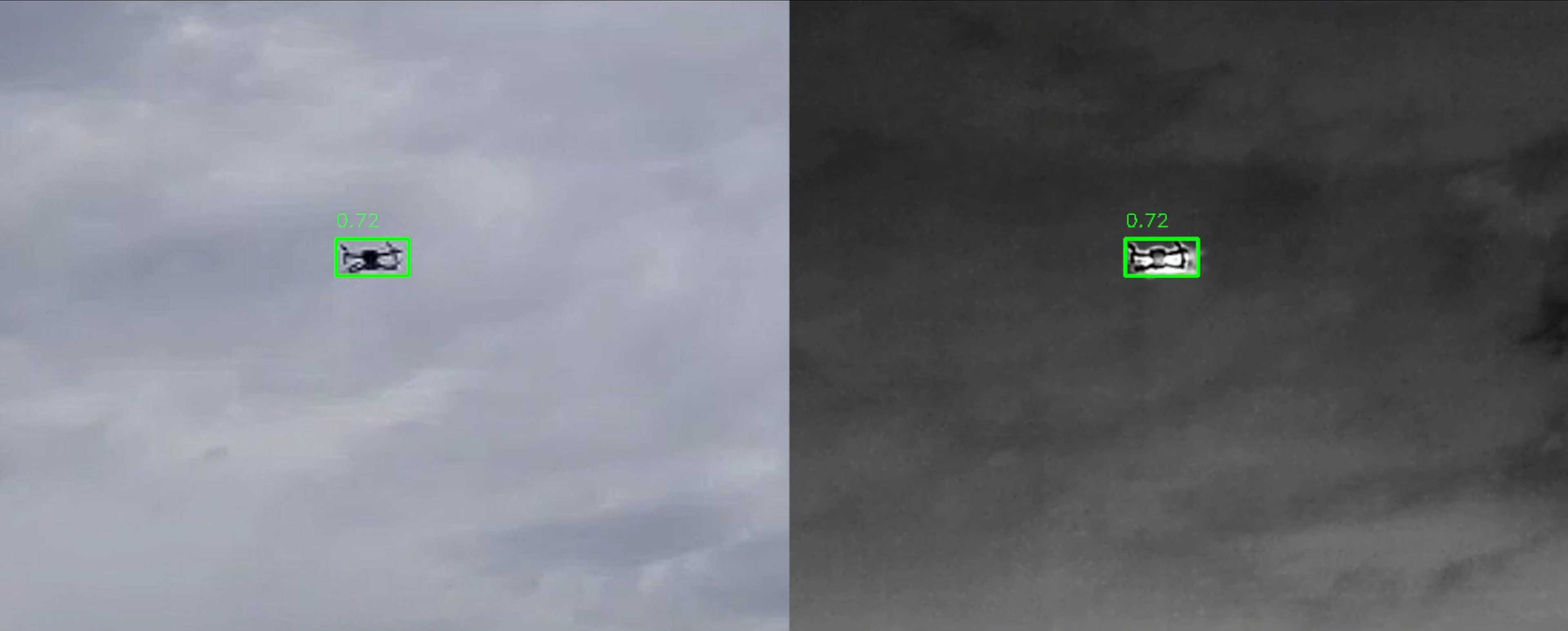}
	\end{subfigure}\hfill
	\begin{subfigure}{0.11\textwidth}
		\centering
		\includegraphics[width=\linewidth]{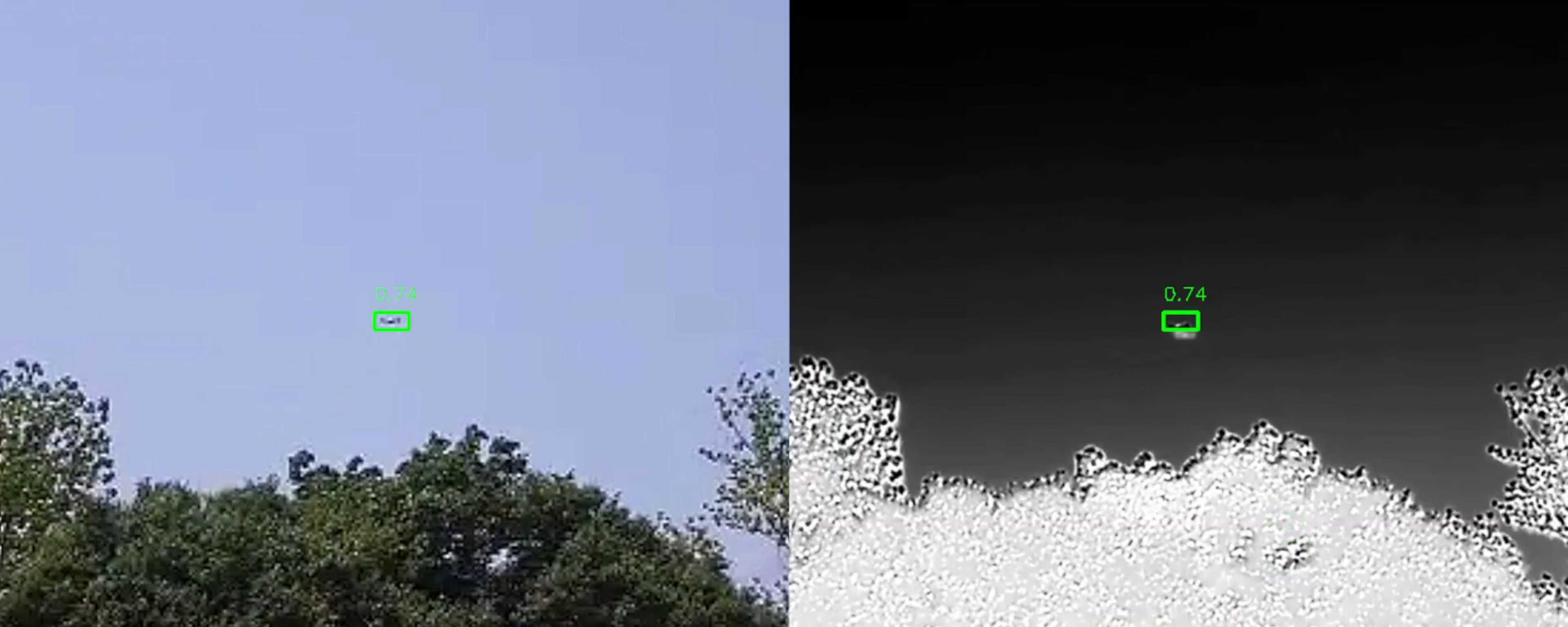}
	\end{subfigure}\hfill
	\begin{subfigure}{0.11\textwidth}
		\centering
		\includegraphics[width=\linewidth]{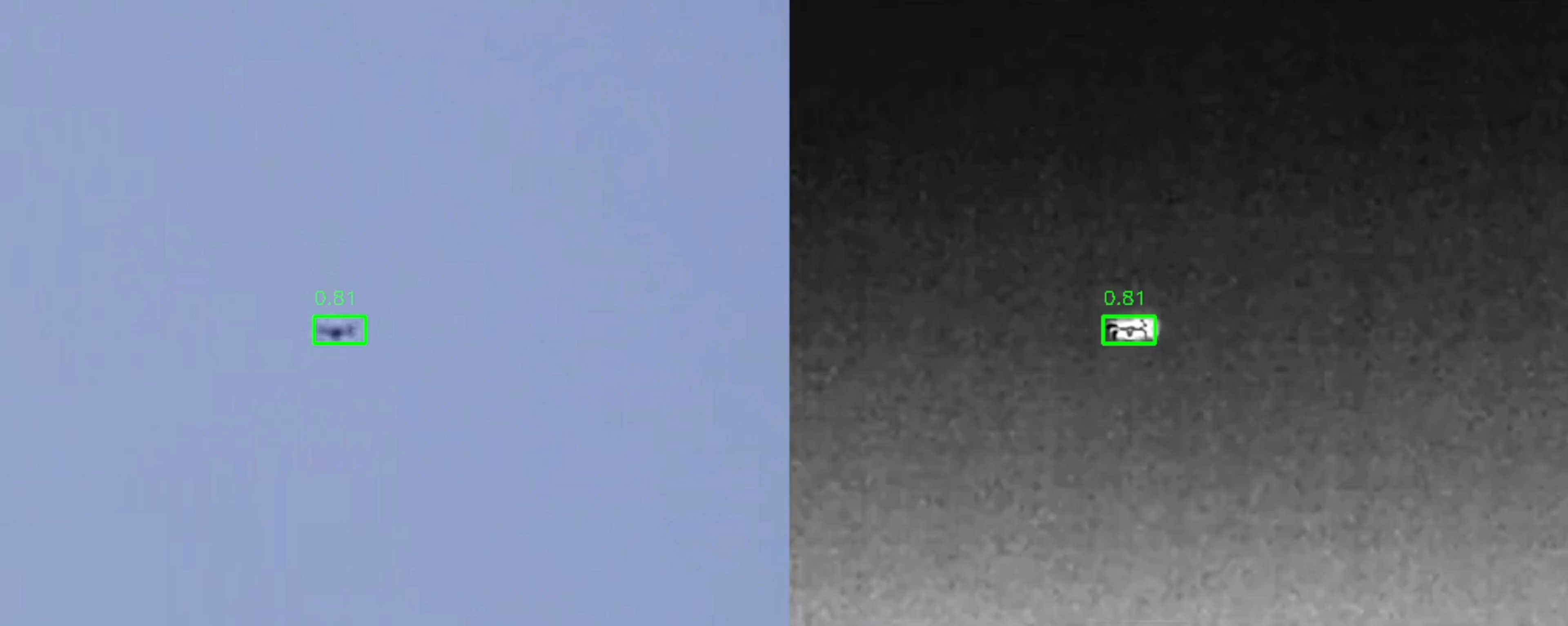}
	\end{subfigure}\hfill
	\begin{subfigure}{0.11\textwidth}
		\centering
		\includegraphics[width=\linewidth]{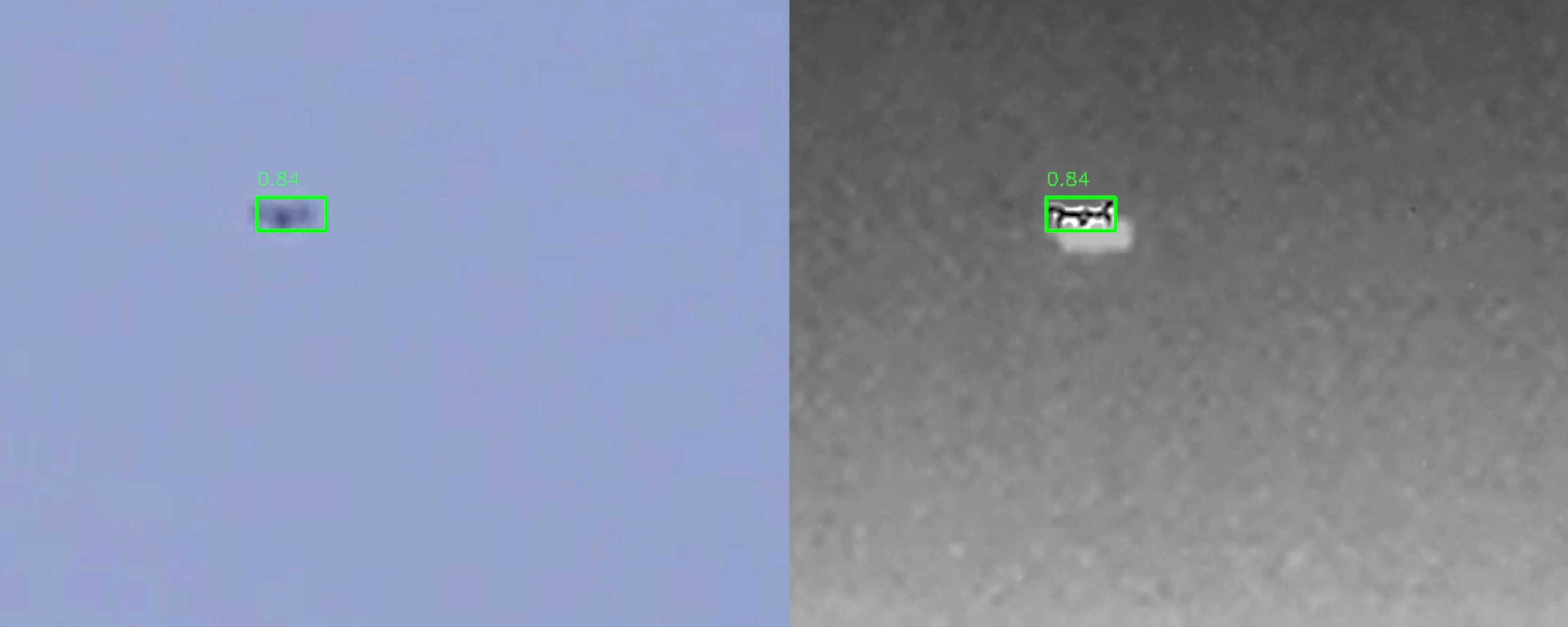}
	\end{subfigure}\hfill
	\begin{subfigure}{0.11\textwidth}
		\centering
		\includegraphics[width=\linewidth]{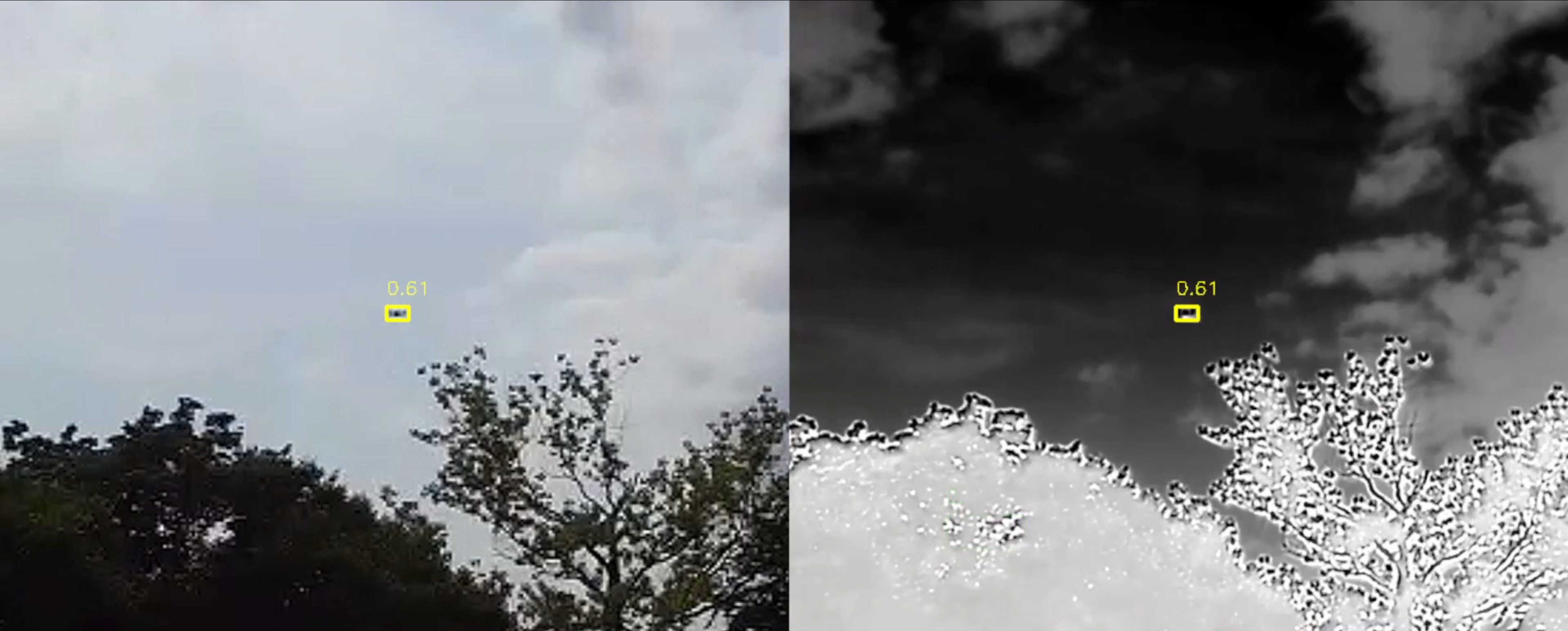}
	\end{subfigure}\hfill
	\begin{subfigure}{0.11\textwidth}
		\centering
		\includegraphics[width=\linewidth]{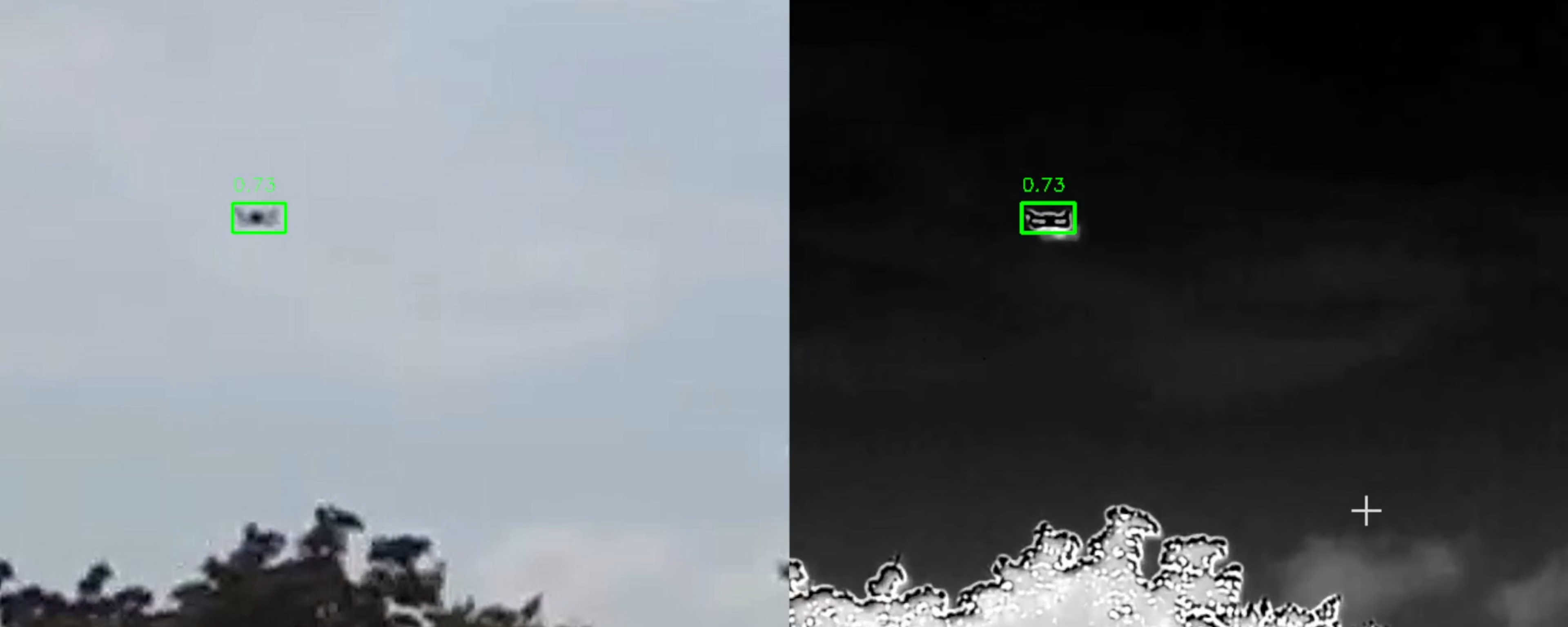}
	\end{subfigure}\hfill
	\begin{subfigure}{0.11\textwidth}
		\centering
		\includegraphics[width=\linewidth]{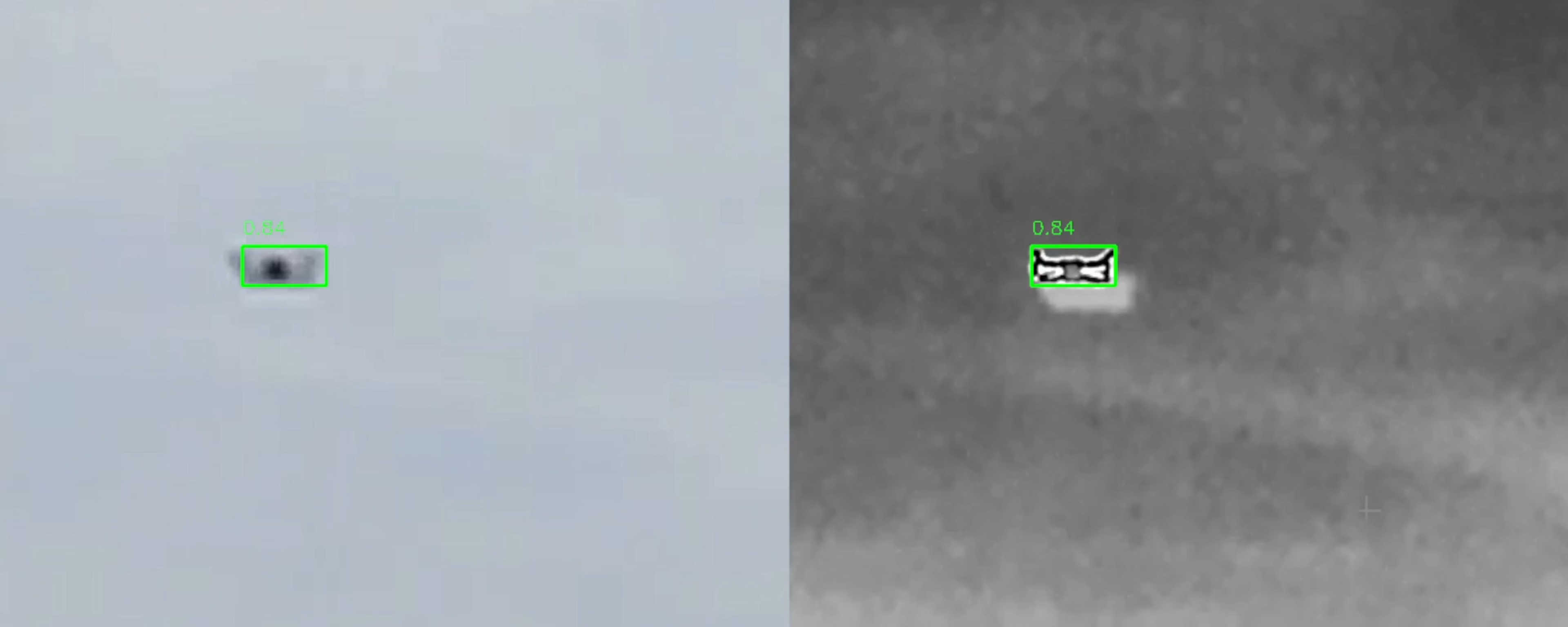}
	\end{subfigure}
	\caption{Ground and sky comparisons at different times and resolutions. From the top left corner of the image, it is possible to see the first line, regarding Early Fusion RGT process on the ground at different zoom levels (1X, 2X, 4X) in different times of day (9:00, 12:00, 15:00). Following it, there is a second line showing the same fusion techniques applied for detecting over a sky background. Finally, the third and fourth lines contain images of tests conducted in the same conditions presented before while using Late Fusion instead.}
\end{figure*}
Different lighting scenarios were faced during the experiments. For safety reasons, a maximum distance of 60 meters and an altitude of 20 m were set for all the tests. Zooming in was also used in borderline cases, where the background was wooded and the drone was camouflaged among the leaves, thus difficult to detect. Confidence scores were increasing when zooming in and the drone was still detected with all degrees of zoom and at great distances. However, detection at 1X is sometimes intermittent, whereas 2X and 4X work reliably, making the system more stable. Furthermore, the score drops when the target flies in front of an obstacle. At those hours, temperatures were lower, which prevented disturbances from light and heat. On the other hand, the sky background greatly improved the confidence score, even with clouds. In fact, the cloud formation let light pass through and the environment was at least sufficiently illuminated, enhancing the contrast between the target UAV and the sky color.\\
Late-fusion technique was more challenging due to the management of two independent models. As can be seen, the stream was aligned during detection, in order to have both frames in the same plane with a horizontal sync. This method proves particularly useful in environments in which one of the two modalities is less reliable. In fact, the best of the two helps the other to perform detection, thanks to the weighted average strategy developed. As before, sky offers higher confidence scores, due to the better contrast between the UAV and the background. However, some small issues have been discovered with trees and terrain setup, as they generate clutter, a phenomenon that challenges the optical sensors. In addition, leaves make detection even more difficult due to the openings inside the trees, which let light in, reducing the possibility of detection. Moreover, since we recorded on summer days, the heat and brightness affected the infrared sensors, generating noise. Another aspect which should be taken into consideration regards UAV movements. If the drone is stationary or changes direction quickly, it can be lost. However, if the drone is in the air, it is much better identified, even at great distances.
\begin{table}[!htbp]
\scriptsize
\centering
\caption{Summary of the experimental evaluation conducted at different hours of the day, under different environmental conditions. The target drone was moving at different altitudes and distances from the master drone, while covering two main backgrounds: trees and sky. In the same way, three zoom levels have been tested.}
\label{tab:experimental_summary}
\begin{tabular}{cccccc}
\toprule
\textbf{Time} & \textbf{Fusion} & \textbf{Zoom} & \textbf{Background} & \textbf{Distance} & \textbf{Altitude} \\
\midrule
09:00 & Early & 1$\times$, 2$\times$, 4$\times$ & Trees & 5, 25, 60 m & 5, 10, 20 m \\
09:00 & Early & 1$\times$, 2$\times$, 4$\times$ & Sky & 5, 25, 60 m & 5, 10, 20 m \\
09:00 & Late  & 1$\times$, 2$\times$, 4$\times$ & Trees & 5, 25, 60 m & 5, 10, 20 m \\
09:00 & Late  & 1$\times$, 2$\times$, 4$\times$ & Sky & 5, 25, 60 m & 5, 10, 20 m \\
\midrule
12:00 & Early & 1$\times$, 2$\times$, 4$\times$ & Trees & 5, 25, 60 m & 5, 10, 20 m \\
12:00 & Early & 1$\times$, 2$\times$, 4$\times$ & Sky & 5, 25, 60 m & 5, 10, 20 m \\
12:00 & Late & 1$\times$, 2$\times$, 4$\times$ & Trees & 5, 25, 60 m & 5, 10, 20 m \\
12:00 & Late & 1$\times$, 2$\times$, 4$\times$ & Sky & 5, 25, 60 m & 5, 10, 20 m \\
\midrule
15:00 & Early & 1$\times$, 2$\times$, 4$\times$ & Trees & 5, 25, 60 m & 5, 10, 20 m \\
15:00 & Early  & 1$\times$, 2$\times$, 4$\times$ & Sky & 5, 25, 60 m & 5, 10, 20 m \\
15:00 & Late & 1$\times$, 2$\times$, 4$\times$ & Trees & 5, 25, 60 m & 5, 10, 20 m  \\
15:00 & Late  & 1$\times$, 2$\times$, 4$\times$ & Sky & 5, 25, 60 m & 5, 10, 20 m \\
\bottomrule
\end{tabular}
\end{table}

Late-Fusion requires two detection models, increasing inference time and computational cost. Each model has 21.5~MB of weights and processes a 640$\times$640 input, resulting in approximately twice the model and inference requirements of RGT.

\section{Results}
\begin{figure}[!htbp]
	\centering
	\begin{minipage}{0.15\textwidth}
		\centering
		\includegraphics[width=1.1\linewidth]{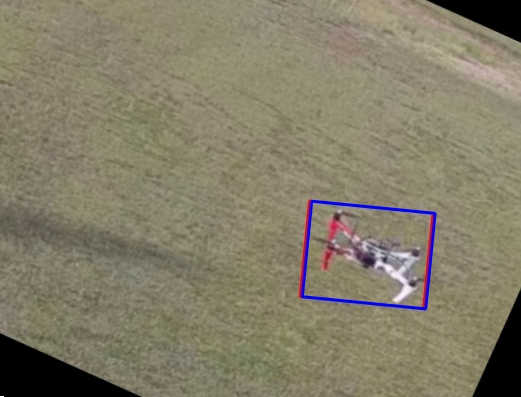}
		\label{fig:rgb_results}
	\end{minipage}%
	\hspace{2pt}
	\begin{minipage}{0.15\textwidth}
		\centering
		\includegraphics[width=1.04\linewidth]{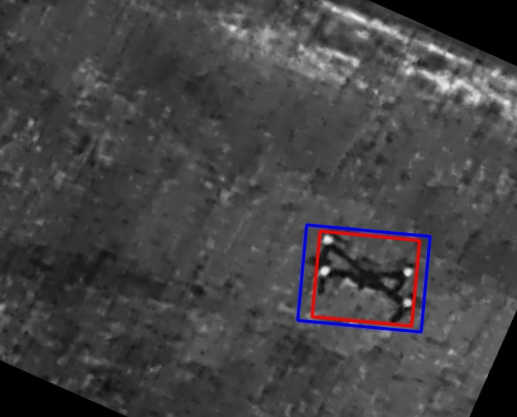}
		\label{fig:ir_results}
	\end{minipage}%
	\hfill
	\begin{minipage}{0.15\textwidth}
		\centering
		\includegraphics[width=1.12\linewidth]{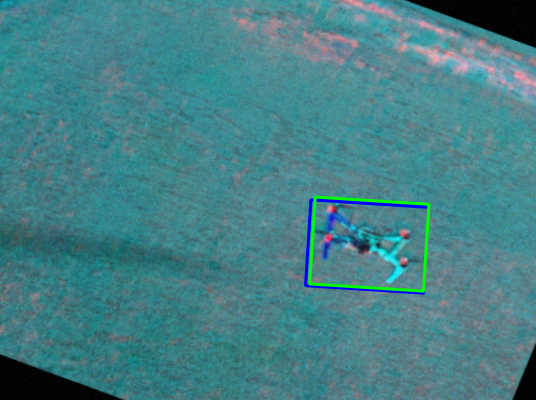}
		\label{fig:rgt_results}
	\end{minipage}
	\centering
	\begin{minipage}{0.15\textwidth}
		\centering
		\includegraphics[width=1.1\linewidth]{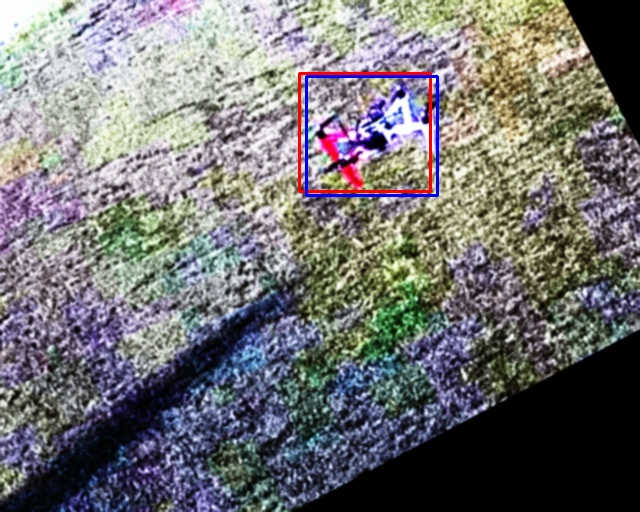}
		\label{fig:rgb_results2}
	\end{minipage}%
	\begin{minipage}{0.15\textwidth}
		\centering
		\includegraphics[width=1.1\linewidth]{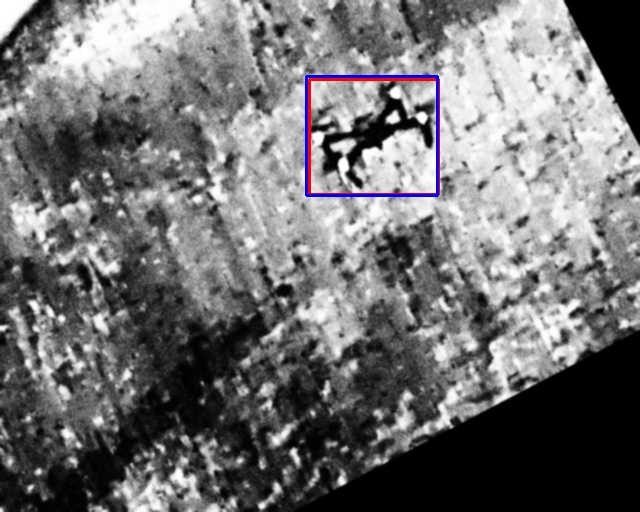}
		\label{fig:ir_results2}
	\end{minipage}%
	\hfill
	\begin{minipage}{0.15\textwidth}
		\centering
		\includegraphics[width=1.1\linewidth]{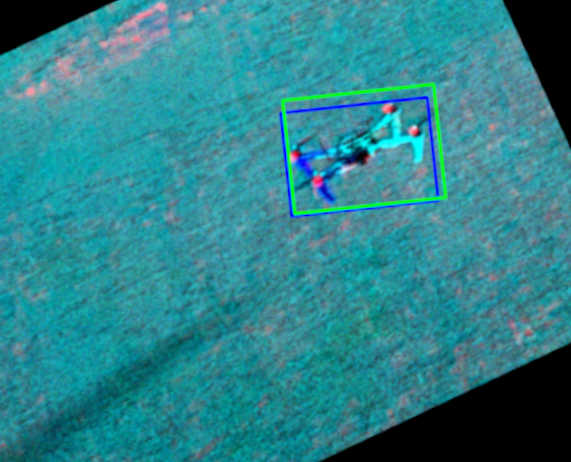}
		\label{fig:rgt_results2}
	\end{minipage}
	\caption{Comparison between blue \textbf{ground truth} boxes and models \textbf{predicted} boxes (red for Late-Fusion and green for Early-Fusion).}
    \label{fig:sovraposition}
\end{figure}

This section presents a qualitative and quantitative comparison between the two proposed fusion approaches. Table \ref{tab:mean_fusion_comparison} summarizes the validation metrics computed from both modalities.
\begin{table}[!ht]
	\centering
	\footnotesize
	\caption{Comparison of mean metrics between Late-Fusion (RGB-IR) and Early-Fusion (RGT) strategies (where mR is mean Recall and mF1 is mean F1).}
	\begin{tabular}{|c|c|c|c|}
		\hline
		\textbf{Fusion Method} & \textbf{mAP@50} & \textbf{mR} & \textbf{mF1-score} \\
		\hline
		Late-Fusion (RGB-IR) & 72\% & 77\% & 73.7\% \\
		\textbf{Early-Fusion (RGT)} & \textbf{75.7\%} & \textbf{83\%} & \textbf{77.8\%} \\
		\hline
	\end{tabular}
	\label{tab:mean_fusion_comparison}
\end{table}
In particular, Late-Fusion requires two independent detection models to process the RGB and IR inputs, resulting in increased inference time and computational cost compared to the Early-Fusion approach. and processes a 640$\times$640 input, resulting in approximately twice the model and inference requirements of RGT.\\
The evaluation process for both modalities follows a common series of steps. In particular, validation was conducted on a pool of $N=400$ spatially aligned images, while inference was performed using a default Intersection-over-Union (IoU) threshold of $0.5$. The image pool comprises several images acquired in the same scenarios to those considered in the experiments, involving different backgrounds, flight altitudes and lighting conditions. All captured images were aggregated into a single evaluation pool, as considering the individual scenarios separately would have resulted in datasets that were too small to support a meaningful comparison. In addition, this diversity of acquisition conditions increases the variability of the evaluation pool, providing a more representative assessment of the model's ability to generalize across different operating conditions.\\
The experimental setup includes various scenarios involving different times of day, environmental conditions, and flight altitudes, whereas the final comparison is
presented only as a set of aggregated results. It is recommended that the authors report the detection results of the early-fusion and decision-level-fusion methods separately
under different experimental conditions. For example, the results could be grouped according to time period, background type, flight altitude, target distance, or zoom level.\\
For the Early-Fusion (RGT) approach, predictions generated by the detector for each image were compared with the corresponding ground-truth annotations using an overlap mechanism. This step allowed the computation of global True Positive, False Positive and False Negatives in the entire pool of images.
\begin{equation}
TP = \sum_{i=1}^{N} TP_i
\qquad
FP = \sum_{i=1}^{N} FP_i
\qquad
FN = \sum_{i=1}^{N} FN_i
\end{equation}
These quantities were then used to compute average precision, recall and F1-score.\\
In contrast, the Late-Fusion approach performs inference independently over each RGB-IR image pair. Then, a set of fused bounding boxes is produced by the weighted average algorithm described in Equation \ref{eq:fbb}. These fused detections are defined as:
\begin{equation}
\hat{\mathcal{D}}_i =
\left\{
B_f^{(1)},
B_f^{(2)},
\ldots,
B_f^{(M_i)}
\right\},
\end{equation}
This set was then compared with the set of ground-truth annotations $\mathcal{G}_i$.
\begin{equation}
\mathcal{G}_i =
\left\{
G_i^{(1)},
G_i^{(2)},
\ldots,
G_i^{(N_i)}
\right\},
\end{equation}
Since the image pair is composed of geometrically aligned RGB and IR frames, they share a single annotation file. Therefore, the label file is unique across both spectra. This allows the comparison between fused boxes and ground-truth.\\
From the results shown in Table \ref{tab:mean_fusion_comparison}, Early-Fusion turned out to be the better of the two, achieving the best overall performance. This point also reinforces the results obtained from previous trainings in Table \ref{tab:metrics}. Despite the final outcome, Late-Fusion strategy still provides competitive results, demonstrating the effectiveness of combining complementary information for aerial target detection.\\
Finally, some detection results showing the overlapping step performed by the validation process are visible in Figure \ref{fig:sovraposition}.

\section{Conclusion and Future Work}\label{Sec:Conclusion}
It has been quite difficult to identify the best strategy of the two fusion techniques. In fact, both early and late fusions gave promising outputs. However, they could be exploited in different contexts. For example, late fusion might be advantageous at night, when the RGB counterpart is completely lost, whereas IR offers better performance and could support it. On the other hand, early fusion is the best for general use, as it improves detection by adopting channel blending, considering the tests conducted and the results acquired. Despite this advantage, early fusion is harder to implement and run, due to the need to train a dedicated model over a fused dataset and its dependence on an accurate pre-alignment process. \\
Although the system reached good overall results, both mathematically and empirically, the need for more multimodal data for future research remains crucial. In particular, the proposed solution was specifically designed for drone-to-drone detection architectures. For this reason, it is necessary to integrate this work with more data, in order to adapt this strategy to different detection pipelines.\\
Some future improvements to the project will surely involve an upgrade to the drone platform, enabling the use of newer and more accurate sensors for detection. This is indeed linked to the exploitation of different and complex image types, such as multispectral or hyperspectral images. However, the use of other sensor data should also be taken into consideration, since radar platforms are widely used for developing sensor fusion paradigms in modern aviation.

\bibliographystyle{IEEEtran}
\bibliography{references}

@article{Samadzadegan2022DroneDetection,
  author       = {Samadzadegan, Farhad and Dadrass Javan, Farzaneh and Ashtari Mahini, Farnaz and Gholamshahi, Mehrnaz},
  title        = {Detection and Recognition of Drones Based on a Deep Convolutional Neural Network Using Visible Imagery},
  journal      = {Aerospace},
  year         = {2022},
  volume       = {9},
  number       = {1},
  pages        = {31},
  doi          = {10.3390/aerospace9010031},
  url          = {https://doi.org/10.3390/aerospace9010031},
  publisher    = {MDPI}
}

@ARTICLE{9743918,
  author={Scholes, Stirling and Ruget, Alice and Mora-Martín, Germán and Zhu, Feng and Gyongy, Istvan and Leach, Jonathan},
  journal={IEEE Access},
  title={DroneSense: The Identification, Segmentation, and Orientation Detection of Drones via Neural Networks},
  year={2022},
  volume={10},
  number={},
  pages={38154-38164},
  doi={10.1109/ACCESS.2022.3162866}
}

@misc{yolov8_multispectral_2025,
  author       = {Glenn Jocher and Ayush Chaurasia and Jing Qiu},
  title        = {Ultralytics YOLOv8 (v8.3.112)},
  howpublished = {https://github.com/ultralytics/ultralytics},
  year         = {2025},
  note         = {Accessed: 2026-05-17},
}

@article{zhao2022vision,
	author = {Jie Zhao and Jingshu Zhang and Dongdong Li and Dong Wang},
	title = {Vision-Based Anti-UAV Detection and Tracking},
	journal = {IEEE Transactions on Intelligent Transportation Systems},
	volume = {23},
	number = {12},
	pages = {25323--25334},
	year = {2022}
}

@article{skorka2022tradeoffs,
	author = {Orit Skorka and Radu Ispasoiu},
	title = {Tradeoffs With RGB-IR Image Sensors},
	journal = {IEEE Transactions on Electron Devices},
	volume = {69},
	number = {6},
	pages = {2915--2923},
	year = {2022}
}

@INPROCEEDINGS{8588047,
  author={Escobar Díaz, Osvaldo and Sánchez López, Abraham and Bernábe Loranca, María Beatriz and González Velázquez, Rogelio},
  booktitle={2016 Fifteenth Mexican International Conference on Artificial Intelligence (MICAI)},
  title={Real Time Drone Object Tracking Using Histograms of Oriented Gradients and Particle Filters},
  year={2016},
  pages={35-40},
  doi={10.1109/MICAI-2016.2016.00014},
  ISSN={},
  month={Oct}
}

@InProceedings{10.1007/978-3-319-46448-0_2,
author="Liu, Wei
and Anguelov, Dragomir
and Erhan, Dumitru
and Szegedy, Christian
and Reed, Scott
and Fu, Cheng-Yang
and Berg, Alexander C.",
editor="Leibe, Bastian
and Matas, Jiri
and Sebe, Nicu
and Welling, Max",
title="SSD: Single Shot MultiBox Detector",
booktitle="Computer Vision -- ECCV 2016",
year="2016",
publisher="Springer International Publishing",
address="Cham",
pages="21--37",
isbn="978-3-319-46448-0"
}

@INPROCEEDINGS{7780460,
  author={Redmon, Joseph and Divvala, Santosh and Girshick, Ross and Farhadi, Ali},
  booktitle={2016 IEEE Conference on Computer Vision and Pattern Recognition (CVPR)},
  title={You Only Look Once: Unified, Real-Time Object Detection},
  year={2016},
  volume={},
  number={},
  pages={779-788},
  doi={10.1109/CVPR.2016.91}}

@InProceedings{10.1007/978-3-030-58452-8_13,
author="Carion, Nicolas
and Massa, Francisco
and Synnaeve, Gabriel
and Usunier, Nicolas
and Kirillov, Alexander
and Zagoruyko, Sergey",
editor="Vedaldi, Andrea
and Bischof, Horst
and Brox, Thomas
and Frahm, Jan-Michael",
title="End-to-End Object Detection with Transformers",
booktitle="Computer Vision -- ECCV 2020",
year="2020",
publisher="Springer International Publishing",
address="Cham",
pages="213--229",
isbn="978-3-030-58452-8"
}

@inproceedings{sangam2023transvisdrone,
  title        = {TransVisDrone: Spatio-Temporal Transformer for Vision-based Drone-to-Drone Detection in Aerial Videos},
  author       = {Tushar Sangam and Ishan Rajendrakumar Dave and Waqas Sultani and Mubarak Shah},
  booktitle    = {Proceedings of the 2023 IEEE International Conference on Robotics and Automation (ICRA 2023)},
  year         = {2023},
  address      = {London, UK},
  publisher    = {IEEE}
}

@inproceedings{rebbapragada2024c2fdrone,
  title={C2FDrone: Coarse-to-fine drone-to-drone detection using vision transformer networks},
  author={Rebbapragada, Sairam VC and Panda, Pranoy and Balasubramanian, Vineeth N},
  booktitle={2024 IEEE International Conference on Robotics and Automation (ICRA)},
  pages={6627--6633},
  year={2024},
  organization={IEEE}
}

@article{liu2022infrared,
	author = {Yuanyuan Liu and Zhiyong Wu and Xizhen Han and Qiang Sun and Jian Zhao and Jianzhuo Liu},
	title = {Infrared and Visible Image Fusion Based on Visual Saliency Map and Image Contrast Enhancement},
	journal = {Sensors},
	volume = {22},
	number = {17},
	pages = {6390},
	year = {2022}
}

@article{ma2022swinfusion,
	author = {Jiayi Ma and Linfeng Tang and Fan Fan and Jun Huang and Xiaoguang Mei and Yong Ma},
	title = {SwinFusion: Cross-domain Long-range Learning for General Image Fusion via Swin Transformer},
	journal = {IEEE/CAA Journal of Automatica Sinica},
	volume = {9},
	number = {7},
	pages = {1200--1217},
	year = {2022}
}

@article{tang2022superfusion,
	author = {Linfeng Tang and Yuxin Deng and Yong Ma and Jun Huang and Jiayi Ma},
	title = {SuperFusion: A Versatile Image Registration and Fusion Network with Semantic Awareness},
	journal = {IEEE/CAA Journal of Automatica Sinica},
	volume = {9},
	number = {12},
	pages = {2121--2137},
	year = {2022}
}

@inproceedings{cao2023multimodal,
	author = {Yue Cao and Junchi Bin and Jozsef Hamari and Erik Blasch and Zheng Liu},
	title = {Multimodal Object Detection by Channel Switching and Spatial Attention},
	booktitle = {2023 IEEE/CVF Conference on Computer Vision and Pattern Recognition Workshops (CVPRW)},
	pages = {403--411},
	year = {2023}
}

@inproceedings{xu2024antidrone,
  author = {Fan Xu and Zebin Lin and Chuibin Chen and Ruoshi Yang and Hui Zhai},
  title = {Anti-Drone Detection in Aerial Multispectral Images},
  booktitle = {IGARSS 2024 - 2024 IEEE International Geoscience and Remote Sensing Symposium},
  pages = {9388--9391},
  year = {2024}
}

@inproceedings{zhang2023sfyolo,
  author = {Shoulong Zhang and Ting Xie and Yiqi Wang and Yingying Liu and Yuling Dou and Shengke Wang},
  title = {SF-YOLO: RGB-T Fusion Object Detection in UAV Scenes},
  booktitle = {2023 8th International Conference on Image, Vision and Computing (ICIVC)},
  pages = {51--59},
  year = {2023}
}

@article{yang2024iaifnet,
	author = {Qiao Yang and Yu Zhang and Zijing Zhao and Jian Zhang and Shunli Zhang},
	title = {IAIFNet: An Illumination-Aware Infrared and Visible Image Fusion Network},
	journal = {IEEE Signal Processing Letters},
	volume = {31},
	pages = {1374--1378},
	year = {2024}
}

@ARTICLE{li2024fcdfusion,
  author={Li, Hesong and Fu, Ying},
  journal={Computational Visual Media},
  title={FCDFusion: A Fast, Low Color Deviation Method for Fusing Visible and Infrared Image Pairs},
  year={2025},
  volume={11},
  number={1},
  pages={195-211},
  doi={10.26599/CVM.2025.9450330}
}

@inproceedings{wagner2016multispectral,
  author = {Jörg Wagner and Volker Fischer and Michael Herman and Sven Behnke},
  title = {Multispectral Pedestrian Detection Using Deep Fusion Convolutional Neural Networks},
  booktitle = {24th European Symposium on Artificial Neural Networks, Computational Intelligence and Machine Learning (ESANN)},
  pages = {509--514},
  howpublished = {\url{https://www.ais.uni-bonn.de/papers/ESANN_2016_Wagner.pdf}},
  year = {2016},
  volume={587}
}

@ARTICLE{zhao2024ieee,
  author={Zhao, Tianyi and Yuan, Maoxun and Jiang, Feng and Wang, Nan and Wei, Xingxing},
  journal={IEEE Transactions on Intelligent Transportation Systems},
  title={Removal Then Selection: A Coarse-to-Fine Fusion Perspective for RGB-Infrared Object Detection},
  year={2026},
  volume={27},
  number={2},
  pages={2504-2519},
  doi={10.1109/TITS.2025.3638627}}

@InProceedings{chen2022eccv,
author="Chen, Yi-Ting and Shi, Jinghao and Ye, Zelin and Mertz, Christoph and Ramanan, Deva and Kong, Shu",
editor="Avidan, Shai and Brostow, Gabriel and Ciss{\'e}, Moustapha and Farinella, Giovanni Maria and Hassner, Tal",
title="Multimodal Object Detection via Probabilistic Ensembling",
booktitle="Computer Vision -- ECCV 2022",
year="2022",
publisher="Springer Nature Switzerland",
address="Cham",
pages="139--158",
isbn="978-3-031-20077-9"
}

@inproceedings{liu2023self,
  title     = {Self-Supervised Learning for Fusion of IR and RGB Images in Visual Teach and Repeat Navigation},
  author    = {Liu, Xinyu and Rozsyp{\'a}lek, Zden{\v{e}}k and Krajn{\'\i}k, Tom{\'a}{\v{s}}},
  booktitle = {2023 European Conference on Mobile Robots (ECMR)},
  pages     = {1--7},
  year      = {2023},
  publisher = {IEEE}
}

@inproceedings{wang2019rgb,
  title     = {RGB-Infrared Cross-Modality Person Re-Identification via Joint Pixel and Feature Alignment},
  author    = {Wang, Guan'an and Zhang, Tianzhu and Cheng, Jian and Liu, Si and Yang, Yang and Hou, Zengguang},
  booktitle = {2019 IEEE/CVF International Conference on Computer Vision (ICCV)},
  pages     = {3622--3631},
  year      = {2019},
  publisher = {IEEE}
}

@inproceedings{zhu2017unpaired,
  author = {Jun-Yan Zhu and Taesung Park and Phillip Isola and Alexei A. Efros},
  title = {Unpaired Image-to-Image Translation Using Cycle-Consistent Adversarial Networks},
  booktitle = {2017 IEEE International Conference on Computer Vision (ICCV)},
  pages = {2242--2251},
  year = {2017}
}

@inproceedings{isola2017image,
  author = {Phillip Isola and Jun-Yan Zhu and Tinghui Zhou and Alexei A. Efros},
  title = {Image-to-Image Translation with Conditional Adversarial Networks},
  booktitle = {2017 IEEE Conference on Computer Vision and Pattern Recognition (CVPR)},
  pages = {1125--1134},
  year = {2017}
}

@ARTICLE{6423909,
  author = {Shutao Li and Xudong Kang and Jianwen Hu},
  journal={IEEE Transactions on Image Processing},
  title={Image Fusion With Guided Filtering},
  year={2013},
  volume={22},
  number={7},
  pages={2864-2875},
  doi={10.1109/TIP.2013.2244222}
}

@article{Wang_Chen_Choi_Kuo_2019,
    title={Towards Visible and Thermal Drone Monitoring with Convolutional Neural Networks},
    volume={8}, DOI={10.1017/ATSIP.2018.30},
    journal={APSIPA Transactions on Signal and Information Processing},
    author = {Ye Wang and Yueru Chen and Jongmoo Choi and C.-C. Jay Kuo},
    year={2019},
    pages={e5}
}

@ARTICLE{9916096,

  author={Sengupta, Arindam and Cheng, Lei and Cao, Siyang},

  journal={IEEE Sensors Letters},

  title={Robust Multiobject Tracking Using Mmwave Radar-Camera Sensor Fusion},

  year={2022},

  volume={6},

  number={10},

  pages={1-4},

  doi={10.1109/LSENS.2022.3213529}}

@ARTICLE{9716741,

  author={Han, Kai and Wang, Yunhe and Chen, Hanting and Chen, Xinghao and Guo, Jianyuan and Liu, Zhenhua and Tang, Yehui and Xiao, An and Xu, Chunjing and Xu, Yixing and Yang, Zhaohui and Zhang, Yiman and Tao, Dacheng},

  journal={IEEE Transactions on Pattern Analysis and Machine Intelligence},

  title={A Survey on Vision Transformer},

  year={2023},

  volume={45},

  number={1},

  pages={87-110},

  doi={10.1109/TPAMI.2022.3152247}}

@ARTICLE{7485869,

  author={Ren, Shaoqing and He, Kaiming and Girshick, Ross and Sun, Jian},

  journal={IEEE Transactions on Pattern Analysis and Machine Intelligence},

  title={Faster R-CNN: Towards Real-Time Object Detection with Region Proposal Networks},

  year={2017},

  volume={39},

  number={6},

  pages={1137-1149},

  doi={10.1109/TPAMI.2016.2577031}
}

@inproceedings{sun2024visible,
	author = {Yuxuan Sun and Yuanqin Meng and Qingbo Wang and Minghua Tang and Tao Shen and Qingwang Wang},
	title = {Visible and Infrared Image Fusion for Object Detection: A Survey},
	booktitle = {Proceedings of International Conference on Image, Vision and Intelligent Systems 2023 (ICIVIS 2023)},
	pages = {236--248},
	publisher = {Springer Nature Singapore},
	year = {2024}
}

@article{zamri2024deep,
	author = {Fatin Najihah Muhamad Zamri and Teddy Surya Gunawan and Mira Kartiwi and Agus Pratondo and Siti Hajar Yusoff and Yasir Mohd. Mustafah},
	title = {Deep Learning Techniques for Advanced Drone Detection Systems: A Comprehensive Review of Techniques, Challenges and Future Directions},
	journal = {Indonesian Journal of Electrical Engineering and Informatics},
	volume = {12},
	number = {4},
	pages = {931--944},
	year = {2024}
}

@article{liu2024boost,
	author = {Fan Liu and Liang Yao and Chuanyi Zhang and Ting Wu and Xinlei Zhang and Xiruo Jiang and Jun Zhou},
	title = {Boost UAV-based Object Detection via Scale-Invariant Feature Disentanglement and Adversarial Learning},
	journal = {arXiv preprint},
	volume = {abs/2405.15465},
	pages = {1--12},
	year = {2024}
}

@inproceedings{ashraf2021dogfight,
	author = {Muhammad Waseem Ashraf and Waqas Sultani and Mubarak Shah},
	title = {Dogfight: Detecting Drones from Drone Videos},
	booktitle = {Proceedings of the IEEE/CVF Conference on Computer Vision and Pattern Recognition (CVPR)},
	pages = {7067--7076},
	year = {2021}
}

@article{hidayatullah2025yolov8,
	author = {Priyanto Hidayatullah and Nurjannah Syakrani and Muhammad Rizqi Sholahuddin and Trisna Gelar and Refidinal Tubagus},
	title = {YOLOv8 to YOLO11: A Comprehensive Architecture In-depth Comparative Review},
	journal = {arXiv preprint},
	volume = {abs/2501.13400},
	pages = {1--15},
	year = {2025}
}

@inproceedings{bilal2024cost,
	author = {Muhammad Bilal and Sadaf Rehmat and Naeem Akhtar and Hafsa Gulzar},
	title = {Cost Effective Drone Detection with Deep Neural Networks For Day and Night Surveillance},
	booktitle = {2024 International Conference on Frontiers of Information Technology (FIT)},
	pages = {1--6},
	year = {2024}
}

@article{ozer2024visirnet,
	author = {Sedat Özer and Alain P. Ndigande},
	title = {VisIRNet: Deep Image Alignment for UAV-Taken Visible and Infrared Image Pairs},
	journal = {IEEE Transactions on Geoscience and Remote Sensing},
	volume = {62},
	pages = {1--11},
	year = {2024}
}

@article{guan2018fusion,
  author = {Dayan Guan and Yanpeng Cao and Jun Liang and Yanlong Cao and Michael Ying Yang},
  title = {Fusion of Multispectral Data Through Illumination-aware Deep Neural Networks for Pedestrian Detection},
  journal = {arXiv preprint},
  volume = {abs/1802.09972},
  pages = {1--9},
  year = {2018}
}

@article{tang2022piafusion,
  author = {Linfeng Tang and Jiteng Yuan and Hao Zhang and Xingyu Jiang and Jiayi Ma},
  title = {PIAFusion: A Progressive Infrared and Visible Image Fusion Network Based on Illumination Aware},
  journal = {Information Fusion},
  volume = {83--84},
  pages = {79--92},
  year = {2022}
}

@article{li2019densefuse,
  author = {Hui Li and Xiao-Jun Wu},
  title = {DenseFuse: A Fusion Approach to Infrared and Visible Images},
  journal = {IEEE Transactions on Image Processing},
  volume = {28},
  number = {5},
  pages = {2614--2623},
  month = {May},
  year = {2019}
}

@inproceedings{yang2022drone,
  author = {Lizhi Yang and Ruhang Ma and Avideh Zakhor},
  title = {Drone Object Detection Using RGB/IR Fusion},
  booktitle = {Electronic Imaging Symposium, Computational Imaging XX Conference},
  volume = {2022},
  pages = {1--6},
  year = {2022}
}

@inproceedings{gadzicki2020early,
  title     = {Early vs Late Fusion in Multimodal Convolutional Neural Networks},
  author    = {Gadzicki, Konrad and Khamsehashari, Razieh and Zetzsche, Christoph},
  booktitle = {2020 IEEE 23rd International Conference on Information Fusion (FUSION)},
  pages     = {1--6},
  year      = {2020},
  publisher = {IEEE}
}

@inproceedings{tian2023rgb,
  title     = {RGB-Infrared Multi-Modal Remote Sensing Object Detection Using CNN and Transformer Based Feature Fusion},
  author    = {Tian, Tao and Cai, Jiang and Xu, Yang and Wu, Zebin and Wei, Zhihui and Chanussot, Jocelyn},
  booktitle = {IGARSS 2023 -- 2023 IEEE International Geoscience and Remote Sensing Symposium},
  pages     = {5728--5731},
  year      = {2023},
  publisher = {IEEE}
}

@ARTICLE{5350902,
  author={Cheng, Hui-Chuan and Ben-David, Ilan and Wu, Shin-Tson},
  journal={Journal of Display Technology},
  title={Five-Primary-Color LCDs},
  year={2010},
  volume={6},
  number={1},
  pages={3-7},
  doi={10.1109/JDT.2009.2032247}}

@Article{seidaliyeva2024advances,
	AUTHOR = {Seidaliyeva, Ulzhalgas and Ilipbayeva, Lyazzat and Taissariyeva, Kyrmyzy and Smailov, Nurzhigit and Matson, Eric T.},
	TITLE = {Advances and Challenges in Drone Detection and Classification Techniques: A State-of-the-Art Review},
	JOURNAL = {Sensors},
	YEAR = {2024},
	NUMBER = {1},
	ARTICLE-NUMBER = {125},
	URL = {https://www.mdpi.com/1424-8220/24/1/125},
	PubMedID = {38202987},
	ISSN = {1424-8220},
	DOI = {10.3390/s24010125},
    volume = {24}
}

@article{YANG2024104179,
	title = {A review on infrared and visible image fusion algorithms based on neural networks},
	journal = {Journal of Visual Communication and Image Representation},
	volume = {101},
	pages = {104179},
	year = {2024},
	issn = {1047-3203},
	doi = {https://doi.org/10.1016/j.jvcir.2024.104179},
	url = {https://www.sciencedirect.com/science/article/pii/S1047320324001342},
	author = {Kaixuan Yang and Wei Xiang and Zhenshuai Chen and Jian Zhang and Yunpeng Liu}
}

@misc{mngeo_cir_imagery,
	author       = {{Minnesota Geospatial Information Office}},
	title        = {Color-Infrared (CIR) Imagery},
	howpublished = {\url{https://www.mngeo.state.mn.us/chouse/airphoto/cir.html}},
	note	 = {Accessed: 2026-05-17}
}

@ARTICLE{11177201,
  author={Tavaris, Denis and De Zan, Alberto and Toma, Andrea and Foresti, Gian Luca and Scagnetto, Ivan and Martinel, Niki},
  journal={IEEE Access},
  title={Multi-Perspective RGB and Infrared Dataset for UAV Detection},
  year={2025},
  volume={13},
  number={},
  pages={168792-168803},
  doi={10.1109/ACCESS.2025.3614069},
  ISSN={2169-3536}
}

@article{architecture,
  title={Autonomous real-time drone-to-drone visual detection by onboard hardware platform},
  author={De Zan, Alberto and Tavaris, Denis and Foresti, Gian Luca and Scagnetto, Ivan and Martinel, Niki and Toma, Andrea},
  journal={Integrated Computer-Aided Engineering},
  pages={10692509261452776},
  year={2026},
  volume={33},
  number={3},
  publisher={SAGE Publications Sage UK: London, England},
  doi={10.1177/10692509261452776}
}

@misc{usgs_cir,
  title        = {What Do the Different Colors in a Color Infrared (CIR) Aerial Photograph Represent?},
  author       = {U.S. Geological Survey (USGS)},
  url          = {https://www.usgs.gov/faqs/what-do-different-colors-a-color-infrared-aerial-photograph-represent},
  note         = {Accessed: 2026-05-05}
}

@article{aldosari2024drone,
  author    = {Kholoud AlDosari and AIbtisam Osman and Omar Elharrouss and Somaya Almaadeed and Mohamed Zied Chaari},
  title     = {Drone-type-Set: Drone types detection benchmark for drone detection and tracking},
  journal   = {CoRR},
  volume    = {abs/2405.10398},
  year      = {2024},
  url       = {https://dblp.org/rec/journals/corr/abs-2405-10398.html}
}

@article{hao2025tmrgbtd2d,
  title     = {TMRGBT-D2D: A Temporal Misaligned RGB-Thermal Dataset for Drone-to-Drone Target Detection},
  author    = {Hao, Hexiang and Peng, Yueping and Ye, Zecong and Han, Baixuan and Tang, Wei and Kang, Wenchao and Zhang, Xuekai and Li, Qilong and Liu, Wenchao},
  journal   = {Drones},
  volume    = {9},
  number    = {10},
  pages     = {694},
  year      = {2025},
  doi       = {10.3390/drones9100694},
  publisher = {MDPI}
}

@article{lu2025mtcnet,
  author  = {Lu, Ruitao and Zhuo, Zhanhong and Wang, Siyu and Zhang, Dingwen and Xi, Bobo and Zheng, Xiangtao and Hou, Zhiqiang and Yang, Xiaogang},
  title   = {{MTCNet}: Multimodal Time-Sensitive Target Fast Concealment Network for {UAV} Aerial Imagery via Real-Time Mask Generation},
  journal = {IEEE Transactions on Geoscience and Remote Sensing},
  volume  = {63},
  year    = {2025},
  pages   = {5643818}
}

@article{tang2025r2ploc,
  author  = {Tang, Bin and Lu, Ruitao and Yang, Xiaogang and Li, Yansheng and Li, Yunsong and Zhang, Dingwen and Chen, Shiwei},
  title   = {{R2PLoc}: A Region-to-Point {UAV} Visual Geo-Localization Framework Leveraging Hierarchical Semantic Representation},
  journal = {IEEE Transactions on Geoscience and Remote Sensing},
  volume  = {63},
  year    = {2025},
  pages   = {5643818}
}

@article{zhang2026ccsfuse,
  author  = {Zhang, Tao and Lu, Ruitao and Yang, Xiaogang and Zhang, Dingwen and Li, Yansheng and Xie, Xueli and Li, Yunsong},
  title   = {{CCSFuse}: Collaborative Compensation and Selective Fusion for {UAV}-Based {RGB-IR} Object Detection},
  journal = {IEEE Transactions on Geoscience and Remote Sensing},
  volume  = {64},
  year    = {2026},
  pages   = {5400314}
}

@article{zhang2020IFCNN,
  title={IFCNN: A General Image Fusion Framework Based on Convolutional Neural Network},
  author={Zhang, Yu and Liu, Yu and Sun, Peng and Yan, Han and Zhao, Xiaolin and Zhang, Li},
  journal={Information Fusion},
  volume={54},
  pages={99--118},
  year={2020},
  publisher={Elsevier}
}

\newpage

\begin{IEEEbiography}[{\includegraphics[width=1in,height=1.25in,clip,keepaspectratio]{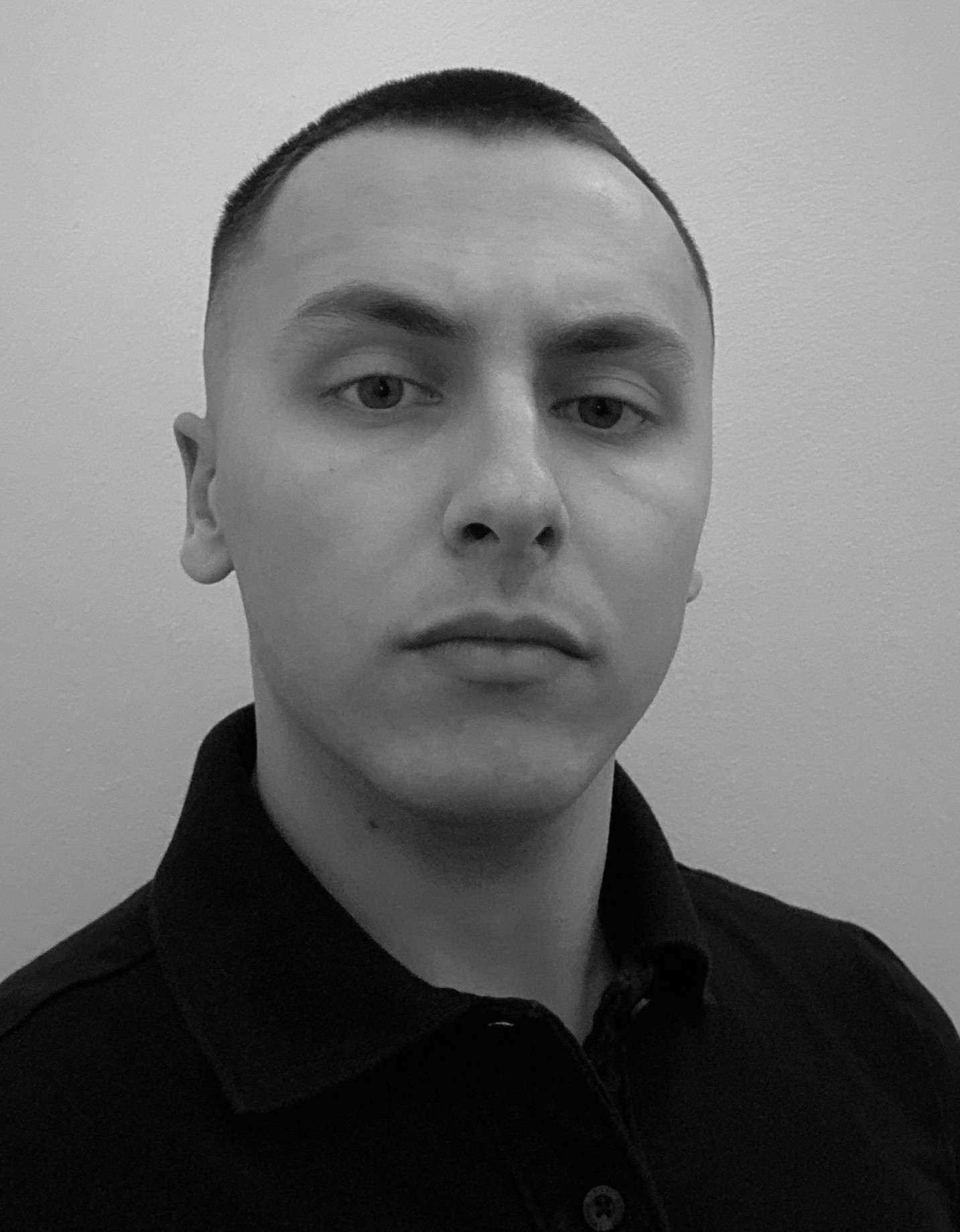}}]{\textbf{Alberto De Zan}} received his Bachelor Degree (cum laude) in Computer Science at the University of Udine, Italy in 2023, on a thesis with title "Drones Cluster Development for Air Surveillance and Defense". He continued his studies with a Double Degree program in Artificial Intelligence \& Cybersecurity at the University of Udine (Italy) and Klagenfurt (Austria). He graduated with maximum marks in 2025 on a thesis with title "Unmanned Aerial Vehicles Detection adopting Multiple Sensor Fusion". He is currently working as a Research Fellow on some projects regarding Computer vision, Radar Platforms, Drone Detection, Open Source Intelligence and Quantum Computing in the Autonomous System Lab of the University of Udine.
\end{IEEEbiography}

\begin{IEEEbiography}[{\includegraphics[width=1in,height=1.25in,clip,keepaspectratio]{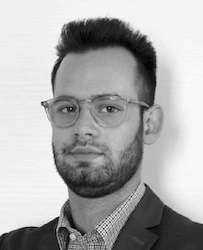}}]{\textbf{Denis Tavaris}} is a Ph.D. student in Artificial Intelligence at the University of Udine, specializing in Computer Vision applied to robotics. His research focuses on the development of autonomous and cooperative multi-drone systems for environmental monitoring, integrating vision-based technologies and sensor fusion. Denis is involved in several research projects on autonomous systems, multi-domain robotics (aerial, surface, and underwater), and AI-driven environmental analysis. He has published papers in international journals and conferences, with a focus on efficient and scalable solutions for autonomous monitoring. Denis has a strong background in data analysis, system design, and software development, with expertise in Python, Linux, and sensor integration.
\end{IEEEbiography}

\begin{IEEEbiography}[{\includegraphics[width=1in,height=1.25in,clip,keepaspectratio]{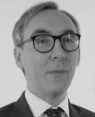}}]{\textbf{Gian Luca Foresti}} received the Laurea degree (cum laude) in electronic engineering and the Ph.D. degree in computer science from the University of Genoa, Genoa, Italy, in 1990 and 1994, respectively. He is a Full Professor of computer science with the University of Udine, Udine, Italy, where he is the Deputy Director of the Department of Mathematics, Computer Science and Physics. Since 2006, he has been a Visiting Professor in artificial vision with the University of Klagenfurt, Klagenfurt, Austria. From 2000 to 2009, he was the appointed Italian Member of the NATO RTO Information System Technology Panel. He has authored over 300 papers published in international journals and international conferences. He has co-edited several books in the field of multimedia and video surveillance. His current research interests include computer vision and image processing, multisensor data and information fusion, pattern recognition, and neural networks. Mr. Foresti was a recipient of the Best IEEE Vehicular Electronics Paper in 2002. He was the Finance Chair of the 11th IEEE Conference on Image Processing, the General Chair of the 16th International Conference on Image Analysis and Processing and the 8th IEEE International Conference on Advanced Video and Signal Based Surveillance. He has been the Guest Editor of a Special Issue of the Proceedings of the IEEE on “Video Communications, Processing and Understanding for Third Generation Surveillance Systems.” He is a member of IAPR and CVPL.
\end{IEEEbiography}

\begin{IEEEbiography}
[{\includegraphics[width=1in,height=1.25in,clip,keepaspectratio]{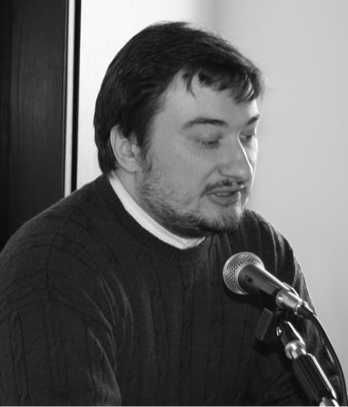}}]{\textbf{Ivan Scagnetto}} received the Ph.D. in Computer Science in 2002 from the University of Udine. Since 2005, he has been an assistant professor at the same University. His research interests range from software certification by means of formal techniques based on Computer Aided Formal Reasoning to mobile computing, information retrieval with mobile devices, and IoT systems. He is author of several scientific papers with national and international peer review and he has been scientific reviewer of international journals and conferences. He worked in several research projects funded by the European Commission, by MUR (the Italian Ministry of University and Research), by the INTERREG programme, and by the PNRM (National Military Research Plan).
\end{IEEEbiography}

\end{document}